\documentclass[11pt]{article}
\usepackage[preprint]{acl}

\usepackage{times}
\usepackage{latexsym}

\usepackage[T1]{fontenc}
\usepackage[utf8]{inputenc}
\usepackage{microtype}
\usepackage{inconsolata}
\usepackage{graphicx}

\usepackage{enumitem}
\usepackage[many]{tcolorbox}
\newcounter{promptbox} 
\usepackage{listings}
\usepackage{float}
\usepackage{booktabs}
\usepackage{placeins}
\usepackage{tabularx}
\usepackage[table]{xcolor}
\usepackage{multirow}
\usepackage{colortbl}
\usepackage{amssymb}
\usepackage{soul}
\usepackage{bm}
\usepackage{standalone}
\usepackage{tikz}
\usetikzlibrary{shapes.geometric, arrows.meta, positioning, calc, shapes.symbols, shadows, fit, backgrounds}
\usepackage{fontawesome5}
\usepackage{array}
\newcolumntype{H}{>{\setbox0=\hbox\bgroup}c<{\egroup}@{}}
\newcolumntype{R}[1]{>{\raggedright\arraybackslash}p{#1}}
\usepackage{subcaption}

\newcommand{\hlc}[2][yellow]{{%
    \colorlet{foo}{#1}%
    \sethlcolor{foo}\hl{#2}}%
}

\newcommand{\fs}[1]{}
\newcommand{\fss}[1]{}
\newcommand{\ah}[1]{}
\newcommand{\mps}[1]{}
\newcommand{\zz}[1]{}

\newcommand{\fsub}{\textsubscript}

\newcommand{\fqt}[1]{``#1''}

\newcommand{\fbf}{\textbf}
\newcommand{\fsc}{\textsc}
\newcommand{\fsl}{\textsl}

\newcommand{\whitecell}[1]{\cellcolor{white}#1}

\usepackage{siunitx} 
\newcommand{\qtyP}[1]{\qty{#1}{\percent}}

\definecolor{ncsuBlue}{RGB}{0,142, 203} 
\definecolor{ncsuTeal}{RGB}{0, 176, 183} 
\definecolor{ncsuRed}{RGB}{199, 53, 45} 
\definecolor{ncsuOrange}{RGB}{231, 135, 35} 
\definecolor{ncsuGreen}{RGB}{124, 185, 43} 
\definecolor{ncsuPurple}{RGB}{141, 109, 172} 
\definecolor{ncsuYellow}{RGB}{237, 179, 31} 
\definecolor{ncsuGray}{RGB}{134, 135, 139} 
\definecolor{ncsuLightBlue}{RGB}{152, 221, 243} 
\definecolor{ncsuLightTeal}{RGB}{172, 223, 223} 
\definecolor{ncsuLightRed}{RGB}{247, 203, 190} 
\definecolor{ncsuLightOrange}{RGB}{242, 211,159} 
\definecolor{ncsuLightGreen}{RGB}{212, 230, 173} 
\definecolor{ncsuLightPurple}{RGB}{225, 205, 227} 
\definecolor{ncsuLightYellow}{RGB}{249, 223, 139} 
\definecolor{ncsuLightGray}{RGB}{205, 206, 211} 
\definecolor{ncsuDarkBlue}{RGB}{33, 70, 109} 
\definecolor{ncsuDarkTeal}{RGB}{40, 92, 103} 
\definecolor{ncsuDarkRed}{RGB}{141, 28, 36} 
\definecolor{ncsuDarkOrange}{RGB}{158, 77, 32} 
\definecolor{ncsuDarkGreen}{RGB}{71, 99, 51} 
\definecolor{ncsuDarkPurple}{RGB}{86, 52, 127} 
\definecolor{ncsuDarkYellow}{RGB}{152, 124, 51} 
\definecolor{ncsuDarkGray}{RGB}{60, 61, 65} 
\definecolor{ncsuGold}{RGB}{255, 184, 28} 

\definecolor{ncsuBanana}{RGB}{254,221,0} 
\definecolor{barDarkBlue}{RGB}{0,0,255} 
\definecolor{barLightBlue}{RGB}{179,179,225} 
\definecolor{barDarkRed}{RGB}{255,0,0} 
\definecolor{barLightRed}{RGB}{255,179,179} 
\definecolor{barDarkGreen}{RGB}{0,255,0} 
\definecolor{barLightGreen}{RGB}{179,255,179} 
\definecolor{pumpkin}{RGB}{255, 116, 23} 
\definecolor{mint}{RGB}{16, 185, 129} 
\definecolor{GoodGreen}{RGB}{76, 175, 80}
\definecolor{EvilRed}{RGB}{229, 57, 53}
\definecolor{PhaseBlue}{RGB}{232, 241, 250}
\definecolor{PhaseYellow}{RGB}{255, 253, 231}
\definecolor{PhaseRed}{RGB}{255, 235, 238}
\definecolor{LeaderGold}{RGB}{255, 215, 0}
\definecolor{cgreen}{HTML}{16a34a}
\definecolor{bgreen}{HTML}{f0fdf4}
\definecolor{camber}{HTML}{d97706}
\definecolor{bamber}{HTML}{fffbeb}
\definecolor{cslate}{HTML}{64748b}
\definecolor{bslate}{HTML}{f8fafc}
\definecolor{ctag}{HTML}{dde3ea}
\definecolor{cnum}{HTML}{d1d5db}

\title{\includegraphics[scale=0.5]{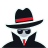}\,\fsc{AVCORP}: Identifying and Resisting Deception via Communication Theory and Theory of Mind}

\author{
  Fardin Saad\textsuperscript{1}, 
  Zhe Zhang\textsuperscript{2}, 
  Sanjana Cheerla\textsuperscript{1},
  Amanul Haque\textsuperscript{3},
  Munindar P. Singh\textsuperscript{1} \\
  \textsuperscript{1}North Carolina State University, \textsuperscript{2}Figma,
  \textsuperscript{3}LexisNexis\\ 
  {\small\textbf{Correspondence:} \url{fsaad@ncsu.edu}} \\
   {\faGithub\,\small\textbf{Code and Corpus:} \url{https://github.com/fardinsaad/_AVCORP_}} \\
   {\raisebox{-0.2em}{\includegraphics[height=1em]{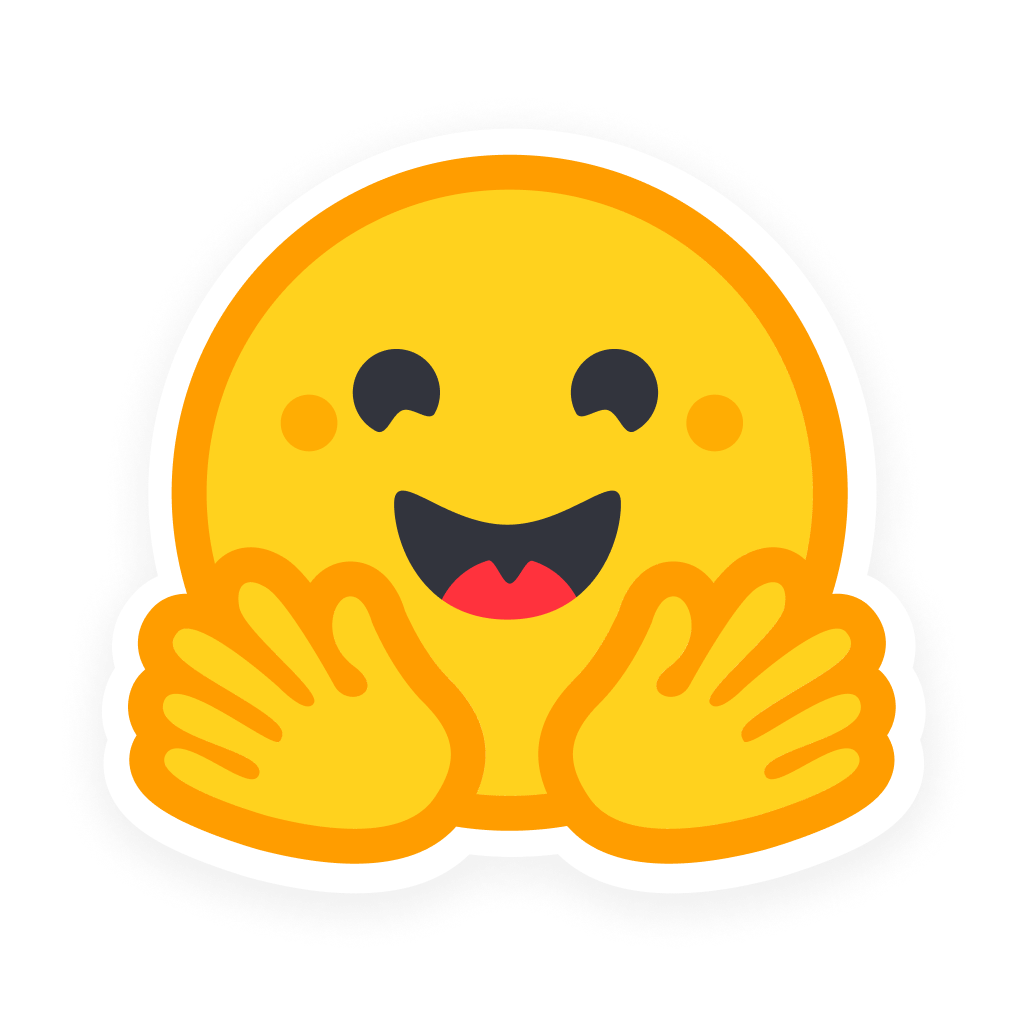}}\,\small\textbf{Dataset and Detector:} \url{https://huggingface.co/collections/fardinsaad/avcorp}}
}

\begin{document}
\maketitle

\begin{abstract}

Today's LLM-based agents are vulnerable to deception by users and other agents.
Defending against deception requires Theory of Mind (ToM) reasoning to infer the beliefs and goals of others as well as modeling \fbf{tactics} based on communication theory.
To enable training models to detect deception, we adopt the well-known deception game, Avalon.
We present a corpus, \fsc{Avcorp}, of 5,000 annotated utterances from 250 games of Avalon.
\fsc{Avcorp} includes (1) 37 \fbf{tactics} mapped to a 4$\times$4 behavior matrix, and (2) \fbf{reasoning traces} capturing abductive hypotheses, suspicion levels, and nested ToM beliefs. 
We train Qwen3-8B on \fsc{Avcorp} and evaluate it on 25 held-out games and 20 games played by humans. 
On held-out games, our best trained detector outperforms its untrained baseline and frontier models (GPT-5.6 Sol, Gemini~3.1 Pro, DeepSeek-V4-Pro), matching Claude Opus~5.5. 
On human games, every trained detector exceeds the untrained model in F1 score across all rounds.

\end{abstract}

\section{Introduction}
\label{sec:intro}


Agents based on Large Language Models (LLMs) are increasingly deployed in practice \citep{Xi+25:LLMAgentsSurvey}.
Besides capabilities and skills, what makes an agent is its mind and its recognition of the minds of others \citep{Dennett87, Self94:book}.
Such \emph{Theory of Mind (ToM)} \citep{Premack+Woodruff-78:ToM-chimps, Frith-05:ToM} endowed agents promise an enhanced ability to cooperate by dealing with incompleteness and ambiguity of communication \cite{AIJ-26:ToM}.

Although attacks on LLMs, such as jailbreaking \citep{Wei+23:Jailbroken}, are as old as LLMs, the rise of ToM-endowed agents creates a new threat.
Specifically, an attacker may use an agent's cognitive reasoning against it not merely by circumventing its rigid guardrails but by \emph{deceiving} it.

We adopt the following working definition of deception.
An agent \emph{deceives} a target when it intentionally leads the target to form or keep a belief that the agent believes false, to advance its own goal \citep{Whiten+Byrne-88:TacticalDeception}.

We focus on an agent being deceived by a user or another agent.
Users deceive agents: attackers told Claude that it was an employee of a legitimate cybersecurity firm conducting defensive testing, and it then carried out their cyberattacks \citep{Anthropic-25:CyberEspionage}.
Agents deceive agents: frontier models do so when it serves an objective \citep{Olson+26:LieCraft}, and other agents are susceptible to deception as they do not reliably distrust goals passed along by the deceptive agent in pursuit of a shared goal \citep{OpenAI-26:HFIncident}.
Since deception by users and agents is accelerating, defenses must accelerate comparably, which requires characterizing the structure of deceptive behavior \citep{Sarkadi-18:Deception}.

\begin{figure*}[htb]
  \includegraphics[width=\textwidth]{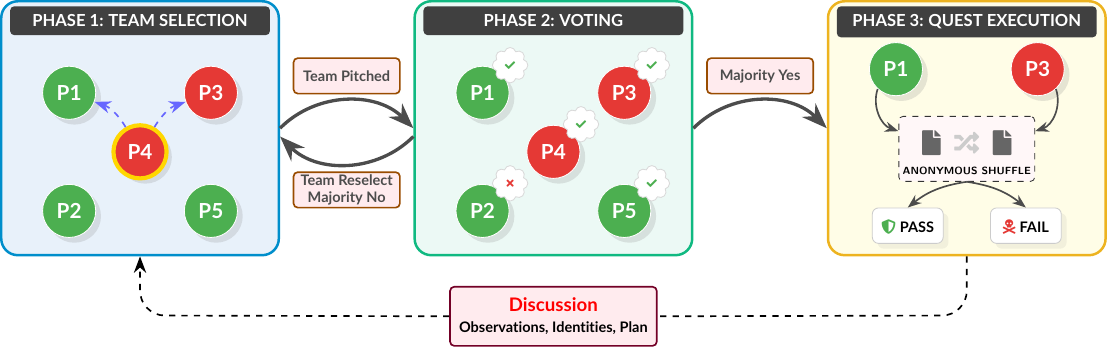}
  \caption{One round of Avalon gameplay. \fbf{Team Selection}: A rotating leader (P4) proposes a quest team (P1 and P3). \fbf{Voting}: All players vote on the proposal; a majority rejection passes leadership to the next player. \fbf{Quest Execution}: Approved team members anonymously submit \fsc{Pass} or \fsc{Fail} cards. A single \fsc{Fail} sabotages the quest. \fbf{Discussion}: Players analyze outcomes, contest identities, and share evidence before the next round begins.}
  \label{fig:flow}
\end{figure*}

By our definition, this structure lies in the mind: a deceiver models the target's beliefs \citep{Walczyk+Fargerson-19:CognitiveDeception}, operating one ToM order above the target.
To combat deception, an investigator must model how the deceiver is modeling them, creating a reciprocal escalation \citep{Sarkadi-23:ArmsRaceToM}. 
Although this escalation demands increasing cognitive depth, behavioral models suggest that humans mostly operate at first- or second-order reasoning \citep{Camerer+04:hierarchy, Chong+Ho+Camerer-16:CognitiveHierarchy}.
Effective detection must therefore operate at this depth, requiring not only attributing beliefs to others (\emph{first-order belief attribution}: \fsl{what does player~X believe?}) but also modeling how others intend to influence those beliefs (\emph{second-order influence attribution}: \fsl{what does player~X want me to believe, and why?}). 
Training a detector to reason at this depth requires corpora that link tactical utterances to these nested ToM reasoning traces.

We present \fsc{Avcorp}, a corpus that makes this connection through \fbf{dual-layer annotations}. 
The first layer annotates utterances using a 37-tactic behavior matrix grounded in Information Manipulation Theory 2 (IMT2) \citep{McCornack+14:IMT2}, Interpersonal Deception Theory (IDT) \citep{Buller+96:IDT}, Truth-Default Theory (TDT) \citep{Levine-14:TDT}, and philosophical accounts of deceptive speech acts \citep{Frankfurt-05:OnBullshit, Carson-10:Lying-Deception} (collectively, Phl).
The second layer records an Investigator's reasoning trace for each round, with abductive hypotheses, suspicion levels, and nested ToM beliefs, and reserves second-order beliefs for strongly suspected players.

Our contributions are as follows:
\begin{description}[nosep]
  \item [Conceptual Model:] Linking \emph{what} is said, through 37 tactics in a 4$\times$4 behavior matrix, to \emph{why}, through nested ToM reasoning (Section~\ref{sec:behavior-matrix}).
  \item [Illustrative Corpus:] 5,000 utterances from 250 games, 4,000 tactic-labeled and 1,000 with ToM reasoning traces, built with a scalable six-stage pipeline (Section~\ref{sec:pipeline}).
  \item [Human-Validated Traces:] The traces deduce both Evil players at game end in 93.2\% of games, and two blind annotators agree with each other and with the judge's selection ($\kappa$ from .92 to .94; Sections~\ref{sec:gold-ceiling} and \ref{sec:iaa}). 
  \item [Trained Detector:] On held-out games, our best detector outperforms three frontier models and matches Claude Opus~5.5; on human games, every trained detector exceeds the untrained model in F1 score (Section~\ref{sec:detector}).
\end{description}

\begin{figure*}[htb]
  \includegraphics[width=\textwidth]{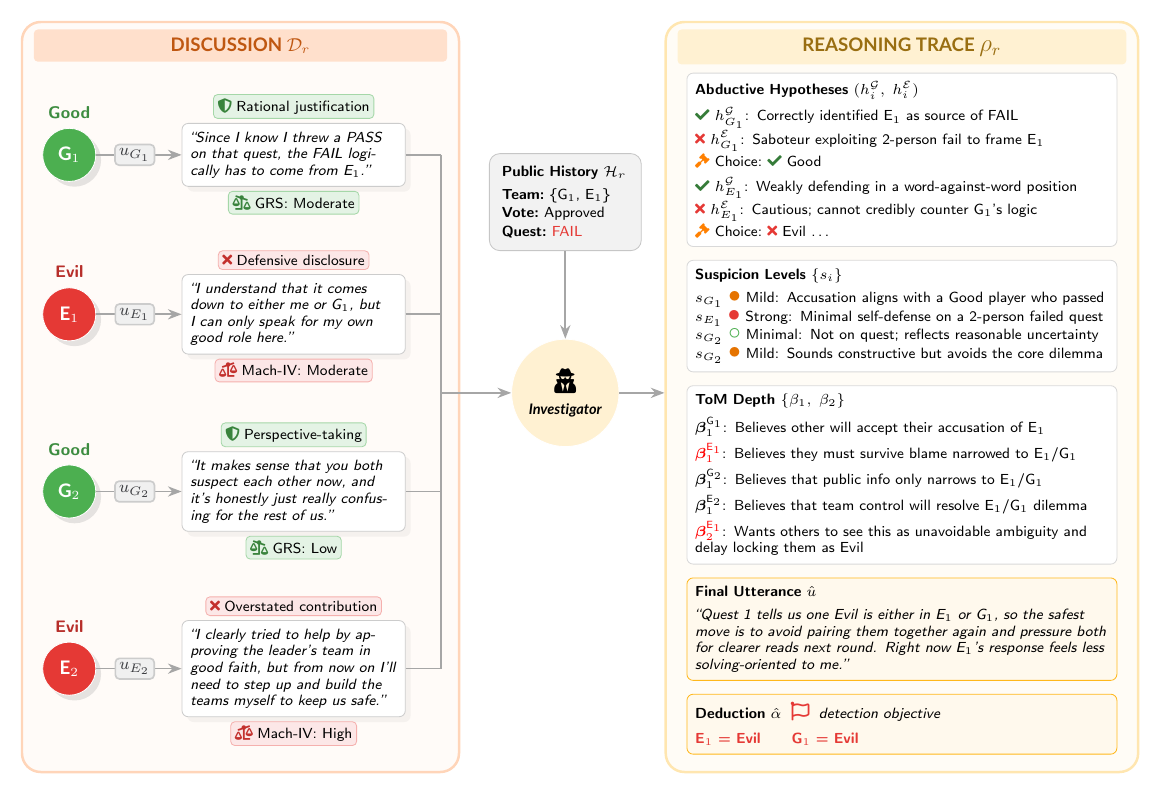}
\caption{One \fsc{Avcorp} sample from Round~1 of Game~60. After a two-person quest returns \textsc{Fail}, G$_1$ accuses E$_1$ (\emph{Rational justification}); E$_1$ deflects (\emph{Defensive disclosure}); G$_2$ expresses uncertainty (\emph{Perspective-taking}); E$_2$ overclaims future leadership (\emph{Overstated contribution}). The Investigator abduces E$_1$ as likely Evil ($h_{E_1}^\mathcal{E}$), assigns Strong suspicion ($s_{E_1}$), and infers E$_1$'s second-order belief that ambiguity will delay accusation ($\beta_2^{E_1}$). The trace concludes with the Investigator's utterance~$\hat{u}$ and deduction~$\hat{\alpha}$ ($h_{i}^\mathcal{G}$, $h_{i}^\mathcal{E}$, $s_i$ and $\beta_1^i$, $\beta_2^i$ are summarized versions).}
  \label{fig:setup}
\end{figure*}

\section{Avalon Gameplay and Design}
\label{sec:gameplay}

\fsl{Avalon: The Resistance} offers a controlled testbed for this problem. 
It is a \fqt{Game of Thoughts} \citep{Wang+24:LLM-recursive-deception} with extreme informational asymmetry: Good players share evidence, whereas Evil players frame and fabricate to conceal their identities. 
Existing resources capture parts of what a detector needs, such as human strategy annotations \citep{Stepputtis+23:AvalonLLM, Li+25:InMind} and statement-level deception labels \citep{Agarwal+25:WOLF}, but they do not connect \emph{what} is said to \emph{why} it is said through nested ToM reasoning.

\fsl{Avalon} is a social deduction game where players are secretly assigned to opposing factions (a Good majority and an Evil minority) \citep{Wiki:AvalonResistance}. 
Evil players know all alignments and conceal themselves to disrupt the Good faction, whereas Good players know only their own and must identify Evil players through open discussion.

We keep only the \textit{Good} and \textit{Evil} alignments, removing special roles (e.g., Merlin, Assassin), and designate one Good player as the \emph{Investigator}.
The non-Investigator players serve as \emph{contextual agents} $(\mathcal{C})$. 
Their discussion provides the behavioral cues that the Investigator uses to detect deception. A game lasts at most five rounds and ends when either faction wins three quests.
Each game has three Good and two Evil players, and each round has three phases followed by a discussion (Figure~\ref{fig:flow}).




\noindent\textbf{Team Selection.} Each game's first leader $\ell_1$ is assigned, and leadership then rotates round-robin. 
In each round $r$, the leader $\ell_r$ proposes a quest team $T_r$ of two or three players (Table~\ref{tab:team-size}, Appendix~\ref{app:design}).
 

\noindent\textbf{Voting.} All players vote on the proposed team ($V_r$), but every proposal receives a passing majority.
This ensures complete games and preserves the voting record as evidence for the Investigator.

\noindent\textbf{Quest Execution.} Approved team members secretly submit a \fsc{Pass} or \fsc{Fail} card, and a single \fsc{Fail} fails the quest ($Q_r$).
Good players always pass, whereas Evil players may fail a quest or pass it to maintain cover. 

\noindent\textbf{Discussion.}
After each quest, players discuss without strict turn-taking. 
Each contextual agent contributes one utterance $u_i$ of one to three sentences (Figure~\ref{fig:setup}), generated from our taxonomy (Section~\ref{sec:behavior-matrix}). 
The Investigator participates last, contributing the final utterance $\hat{u}$ as the penultimate output of its reasoning trace.
%


Each round $r$ adds $\mathcal{H}_r = (\ell_r, T_r, V_r, Q_r)$ to the history; its discussion is $\mathcal{D}_r = \{u_i\}_{i \in \mathcal{C}}$.
After round $r$, the Investigator reasons over all rounds so far using the history $\mathcal{H}_{1:r}$ and discussion $\mathcal{D}_{1:r}$ to construct its reasoning trace $\rho_r$.

\noindent\textbf{Reasoning Trace.}
The Investigator's output for each sample is a \fbf{reasoning trace} $\rho_r$, which serves as the gold label for deception detection.
Formally:
\[
\begin{aligned}
\rho_r = \bigl( & \{h_i\}_{i \in \mathcal{C}},\; \{s_i\}_{i \in \mathcal{C}},\; \{\beta_1^{i}\}_{i \in \mathcal{C}}, \\
              & \{\beta_2^{i}\}_{i \in \mathcal{C}_{s_i \geq \mathrm{Strong}}},\; \hat{u}, \; \hat{\alpha}\bigr) \text{ where,}
\end{aligned}
\]
$h_i=(h_i^{\mathcal{G}}, h_i^\mathcal{E})$ are abductive hypotheses assuming Good and Evil alignments (both grounded in $\mathcal{H}_{1:r}$ and $\mathcal{D}_{1:r}$), $s_i \in$ $\{$Minimal, Mild, Moderate, Strong, Extreme$\}$ is the suspicion level, $\beta_1^{i}$ and $\beta_2^{i}$ are first- and second-order beliefs (the latter only for Strong/Extreme suspicion), $\hat{u}$ is the Investigator's utterance, and $\hat{\alpha}$ is the per-round deduction of suspected Evil players (Figure~\ref{fig:setup}). 


\section{Understanding Deception: A Taxonomy}
\label{sec:behavior-matrix}

Table~\ref{tab:tactic_mapping_with_color} presents our taxonomy of deceptive and truthful communication to guide the dialogue generation.
Rows reflect \textbf{information commitment strategies} (grounded in IMT2 and Phl) and columns reflect \textbf{social goal orientations} (grounded in IDT and TDT).
The cells list tactics (37 in all); Appendix~\ref{app:tactics} provides their detailed definitions.

\begin{table*}[htb]
\centering
\caption{\textbf{4$\times$4 Behavior Taxonomy: 37 Deception Tactics.} Columns are derived from IDT and TDT; rows are from IMT2 and Phl. Social goal refers to whether a player is Cooperative (only if Good), Defensive,  Opportunistic (only if Evil), and Adversarial (only if Evil). The tactics are described in Appendix~\ref{app:tactics}.}
\label{tab:tactic_mapping_with_color}
\resizebox{\textwidth}{!}{
\begin{tabular}{R{3.7cm} R{3.2cm} R{3.3cm} R{3.5cm} R{3.2cm}}
\toprule
& \multicolumn{4}{c}{\textbf{Social Goal}}\\\cmidrule{2-5}
\textbf{Info Commitment} &
\textbf{Cooperative:} Advancing shared goals honestly &
\textbf{Defensive:} Protecting one's own position or image &
\textbf{Opportunistic:} Exploiting situations for personal advantage &
\textbf{Adversarial:} Undermining or sabotaging others \\
\midrule
\textbf{Transparent:} Openly sharing clear, factual information &
\cellcolor{green!20}\begin{tabular}[c]{@{}l@{}}Evidence sharing\\ Rational justification\\ Perspective-taking\end{tabular} &
\cellcolor{blue!20}\begin{tabular}[c]{@{}l@{}}Defensive disclosure\\ Self-justification\end{tabular} &
\cellcolor{red!20}\begin{tabular}[c]{@{}l@{}}Weaponized candor\\ Overstated contribution\end{tabular} &
\cellcolor{red!20}\begin{tabular}[c]{@{}l@{}}Truthful misdirection\\ Misleading emphasis\end{tabular} \\
\midrule
\textbf{Framing:} Shaping discourse through omission or emphasis &
\cellcolor{green!20}\begin{tabular}[c]{@{}l@{}}Bayesian hedging\\ Strategic uncertainty\\ Pragmatic silence\end{tabular} &
\cellcolor{blue!20}\begin{tabular}[c]{@{}l@{}}Omission to save face\\ Half-truths\end{tabular} &
\cellcolor{red!20}\begin{tabular}[c]{@{}l@{}}Cherry-picking\\ False consensus\\ Misframing\end{tabular} &
\cellcolor{red!20}\begin{tabular}[c]{@{}l@{}}Withholding info\\ Strategic omission\end{tabular} \\
\midrule
\textbf{Careless-to-truth:} Speaking without regard for factual accuracy &
\cellcolor{green!20}\begin{tabular}[c]{@{}l@{}}Noisy sincerity\\ Over-explaining\end{tabular} &
\cellcolor{blue!20}\begin{tabular}[c]{@{}l@{}}Face-saving blather\\ Vague filler\end{tabular} &
\cellcolor{red!20}\begin{tabular}[c]{@{}l@{}}Bullshitting\\ Technobabble\end{tabular} &
\cellcolor{red!20}\begin{tabular}[c]{@{}l@{}}Obfuscatory nonsense\\ Confusion tactics\end{tabular} \\
\midrule
\textbf{Counterfactual:} Asserting information known to be false &
\cellcolor{green!20}\begin{tabular}[c]{@{}l@{}}Honest mistake\\ Sincere false belief\end{tabular} &
\cellcolor{blue!20}\begin{tabular}[c]{@{}l@{}}Deflection\\Self-deception\\ Rationalized error\end{tabular} &
\cellcolor{red!20}\begin{tabular}[c]{@{}l@{}}Reverse accusations\\ Strategic fabrication\end{tabular} &
\cellcolor{red!20}\begin{tabular}[c]{@{}l@{}}Hard lying\\ Gaslighting\\ Bold falsehoods\end{tabular} \\
\bottomrule
\end{tabular}
}
\end{table*}
%
%
Figure~\ref{fig:setup} shows four tactics in context.
The Investigator sees only the utterances and history, not tactic labels. 
%
%
We rate a contextual agent's \emph{skill level} $\lambda_i$ ($i \in \mathcal{C}$) on two scales: GRS and Mach-IV.
\begin{figure*}[htb]
    \includegraphics[width=0.99\textwidth]{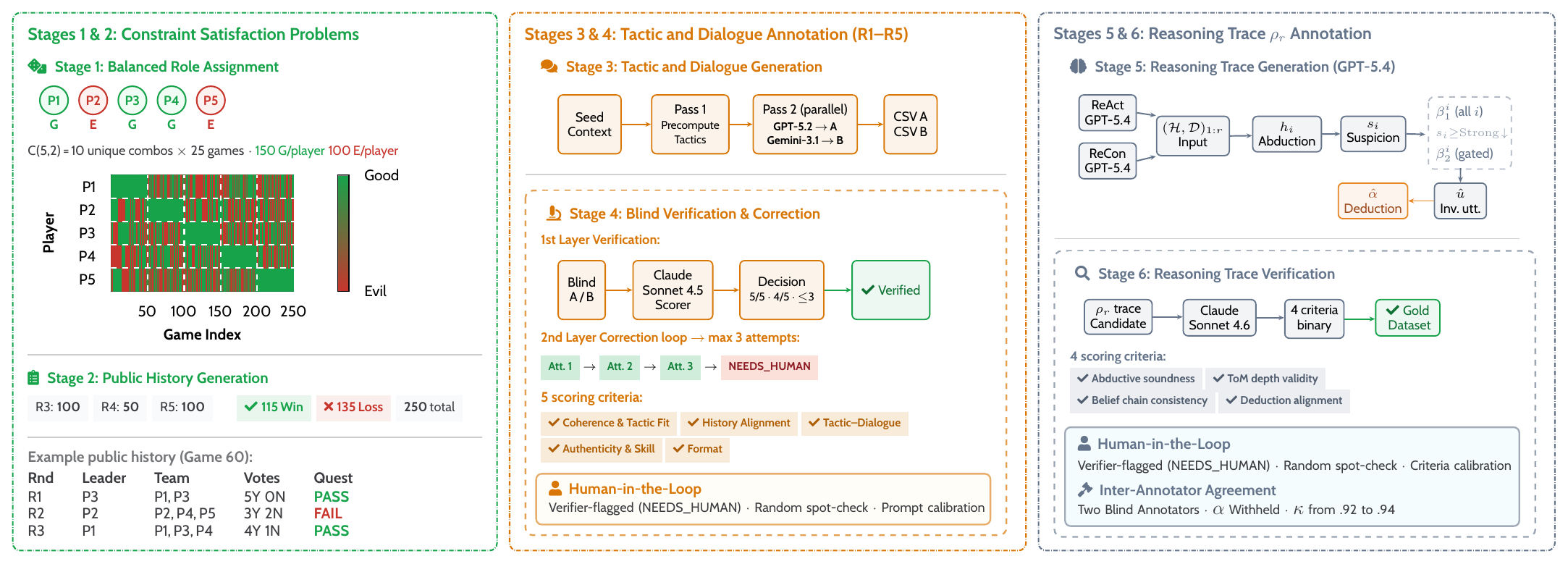}
    \caption{Six-stage pipeline for building \fsc{Avcorp}: Stages~1--2 (far left) construct the game record, Stages~3--4 (center) the tactic-annotated dialogue, and Stages~5--6 (far right) the reasoning traces.
    } 
    \label{fig:combined_stages}
\end{figure*}

\noindent\textbf{Goodness Rationality Spectrum (GRS).} 
GRS is based on theories of cognitive control \citep{Miller+Cohen-01:PrefrontalCortex, Braver-12:Cog-Control}.
It rates a Good player's reasoning in terms of logical soundness, transparency, and reasoning under uncertainty as: 
  \begin{itemize}[nosep]
    \item \textbf{High ($H$)}: 
    Provides rigorous, evidence-based, and well-founded justifications.
    \item \textbf{Moderate ($M$)}: Generally rational but susceptible to minor logical oversights.
    \item \textbf{Low ($L$)}: Displays fragmented reasoning; may appear confused, erratic, or emotional.
  \end{itemize}

\noindent\textbf{Mach-IV (Machiavellianism).} 
We adapt \citeposs{Christie+13:Machiavellianism} Mach-IV scale of manipulation to categorize Evil players into three levels:
  \begin{itemize}[nosep]
    \item \textbf{High ($H$)}: Expert manipulator employing subtle deception and strategic planning.
    \item \textbf{Moderate ($M$)}: Competent deceiver using standard tactics with occasional slips.
    \item \textbf{Low ($L$)}: Clumsy deceiver prone to obvious \emph{tells} like self-contradiction.
  \end{itemize}

The skill level modulates how the tactic is expressed in dialogue. 
For instance, a \fqt{Hard lying} tactic performed by a High Mach-IV Evil player is subtle and believable, whereas the same tactic used by a Low Mach-IV Evil player is easily refuted.
Figure~\ref{fig:setup} demonstrates both scales: G$_1$ (GRS: Moderate) delivers a precise, evidence-grounded accusation, whereas G$_2$ (GRS: Low) offers only an uncertain and confused observation.
E$_1$ (Mach-IV: Moderate) produces a measured self-defense, and E$_2$ (Mach-IV: High) pivots smoothly to overclaiming their fitness to lead next.

\section{Dataset Construction Pipeline}
\label{sec:pipeline}

We build \fsc{Avcorp} with a six-stage pipeline (Figure~\ref{fig:combined_stages}) on a modified \fsl{Avalon} (Section~\ref{sec:gameplay}) in which deception detection is the explicit objective of the \fsl{Investigator}.
The pipeline first fixes each game, with balanced Good and Evil roles (Section~\ref{sec:s1}) and a constraint-satisfying history (Section~\ref{sec:s2}). 
It then generates and verifies both layers using three LLM generators (Gemini~3.1 Pro, GPT-5.2, GPT-5.4) and two LLM judges (Claude Sonnet~4.5 and~4.6) \citep{Zheng+23:LLM-Judge}, keeping each utterance and trace consistent with that game state (Sections~\ref{sec:s3}--\ref{sec:s6}).
Because each utterance is conditioned on its game state, one of the 37 tactics, and one of three skill levels (Section~\ref{sec:behavior-matrix}), no two tactic-labeled utterances are identical.
Two annotators blind to the ground-truth roles $\alpha$ validate the judge's trace selections (Section~\ref{sec:iaa}).

We define each \textbf{game record} as $g = (\mathcal{P}, \alpha, \mathcal{H})$, where $\mathcal{P} = \{p_1, \ldots, p_5\}$ is the player set, $\alpha\colon \mathcal{P} \to \{G, E\}$ assigns exactly two Evil players, and $\mathcal{H} = \mathcal{H}_{1:R}$ is the history of all $R \leq 5$ rounds (Section~\ref{sec:gameplay}).
For each round $r$, an \textbf{annotated sample} extends the game record to $\xi_r = (g, \mathcal{A}_\mathcal{C}, \mathcal{D}_r, \rho_r)$. 
Here, $\mathcal{A}_\mathcal{C}$ assigns a tactic $\tau_i \in \mathcal{T}$ and a skill level $\lambda_i$ to each contextual agent $i \in \mathcal{C}$.
$\mathcal{D}_r$ contains the round's generated utterances, and $\rho_r$ is the Investigator's reasoning trace. 
The six stages (Appendix~\ref{app:six-stages}) construct $g$ (Stages 1--2), $\mathcal{A}_\mathcal{C}$ and $\mathcal{D}_r$ (Stages 3--4), and $\rho_r$ (Stages 5--6). 

\subsection{Stage 1: Player Role Augmentation}
\label{sec:s1}

Stage 1 generates hidden role assignments $\alpha$ for 250 games. 
To ensure role parity across the $\binom{5}{2} = 10$ possible Evil pairs, we generate 225 assignments (supplementing 25 human-annotated seed $R$1 games) satisfying the following constraints.

\noindent\textbf{Combinatorial balance.}
To prevent identity bias, each of the 10 Evil pairs appears 25 times ($250/10$), so each player is Evil in exactly 100 games.

\noindent\textbf{Investigator rotation.}
We split the corpus into five consecutive 50-game blocks and fix player $p_k$ as the Investigator in block $k$, so $p_k$ never appears in an Evil pair within that block.
This \emph{diagonal constraint} appears as a staircase in Figure~\ref{fig:combined_stages}.


\noindent\textbf{Macro-distribution solver.} 
We satisfy these constraints using a solver that assigns patterns to blocks via weighted random sampling. 
At each step, it selects a valid Evil pair (one that excludes the fixed-Good Investigator) with a probability proportional to the remaining capacity.

\subsection{Stage 2: Public History Generation}
\label{sec:s2}

Stage 2 generates the public history $\mathcal{H}$ for 250 games (1,000 rounds), mapping roles to proposals, votes, and outcomes.
We first assign quest outcomes, then sample leaders, role-consistent teams, and votes.
Pairing fixed outcomes with varying roles yields varied game states for Stage~3.

\noindent\textbf{Outcome sequences $Q_r$.}
We assign each game one of 20 predefined outcome sequences (Table~\ref{tab:outcome-sequences}, Appendix~\ref{app:s2}). 
We balance the corpus to include 100 games ending at Round 3, 50 at Round 4, and 100 at Round 5, resulting in 135 Evil and 115 Good victories. Within each block, we cycle through all outcome patterns to ensure representational parity.

\noindent\textbf{Leader rotation $\ell_r$}. 
For the 25 human-annotated seed games, we retain the original $\ell_1$ leadership; for the 225 generated games, we assign $\ell_1$ such that each player leads in exactly 50 of the 250 total games (Figure~\ref{fig:leader-dist}, Appendix~\ref{app:s2}). 
Subsequent leadership rotates clockwise. 

\noindent\textbf{Team selection $T_r$.}
We compose each team using a role-aware selection rule operating on $\alpha$ and a per-player suspicion score $\sigma_p^{(r)}$ (initialized to zero).
On a \fsc{Pass} round, a Good leader chooses the least suspicious players. On a \fsc{Fail} round, they must include at least one Evil player to deliberately test a suspect.
An Evil leader avoids any player whose suspicion exceeds the current group average by more than 1.0 to preserve cover.
We update $\sigma_p^{(r)}$ after each round as $\sigma_p^{(r+1)} = \sigma_p^{(r)} + \Delta_p^{(r)}$ where:
\[
\Delta_p^{(r)} = \begin{cases}
+1.0 & p \in T_r,\; Q_r = \mathrm{FAIL} \\
-0.3 & p \in T_r,\; Q_r = \mathrm{PASS} \\
\phantom{+}0   & \text{otherwise.}
\end{cases}
\]

\noindent\textbf{Voting $V_r$.}  
For each nonmember $p \notin T_r$, let $\pi_p = \Pr[v_p = N]$ denote the No-vote probability:
\fs{read below for interpreting the Voting piecewise function}
{
\[
\pi_p = \begin{cases}
0.6 & \alpha(p)=G,\;\displaystyle\sum_{q \in T_r}\sigma_q^{(r)} > 2.0 \\[4pt]
0.4 & \alpha(p)=G,\;\exists\, q \in T_r: \sigma_q^{(r)} > 1.5 \\[2pt]
0.2 & \alpha(p)=E,\; T_r \cap \alpha^{-1}(E) = \emptyset \\[2pt]
0   & \text{otherwise.}
\end{cases}
\]
}
If $|\{p : v_p = N\}| \geq 3$, we convert the minimum No votes to Yes to restore a majority (see Figure~\ref{fig:voting-pattern} in Appendix~\ref{app:s2} for the distribution). 


\noindent\textbf{Verification.}
We validate Stage~2 against eight hard constraints (e.g., clockwise leader rotation, valid team sizes, and correct game termination).
We observed zero violations across all 250 games.

\subsection{Stage 3: Tactic and Dialogue Generation}
\label{sec:s3}
Stage 3 produces tactic annotations $\mathcal{A}_\mathcal{C}$ and discussion $\mathcal{D}_r$ for the 975 samples across R1--R5, yielding 3,900 contextual-agent utterances. 
The 25 seed games retain their original human annotations.

\noindent\textbf{Tactic annotation.} 
We define six tactic pools, combining alignment and skill levels. 
In Round 1, we cycle through the pools to ensure balanced coverage. 
High-skill agents are restricted to alignment-congruent tactics (Cooperative for Good; Opportunistic/Adversarial for Evil), whereas Moderate- and Low-skill agents may also draw from the 9 shared Defensive tactics. 
From Round 2 onward, GPT-5.2 evolves each agent's tactic from its prior-round assignment, adapting to game-state variables while preserving continuity. 
Evil agents recalibrate based on exposure risk, opportunities for blame, and team composition; Good agents adjust based on accumulated evidence and shifting suspicions. 
Each agent's skill level stays fixed across rounds. 
Full tactic distributions appear in Appendix~\ref{app:s3}.

\noindent\textbf{Dialogue generation.} We generate two candidates (GPT-5.2 and Gemini~3.1 Pro) per sample, conditioned on the game record $g$, assigned tactics $\mathcal{A}_\mathcal{C}$, cumulative history $\mathcal{H}_{1:r}$, preceding verbatim discussion $\mathcal{D}_{r-1}$, and LLM-generated summaries of earlier rounds $\hat{\mathcal{D}}_{1:r-2}$ (Prompts in Appendix~\ref{app:s3}).

\noindent\textbf{Validation.}
We validate each candidate against an eight-point checklist covering format compliance, speaker ordering, tactic-dialogue alignment, and history grounding. All 975 samples satisfy every constraint for both models.

\subsection{Stage 4: Tactic and Dialogue Verification}
\label{sec:s4}
To ensure quality, we employ a two-layer LLM-as-a-judge pipeline, using Claude Sonnet~4.5 as an evaluator blind to model identity.

\noindent\textbf{Scoring criteria and decision logic.} The judge scores candidates on five binary criteria: Coherence and Situational Fit, History Alignment, Tactic-Dialogue Alignment, Skill Consistency, and Format Compliance. We use a layered decision logic:
\begin{itemize}[nosep, leftmargin=1.5em]
    \item \textbf{Layer 1:} If both candidates score 5/5, pairwise comparison determines the winner. 
    If only one scores 5/5, it is selected. 
    If both score 4/5, the judge selects the stronger response and applies a targeted correction to the failing criterion. 
    Scores $\leq$3 trigger full custom regeneration.
    \item \textbf{Layer 2:} All corrected or regenerated outputs undergo a second-pass recheck (max three attempts), after which unverified outputs are flagged as \texttt{NEEDS\_HUMAN} for manual inspection.
\end{itemize}

\noindent\textbf{Human-in-the-loop.}
After Stage~4 verification, we manually review at least 50 sampled dialogues drawn from corrected and regenerated outputs.
We additionally review all samples flagged as \texttt{NEEDS\_HUMAN}.
Based on these reviews, we iteratively refine the generation prompts and verification criteria for Stage~3 and Stage~4.

\noindent\textbf{Verification results.}
Table~\ref{tab:verifier-results} in Appendix~\ref{app:s4} reports per-round outcomes.
GPT-5.2 is selected (directly or after targeted correction) in 84--96\% of samples across rounds, reflecting a quality advantage over Gemini~3.1 Pro (Figure~\ref{fig:s4-trend-avg}).
Pairwise tiebreakers resolve 51--66\% of cases in R2--R5.
R1 requires the most correction (18.2\% targeted), because no earlier round constrains the opening dialogue.
Custom regeneration remains below 1\% across all rounds.
Per-criterion pass rates and prompt configurations appear in Appendix~\ref{app:s4}.

\subsection{Stage 5: Reasoning Trace Augmentation}
\label{sec:s5}
Stage 5 generates the Investigator's reasoning trace $\rho_r$ for all 1,000 verified samples.
We run two prompting frameworks via GPT-5.4, \fsl{ReAct} \citep{Yao+23:ReAct-LLMs} and \fsl{ReCon} \citep{Wang+24:LLM-recursive-deception}, each producing a candidate trace.

\noindent\textbf{Reasoning frameworks.} 
We condition models on the observable game state ($\mathcal{H}_{1:r}$, $\mathcal{D}_{1:r}$) and prior traces $\rho_{1:r-1}$, withholding ground-truth roles $\alpha$ so that no trace sees the answer. 
Our \fsl{ReAct} implementation interleaves two \fsl{Thought--Action--Observation} cycles (first isolating diagnostic history events, then cross-checking them against specific utterances) before a \fsl{Synthesis} step constructs $\rho_r$. 
Our \fsl{ReCon} implementation uses two-pass recursive contemplation: a \fsl{Formulation} pass infers first-order alignments, and a \fsl{Refinement} pass models second-order influence attribution with checks for evidence sufficiency and cross-round consistency. Appendix~\ref{app:s5} details the prompt structures.

\noindent\textbf{Output schema constraints.} Both frameworks output the six-field schema defined in Section~\ref{sec:gameplay}. 
To ensure logical soundness, generations must satisfy the \fbf{Definitive Deduction Rule}: if the Investigator participated in a two-person failed quest, they must definitively identify the partner $p_x$ as Evil (assigning $s_{p_x} = \text{Extreme}$ and ensuring $p_x \in \hat{\alpha}$).

\subsection{Stage 6: Reasoning Trace Verification}
\label{sec:s6}

Stage~6 mirrors Stage~4 with criteria for reasoning traces, selecting the higher-quality trace of the two Stage~5 candidates.
We use Claude Sonnet~4.6 as the judge, which receives Trace~A (ReAct) and Trace~B (ReCon), with identities withheld.

\noindent\textbf{Scoring criteria.}
The judge scores each trace on four binary criteria: (1) Abduction Plausibility, (2) Suspicion Level Consistency, (3) Gated Depth Reasoning, and (4) Statement and Deduction Coherence.
Criterion~1 checks that abductive hypotheses $h_i$ cite specific events.
Criterion~2 ensures suspicion levels $s_i$ match cumulative evidence and enforces the Definitive Deduction Rule.
Criterion~3 verifies first-order beliefs $\beta_1^i$ exist for all players, restricting second-order beliefs $\beta_2^i$ to Strong or Extreme suspects.
Criterion~4 confirms the utterance $\hat{u}$ is evidence-grounded and the deduction $\hat{\alpha}$ includes all high-suspicion players.

\noindent\textbf{Context and decision logic.}
The judge evaluates trace pairs using history $\mathcal{H}_{1:r}$, discussions $\mathcal{D}_{1:r}$ with summaries $\hat{\mathcal{D}}_{1:r}$, prior traces $\rho_{1:r-1}$, and roles $\alpha$. 
We apply a two-layer logic: in Layer 1, if both traces score 4/4, a pairwise comparison selects the one closer to $\alpha$. 
If only one scores 4/4, it is selected directly. 
If both score below 4/4, a targeted correction is applied to the stronger base trace.
In Layer 2, corrected traces undergo up to three rechecks before being flagged.

\noindent\textbf{Human-in-the-loop.}
We follow the same pipeline of manual review and prompt iteration as Stage~4.

\noindent\textbf{Verification results.}
ReAct traces score higher on the criteria than ReCon traces (mean 3.9 against 3.8 of~4; Table~\ref{tab:s6-full}, Appendix~\ref{app:s6}), but the judge selects ReCon in 63.9\% of pairs.
The reason is that 79.1\% of pairs tie at 4/4 (Figure~\ref{fig:s6-sel}), and the pairwise tiebreaker rewards qualitative depth.
Targeted correction applies to only 22 traces (2.2\%), and all of them resolve within Layer~2.

\subsubsection{Deduction Accuracy of Traces}
\label{sec:gold-ceiling}
We score each trace's deductions $\hat{\alpha}$ against the withheld roles $\alpha$ at the round each game ends (R3, R4, or R5).
The traces correctly identify both Evil players in 233 of 250 games (93.2\%, Table~\ref{tab:gold-verdict} in Appendix~\ref{app:gold-ceiling}), against 16.7\% ($1/\binom{4}{2}$) for naming two of the four candidates at random.
Quest outcomes alone single out the Evil pair in only 12 of these games, so the dialogue provides most of the evidence for the verdict (Table~\ref{tab:logic-only}).

\noindent\textbf{Impact of the Definitive Deduction Rule.}
The rule applies in 85 out of 250 games (277 traces).
The traces resolve 84 of these games and 149 of the 165 games in which the rule never applies (90.3\%).

\noindent\textbf{Impact of label priors.}
Each of the 10 Evil pairs occurs in exactly 25 games (Section~\ref{sec:s1}), so the label favors no pair.
Weighting the games that Evil wins (133 of 135 resolved) and that Good wins (100 of 115) equally lowers the verdict only from 93.2\% to 92.7\%.

\subsubsection{Inter-Annotator and Judge Agreement}
\label{sec:iaa}
Two annotators review 100 trace pairs (79 tiebreak vs 21 clean-win) from 25 games sampled at random and stratified by ending round: 10 for R3, 5 for R4, and 10 for R5.
Blind to the framework, the judge's selection, and \emph{ground-truth roles} $\alpha$, they score each trace on the four Stage~6 criteria and select the better one (Appendix~\ref{app:human-study}).
Each framework appears as Trace~A in 50 items and as Trace~B in the other 50 in random order.
Thus, a preference for the first trace shown cannot favor either framework.

The annotators select the same trace on 97 items ($\kappa = .94$, 95\% CI $[.86, 1.00]$), and the judge agrees with them on 96 ($\kappa = .92$ $[.83, .98]$) and 97 items ($\kappa = .94$ $[.86, 1.00]$; Table~\ref{tab:iaa}), all within the almost perfect agreement range ($\kappa \geq .81$) \citep{Landis+Koch-77:kappa-scale}.
On the 79 tiebreak items, where the judge used $\alpha$, it agrees with the annotators on 77 and 78 items and matches all 76 selections they share.
On the 21 clean-win items, both annotators select ReCon over the judge's ReAct on the same two items, in each of which the judge failed the ReCon trace on a criterion that both annotators passed.
Tables~\ref{tab:iaa} and~\ref{tab:iaa-criteria} in Appendix~\ref{app:human-study} report the agreement with PABAK \citep{Byrt+93:PABAK}.

\begin{table*}[t]
\centering
\sisetup{
  text-family-to-math = true,
  text-series-to-math = true,
  round-precision=3,
  round-mode=places
}
\caption{Detector performance across domains. \fbf{Exact}: identifies both Evil players exclusively; \fbf{F1}: game-ending deduction; \fbf{Round F1}: all 82 Avalon-NLU rounds. Trace-only rows: mean of three runs (ranges in parentheses).}
\label{tab:detector}
\resizebox{\textwidth}{!}{
\begin{tabular}{ll cc ccc}
\toprule
\whitecell{} & \whitecell{} & \multicolumn{2}{c}{\textbf{\textcolor{ncsuBlue}{\hlc[ncsuBlue!15]{\mbox{\fsc{Avcorp} (4 candidates)}}}}}
 & \multicolumn{3}{c}{\textbf{\textcolor{ncsuBlue}{\hlc[ncsuBlue!15]{\mbox{Avalon-NLU (5 candidates)}}}}} \\
\cmidrule(lr){3-4}\cmidrule(lr){5-7}
\rowcolor{ncsuBlue!15}
\textbf{\textcolor{ncsuBlue}{Detector}} & \textbf{\textcolor{ncsuBlue}{Stage}} &
\textbf{\textcolor{ncsuBlue}{Exact (\%)}} & \textbf{\textcolor{ncsuBlue}{F1}} &
\textbf{\textcolor{ncsuBlue}{Exact (\%)}} & \textbf{\textcolor{ncsuBlue}{F1}} & \textbf{\textcolor{ncsuBlue}{Round F1}} \\
\midrule
Qwen3-8B, trace prompt  & adapters off & 24.0 & 0.58 & 40.0 & 0.67 & 0.46 \\
Qwen3-8B, tactic prompt & adapters off & 48.0 & 0.72 & 20.0 & 0.55 & 0.43 \\
\midrule
Trace, original rewards & SFT          & 36.0 (24--56) & 0.70 (0.64--0.81) & \bfseries \hlc[ncsuGold!35]{46.7 (40--55)} & 0.71 (0.67--0.76) & 0.58 (0.55--0.60) \\
                        & SFT + GRPO   & 33.3 (24--44) & 0.78 (0.73--0.82) & 30.0 (25--35) & 0.65 (0.64--0.67) & 0.57 (0.53--0.60) \\
Trace, revised rewards  & SFT          & 38.7 (36--44) & 0.69 (0.65--0.74) & 40.0 (30--45) & 0.70 (0.66--0.72) & 0.56 (0.53--0.59) \\
                        & SFT + GRPO   & 64.0 (60--68) & 0.86 (0.84--0.88) & 40.0 (30--50) & \bfseries \hlc[barLightGreen]{0.72 (0.68--0.77)} & \bfseries \hlc[barLightGreen]{0.60 (0.57--0.62)} \\
Tactic, original rewards & SFT         & 52.0 & 0.75 & 35.0 & 0.64 & 0.53 \\
                        & SFT + GRPO   & 48.0 & 0.85 & 20.0 & 0.65 & 0.55 \\
Tactic, revised rewards & SFT          & 48.0 & 0.74 & 45.0 & 0.66 & 0.54 \\
                        & SFT + GRPO   & \bfseries \hl{72.0} & \bfseries \hl{0.88} & 35.0 & 0.70 & 0.58 \\
\bottomrule
\end{tabular}
}
\end{table*}

\section{Trained Detector}
\label{sec:detector}

We finetune Qwen3-8B \citep{QwenTeam-25:Qwen3} with LoRA \citep{Hu+22:LoRA}, using supervised fine-tuning (SFT) followed by Group Relative Policy Optimization (GRPO) \citep{Shao+24:DeepSeek-GRPO}. 
We split the corpus at the game level, stratified by ending round, into 225 training games (900 samples) and 25 held-out games (100 samples).
Each prompt includes the history $\mathcal{H}_{r-2:r}$, the discussion $\mathcal{D}_r$, and the Investigator's prior traces $\rho_{r-2:r-1}$.
SFT runs one epoch on the gold labels, and GRPO then samples 16 completions per prompt for 450 steps.
A full run takes 25.4 to 34.8 GPU hours on an NVIDIA H100 PCIe GPU.
Appendix~\ref{app:parameters} lists all parameters, token caps, and GPU hours. 


\noindent\textbf{Variants and rewards.}
We use two variants for training that differ in their output labels.
The \emph{trace} variant learns $\rho_r$ alone.
The \emph{tactic} variant also learns each contextual agent's tactic.
The \emph{original} set scores (1) schema validity, (2) abduction completeness, (3) suspicion consistency, (4) gated second-order depth, and (5) deduction F1; (3) and (5) use $\alpha$.
The \emph{revised} set replaces (2) with abduction accuracy and (3) with suspicion calibration, which checks that the Evil players receive higher suspicion than the Good players, and extends (1) to check field content; three of them use $\alpha$.
The tactic variant adds a sixth reward for tactic accuracy, and Appendix~\ref{app:rewards} defines every reward.
We train each trace configuration three times and each tactic configuration once, for eight runs.

\noindent\textbf{Evaluation.}
We score the deduction at the round each game ends as an exact match and as F1, and each detector conditions on its own earlier traces (Appendix~\ref{app:gold-chain}).
The untrained baseline is Qwen3-8B with its adapters off, prompted as each variant.
The gold traces resolve 23 of the 25 held-out games (92.0\%).
We test transfer on the 20 Avalon-NLU games of \citet{Stepputtis+23:AvalonLLM} (82 rounds; Appendix~\ref{app:avalon-nlu}), played by six humans as two Loyal Servants, Merlin, Percival, Morgana, and the Assassin.
The Investigator is the first Loyal Servant, and we evaluate two prompts (Appendix~\ref{app:nlu-prompts}).

\noindent\textbf{Results.}
On \fsc{Avcorp}, every trained configuration exceeds its untrained baseline in F1, and all except two tactic configurations, which tie at 48.0\%, also exceed it in exact match (Table~\ref{tab:detector}).
The tactic variant with the revised rewards performs best: it resolves 18 of the 25 games, as many as Claude Opus~5.5 and more than GPT-5.6 Sol (16), Gemini~3.1 Pro (14), and DeepSeek-V4-Pro (12) (Appendix~\ref{app:frontier}).
On Avalon-NLU, every trained detector exceeds the untrained model in round F1, mainly through higher recall (Appendix~\ref{app:detector-rounds}).
Training raises game-ending F1 in seven of the eight configurations, but the exact match ranges from 20.0\% to 46.7\%, against 40.0\% (trace prompt) and 20.0\% (tactic prompt) without training.
Training on \fsc{Avcorp} thus improves inference round by round on human games, whereas the final verdict depends on the training method and the prompt.
Future work includes testing cross-setting transfer (e.g., \fsl{Werewolf}) and auditing the detection abilities of existing safety-trained models.



\section{Related Work}
\label{sec:related-works}

Social deduction games enforce information asymmetry, making them natural testbeds for studying deception.
\fsl{Avalon} studies evaluate LLM agents \citep{Light+23:LLM-Avalon}, annotate persuasion and deception strategies in human dialogue \citep{Stepputtis+23:AvalonLLM}, capture human reasoning traces \citep{Li+25:InMind}, and raise win rates through recursive contemplation \citep{Wang+24:LLM-recursive-deception} or tree search \citep{Light+25:Strategist}.
\fsl{Werewolf} studies examine role-concealment communication \citep{Xu+23:WerewolfComm}, multimodal persuasion \citep{Lai+23:WerewolfPersuasion}, reasoning modules \citep{Wu+24:WerewolfEnhance}, and deception benchmarking \citep{Agarwal+25:WOLF}.
These resources provide strategy labels or reasoning traces, but they do not pair tactic labels of utterances with ToM reasoning traces for training detectors.

Deception emerges in AI systems as a strategy to satisfy goals \citep{Park+24:AI-Deception}, and LLMs lack robust defenses against manipulative contextual cues \citep{Huang+25:DeceptionBench}.
\citet{Sarkadi+19:Deception-ToM} combine IMT2 and IDT with ToM in a formal model of deceptive agents, but provide no corpus.
\fsc{Avcorp} builds on these accounts with a tactic taxonomy and nested ToM traces designed to train detectors.

\section*{Ethical Considerations}
We design \fsc{Avcorp} to facilitate deception detection.
The taxonomy uses \emph{published} deception theory (IMT2, IDT, TDT, and philosophical accounts of lying).
Releasing the taxonomy adds no new capability to deceive.
Every reasoning trace takes the Investigator's perspective: it models suspicion and updates beliefs about the other players.
No trace takes an Evil player's perspective or plans a deception.
Therefore, the traces do not teach how to deceive.
The tactic-labeled utterances do show deceptive moves in context, and a model could learn to imitate them.
We release them with the detection task and the taxonomy so that safety researchers can audit which patterns a detector must cover.

\section*{Limitations}
We identify limitations that concern the corpus design, detector training, evaluation scope, and verification judges, and describe a resolution for each.
\paragraph{Distributional artifacts.}
GPT-5.2 selects tactics for Rounds~2--5 based on the game context.
This selection skews the tactic frequencies (Table~\ref{tab:tactic-count} in Appendix~\ref{app:s3}).
The most frequent tactic, \emph{Evidence sharing}, labels 17.5\% of the utterances.
Further, \fsc{Avcorp} fixes quest outcomes to balance Evil and Good wins (54\% and 46\%).
A detector must exceed these two majority baselines to show that it learns.
Additionally, the corpus contains no rejected proposals, which can provide evidence about player roles.
Future versions can use frequency-aware sampling and allow rejected proposals to reduce these artifacts.



\paragraph{Synthetic dialogue realism.}
An LLM generates every utterance, and the dialogue may lack the hesitations, topic shifts, and emotional cues of human deception.
Researchers can replace the synthetic text with crowdsourced human utterances based on the existing tactic assignments.
They must then regenerate the reasoning traces (Stages~5 and~6) to analyze the new statements.


\paragraph{Scale and language scope.}
The current corpus contains 5,000 utterances from an English language game.
The six-stage pipeline directly supports scaling to more games, but token cost limited its size.
We release the pipeline and tactic taxonomy so that the community can scale the corpus to more games and languages (Chinese Avalon gameplay \citep{Li+25:InMind}), including settings such as \fsl{Werewolf} and \fsl{Among Us} \citep{Sarkar+25:AmongUs}, where the same deception taxonomy applies.


\paragraph{Untested transfer beyond the game.}
We train and test the detector only on \fsl{Avalon} using synthetic gameplay (\fsc{Avcorp}) and human players (Avalon-NLU).
We have not evaluated transfer to deception outside the game.
For example, DeceptionBench \citep{Huang+25:DeceptionBench} tests deception in domains like healthcare and the economy.
Evaluating on such benchmarks requires a new task format because their data lacks hidden roles and our tactic labels.

\paragraph{Training methods.}
GRPO does not consistently improve on SFT.
With the original rewards, GRPO underperforms SFT for exact matches on \fsc{Avcorp}.
The exact match rate drops from 36.0\% to 33.3\% for the trace variant and 52.0\% to 48.0\% for the tactic variant.
On Avalon-NLU, SFT yields more exact matches than GRPO on average.
The deduction F1 reward (Appendix~\ref{app:rewards}) gives partial credit to a deduction that names extra players.
With the original rewards, GRPO names more players per round than SFT on Avalon-NLU (Table~\ref{tab:nlu-rounds} in Appendix~\ref{app:detector-rounds}). 
In contrast, SFT imitates the gold traces, whose game-ending deductions always name two players.
A reward that scores the deduction size might close this gap, but we have not evaluated this approach.

\paragraph{Gold prior traces during training.}
During training, each prompt includes the gold traces from the two preceding rounds, whereas during evaluation, the detector receives its own earlier traces.
Gold traces contain previous gold deductions.
Copying the previous gold deduction alone resolves 18 of the 25 held-out games (Appendix~\ref{app:gold-chain}).
The detector can therefore learn to depend on information that it does not receive during evaluation, and an error in an early round can cascade into later rounds.
To fix this, future training can build prompts from the detector's own earlier traces or scheduled sampling \citep{Mihaylova+Martins19:ScheduledSampling} can progressively replace gold traces with generated traces.

\paragraph{Verification judges.}
Both verification stages use Claude Sonnet~4.5 and~4.6.
The Stage~6 judge always receives the ReAct trace as Trace~A.
We did not swap the trace order or evaluate other model families.
Additionally, rerunning the pipeline with future models may yield different outcomes.
Nevertheless, the blind human study (Section~\ref{sec:iaa}) balances the trace order and agrees with the judge ($\kappa$ from .92 to .94).
We publish all prompts and scripts to enable the community to recalibrate with successor models.









\bibliography{Fardin, delta}

\appendix
\onecolumn
\section*{Appendix}

\section{\fsc{Avcorp} Design}
\label{app:design}
Table~\ref{tab:team-size} records the quest team size for each round in \fsl{Avalon}, which determines how many players participate in the quest phase.

\begin{table}[ht]
\centering
\caption{Team size distribution across \fsl{Avalon} rounds.}
\begin{tabular}{lccccc}
\toprule
\textbf{\textcolor{ncsuDarkPurple}{Round}} & 
\textbf{\textcolor{ncsuDarkPurple}{R1}} & 
\textbf{\textcolor{ncsuDarkPurple}{R2}} & 
\textbf{\textcolor{ncsuDarkPurple}{R3}} & 
\textbf{\textcolor{ncsuDarkPurple}{R4}} & 
\textbf{\textcolor{ncsuDarkPurple}{R5}} \\
\midrule
\rowcolor{mint!8}
\textbf{Team size} & 2 & 3 & 2 & 3 & 3 \\
\bottomrule
\end{tabular}
\label{tab:team-size}
\end{table}


\FloatBarrier

\section{Behavior Matrix Definitions}
\label{app:tactics}

The behavior matrix $B$ organizes all 37 tactics across two axes: \textbf{information commitment strategy} and \textbf{social goal orientation}. The axes are defined below, followed by full definitions of each tactic. Alignment accessibility is noted as \textbf{Good}, \textbf{Evil}, or \textbf{Both}.

Information commitment strategy is derived from IMT2 and Phl, this axis distinguishes how utterances relate to the known truth:
\begin{itemize}[nosep]
  \item \fbf{Transparent:} Openly sharing clear, evidence-based information.
  \item \fbf{Framing:} Deliberate shaping of discourse through omission or emphasis.
  \item \fbf{Careless-to-truth:} Assertions made without regard for factual accuracy.
  \item \fbf{Counterfactual:} Direct assertions of information contrary to reality.
\end{itemize}

Social goal orientation is derived from IDT and TDT, this axis characterizes the social goal an utterance serves:
\begin{itemize}[nosep]
  \item \fbf{Cooperative:} Acting in harmony with group goals and mutual benefit.
  \item \fbf{Defensive:} Protecting one's own image, reputation, or position.
  \item \fbf{Opportunistic:} Exploiting situations for personal gain at the expense of others.
  \item \fbf{Adversarial:} Seeking to undermine, harm, or sabotage others.
\end{itemize}

\subsection*{(A) Transparent}

\noindent\textbf{Cooperative: Good.}
\begin{itemize}[nosep, leftmargin=1.5em]
  \item \textbf{Evidence sharing}: Providing concrete facts to support a claim.
  \item \textbf{Rational justification}: Explaining actions through logical reasons.
  \item \textbf{Perspective-taking}: Acknowledging another's viewpoint to build trust.
\end{itemize}

\noindent\textbf{Defensive: Both.}
\begin{itemize}[nosep, leftmargin=1.5em]
  \item \textbf{Defensive disclosure}: Revealing only what is necessary to protect oneself.
  \item \textbf{Self-justification}: Excusing one's own actions to avoid blame.
\end{itemize}

\noindent\textbf{Opportunistic: Evil.}
\begin{itemize}[nosep, leftmargin=1.5em]
  \item \textbf{Weaponized candor}: Using blunt truth strategically to gain advantage over others.
  \item \textbf{Overstated contribution}: Exaggerating one's effort to appear indispensable.
\end{itemize}

\noindent\textbf{Adversarial: Evil.}
\begin{itemize}[nosep, leftmargin=1.5em]
  \item \textbf{Truthful misdirection}: Offering accurate details that nonetheless distract from the relevant truth.
  \item \textbf{Misleading emphasis}: Highlighting true but minor details to make them appear central.
\end{itemize}

\subsection*{(B) Framing}

\noindent\textbf{Cooperative: Good.}
\begin{itemize}[nosep, leftmargin=1.5em]
  \item \textbf{Bayesian hedging}: Presenting probabilities or uncertainty to remain flexible.
  \item \textbf{Strategic uncertainty}: Deliberately leaving outcomes ambiguous to avoid overcommitment.
  \item \textbf{Pragmatic silence}: Omitting information instead of lying outright.
\end{itemize}

\noindent\textbf{Defensive: Both.}
\begin{itemize}[nosep, leftmargin=1.5em]
  \item \textbf{Omission to save face}: Withholding details to avoid personal embarrassment.
  \item \textbf{Half-truths}: Mixing true information with key omissions to create a false impression.
\end{itemize}

\noindent\textbf{Opportunistic: Evil.}
\begin{itemize}[nosep, leftmargin=1.5em]
  \item \textbf{Cherry-picking}: Selectively presenting only favorable evidence.
  \item \textbf{False consensus}: Overstating agreement among others to pressure a target.
  \item \textbf{Misframing}: Presenting accurate information within a skewed or misleading context.
\end{itemize}

\noindent\textbf{Adversarial: Evil.}
\begin{itemize}[nosep, leftmargin=1.5em]
  \item \textbf{Withholding info}: Creating strategic doubt by hiding critical details.
  \item \textbf{Strategic omission}: Structuring a true account so that omitted facts lead to a false conclusion.
\end{itemize}

\subsection*{(C) Careless-to-truth}

\noindent\textbf{Cooperative: Good.}
\begin{itemize}[nosep, leftmargin=1.5em]
  \item \textbf{Noisy sincerity}: Being honest in a confusing or informationally overloaded way.
  \item \textbf{Over-explaining}: Adding excessive detail in an attempt to appear genuine.
\end{itemize}

\noindent\textbf{Defensive: Both.}
\begin{itemize}[nosep, leftmargin=1.5em]
  \item \textbf{Face-saving blather}: Using empty talk to avoid confrontation or accountability.
  \item \textbf{Vague filler}: Relying on imprecise or generic phrases that commit to nothing.
\end{itemize}

\noindent\textbf{Opportunistic: Evil.}
\begin{itemize}[nosep, leftmargin=1.5em]
  \item \textbf{Bullshitting}: Speaking without concern for truth or falsity, purely to produce an effect.
  \item \textbf{Technobabble}: Using jargon or complexity to appear credible without substantive content.
\end{itemize}

\noindent\textbf{Adversarial: Evil.}
\begin{itemize}[nosep, leftmargin=1.5em]
  \item \textbf{Obfuscatory nonsense}: Deploying intentionally incoherent speech to mislead or exhaust.
  \item \textbf{Confusion tactic}: Disrupting others through pivots or interruptions rather than incoherent content.
\end{itemize}

\subsection*{(D) Counterfactual}

\noindent\textbf{Cooperative: Good.}
\begin{itemize}[nosep, leftmargin=1.5em]
  \item \textbf{Honest mistake}: Asserting something false in genuine good faith.
  \item \textbf{Sincere false belief}: Expressing a persistently internalized false belief, not a momentary slip.
\end{itemize}

\noindent\textbf{Defensive: Both.}
\begin{itemize}[nosep, leftmargin=1.5em]
  \item \textbf{Deflection}: Shifting blame or suspicion onto another player by redirecting attention, often through partial truths or emotional appeals.
  \item \textbf{Self-deception}: Convincing oneself of a falsehood and acting on it sincerely.
  \item \textbf{Rationalized error}: Justifying a factual mistake with faulty or motivated logic.
\end{itemize}

\noindent\textbf{Opportunistic: Evil.}
\begin{itemize}[nosep, leftmargin=1.5em]
  \item \textbf{Reverse accusations}: Shifting blame back onto the accuser to neutralize suspicion.
  \item \textbf{Strategic fabrication}: Inventing plausible details to achieve a deceptive goal.
\end{itemize}

\noindent\textbf{Adversarial: Evil.}
\begin{itemize}[nosep, leftmargin=1.5em]
  \item \textbf{Hard lying}: Deliberately asserting the opposite of the known truth.
  \item \textbf{Gaslighting}: Causing others to doubt their own memory, perception, or reasoning.
  \item \textbf{Bold falsehoods}: Asserting lies so brazenly that delivery alone overwhelms scrutiny, regardless of subtlety.
\end{itemize}

\newpage
\section{Six Stages}
\label{app:six-stages}

\begin{figure}[!htb]
\centering
\includegraphics[width=0.99\linewidth]{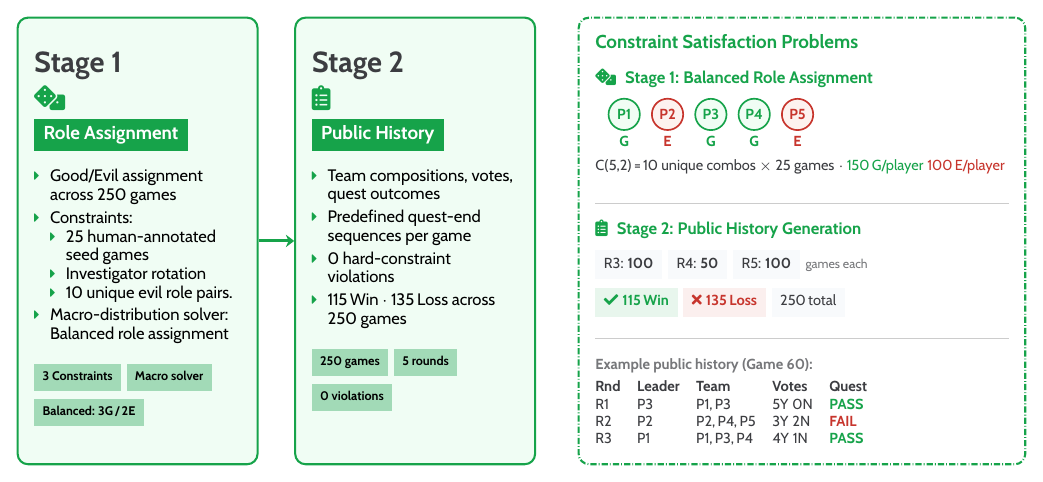}
\caption{Stages~1 and 2 facilitate in construction of $g$.}
\label{fig:s1s2}
\end{figure}

\begin{figure}[!htb]
\centering
\includegraphics[width=0.99\linewidth]{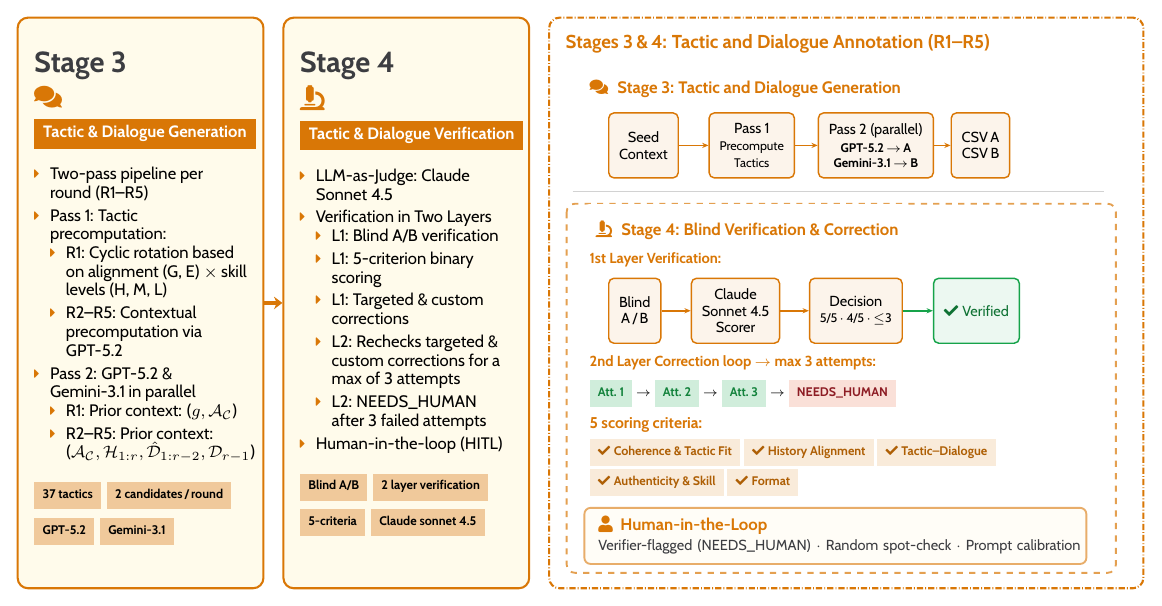}
\caption{Stages~3 and 4 facilitate in construction of  $\mathcal{A}_\mathcal{C}$ and $\mathcal{D}$.}
\label{fig:s3s4}
\end{figure}

\begin{figure}[!htb]
\centering
\includegraphics[width=0.99\linewidth]{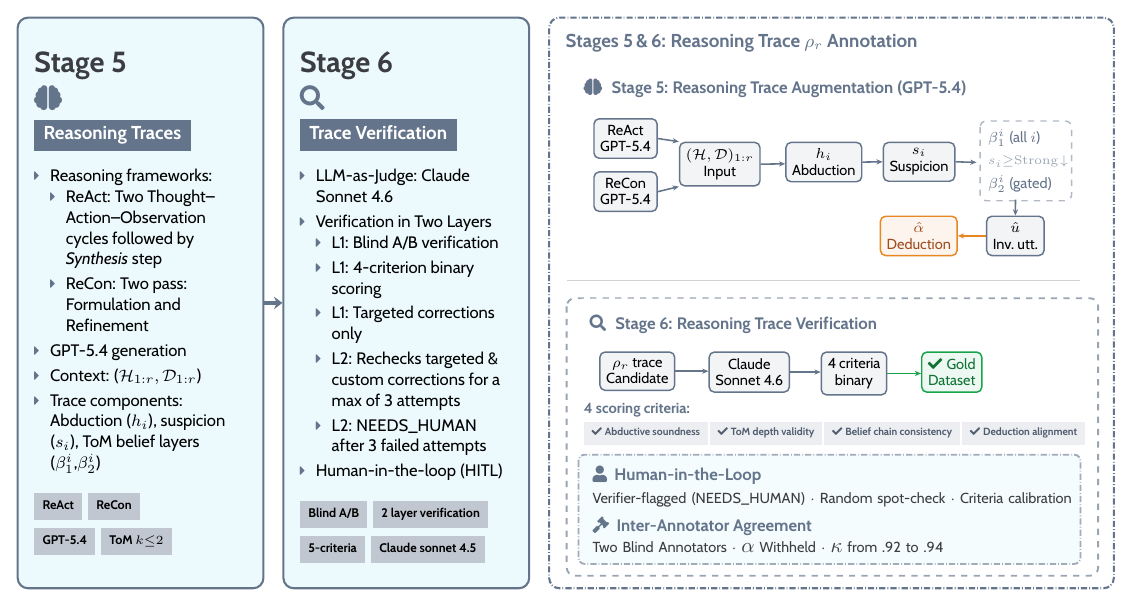}
\caption{Stages~5 and 6 facilitate in constructing $\rho$.}
\label{fig:s5s6}
\end{figure}

\FloatBarrier

\subsection{Stage 1}
\label{app:s1}


\begin{figure}[!htb]
\subcaptionsetup{width=0.9\linewidth}
\centering
\begin{subfigure}[htb]{0.49\textwidth}
\centering
\includegraphics[width=0.8\linewidth]{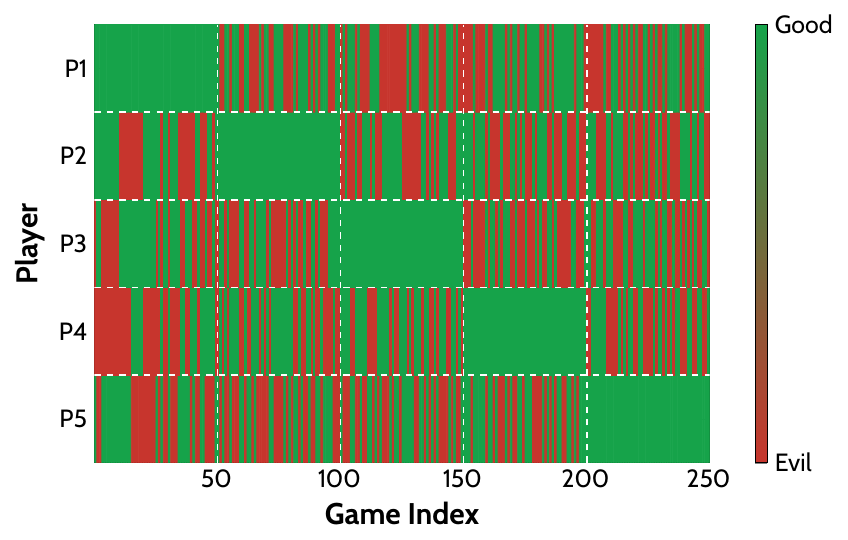}
\caption{Role assignment heatmap ($5$ players $\times$ $250$ games). The Good assignment staircase reflects the diagonal block constraint.}
\label{fig:heatmap}
\end{subfigure}
\hfill
\begin{subfigure}[htb]{0.49\textwidth}
\centering
\includegraphics[width=0.7\linewidth]{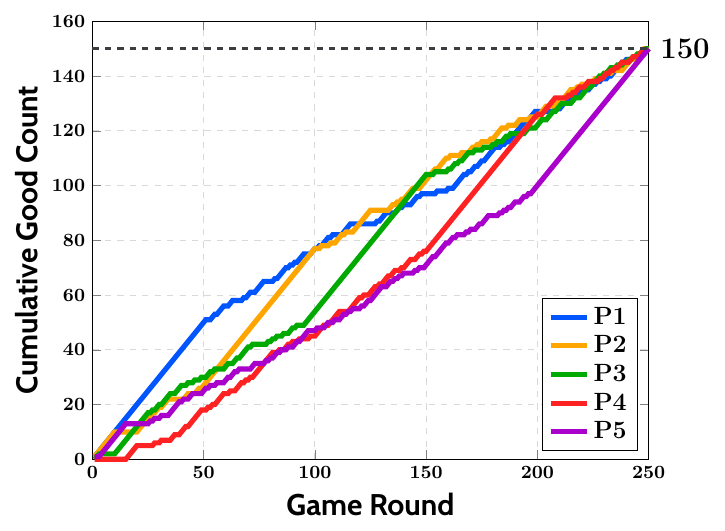}
\caption{Distribution of 150 Good assignments for each player across the 250 games.}
\label{fig:good-distribution}
\end{subfigure}
\caption{Stage 1 Good and Evil Role Distribution.}
\end{figure}


\FloatBarrier

\subsection{Stage 2}
\label{app:s2}
\begin{table}[!htb]
\centering
\caption{The 20 predefined outcome sequences grouped by terminal round and winning team.
\textbf{P} = quest pass (Good wins the quest); \textbf{F} = quest fail (Evil sabotages).
Each sequence terminates on the round delivering the third accumulated win.
Within R5, each row is a matched pair sharing identical first-four outcomes---only the fifth quest determines the winner.}
\label{tab:outcome-sequences}
\begin{tabular}{cll}
\toprule
\textbf{Terminal round} & \textbf{Good wins} & \textbf{Evil wins} \\
\midrule
R3 & \texttt{P-P-P} & \texttt{F-F-F} \\
\midrule
\multirow{3}{*}{R4}
  & \texttt{P-P-F-P} & \texttt{F-F-P-F} \\
  & \texttt{P-F-P-P} & \texttt{F-P-F-F} \\
  & \texttt{F-P-P-P} & \texttt{P-F-F-F} \\
\midrule
\multirow{6}{*}{R5}
  & \texttt{F-F-P-P-P} & \texttt{F-F-P-P-F} \\
  & \texttt{F-P-F-P-P} & \texttt{F-P-F-P-F} \\
  & \texttt{F-P-P-F-P} & \texttt{F-P-P-F-F} \\
  & \texttt{P-F-F-P-P} & \texttt{P-F-F-P-F} \\
  & \texttt{P-F-P-F-P} & \texttt{P-F-P-F-F} \\
  & \texttt{P-P-F-F-P} & \texttt{P-P-F-F-F} \\
\midrule
\textbf{Total} & \textbf{10} & \textbf{10} \\
\bottomrule
\end{tabular}
\end{table}

\begin{figure}[!htb]
\centering
\includegraphics[width=0.6\linewidth]{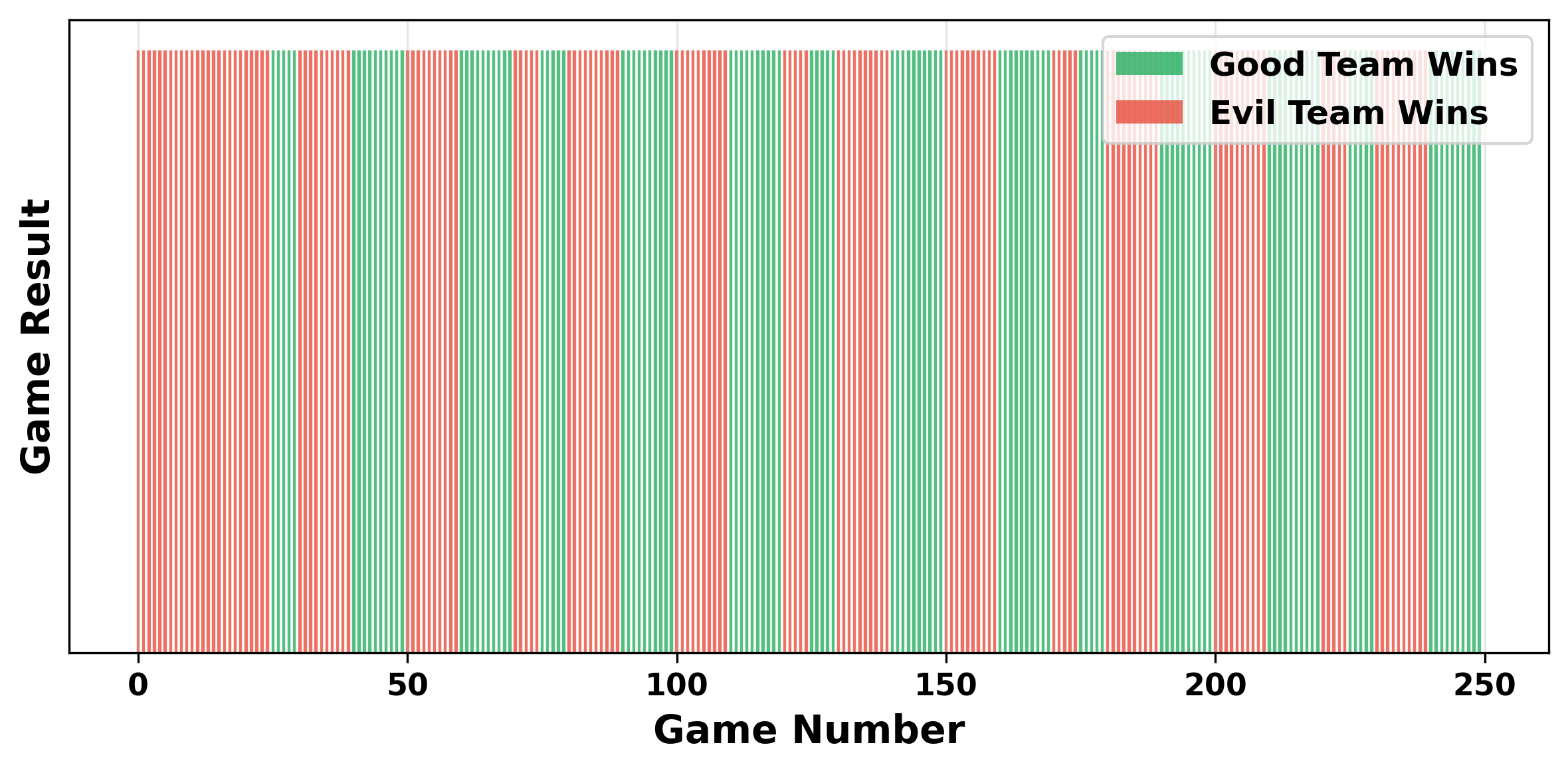}
\caption{Game outcome across all 250 games (green = Good victory, red = Evil victory). The repeating block structure reflects the predefined assignment of the outcome sequence.}
\label{fig:game-timeline}
\end{figure}

\begin{figure}[!htb]
\centering
\includegraphics[width=0.5\linewidth]{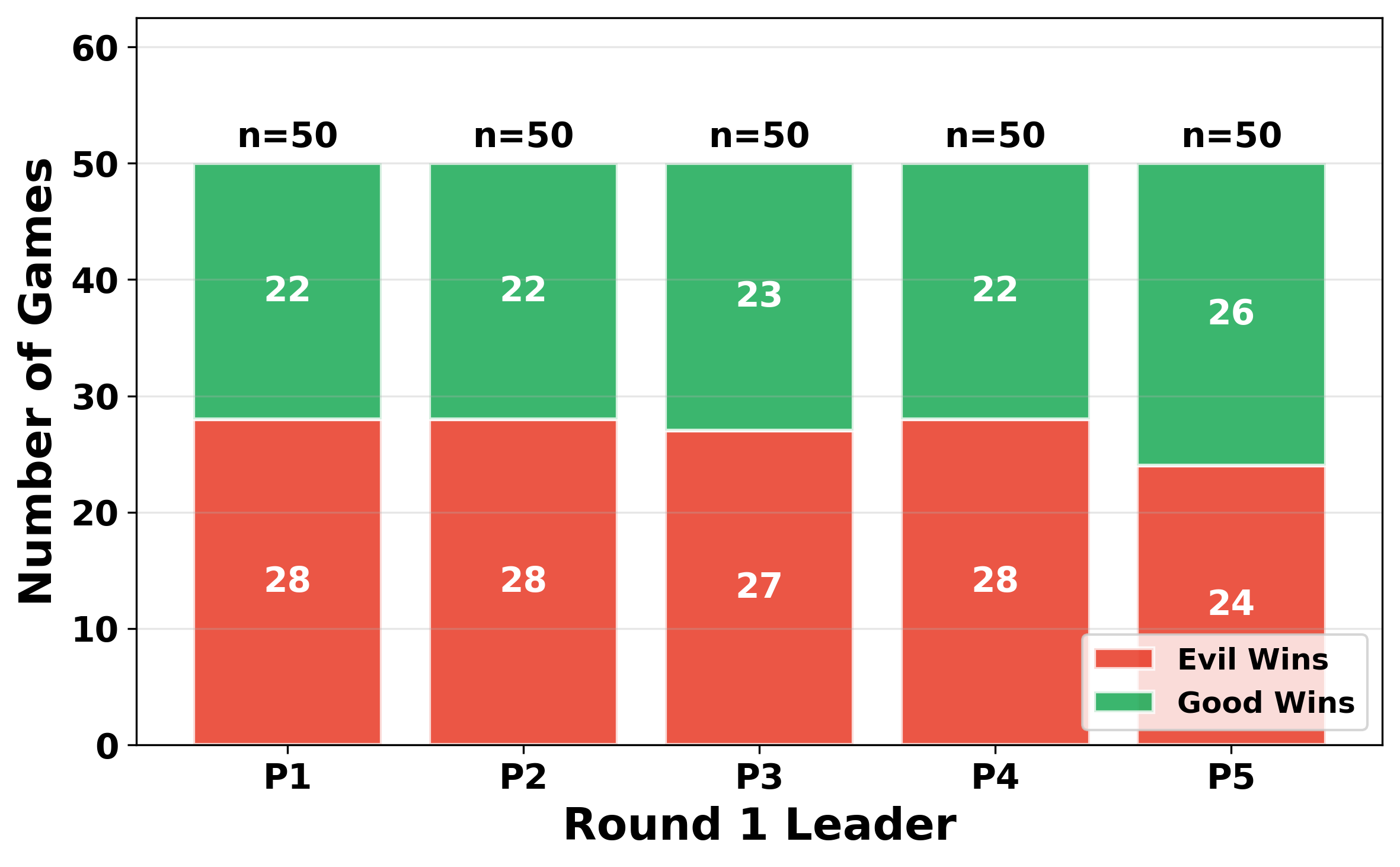}
\caption{Round~1 leader distribution across 250 games, split by game outcome (green = Good victory, red = Evil victory). Each player serves as Round~1 leader in exactly 50 games, confirming the balanced assignment; the outcome split reveals no advantage from leading first.}
\label{fig:leader-dist}
\end{figure}


\begin{figure}[!htb]
\centering
\includegraphics[width=0.55\linewidth]{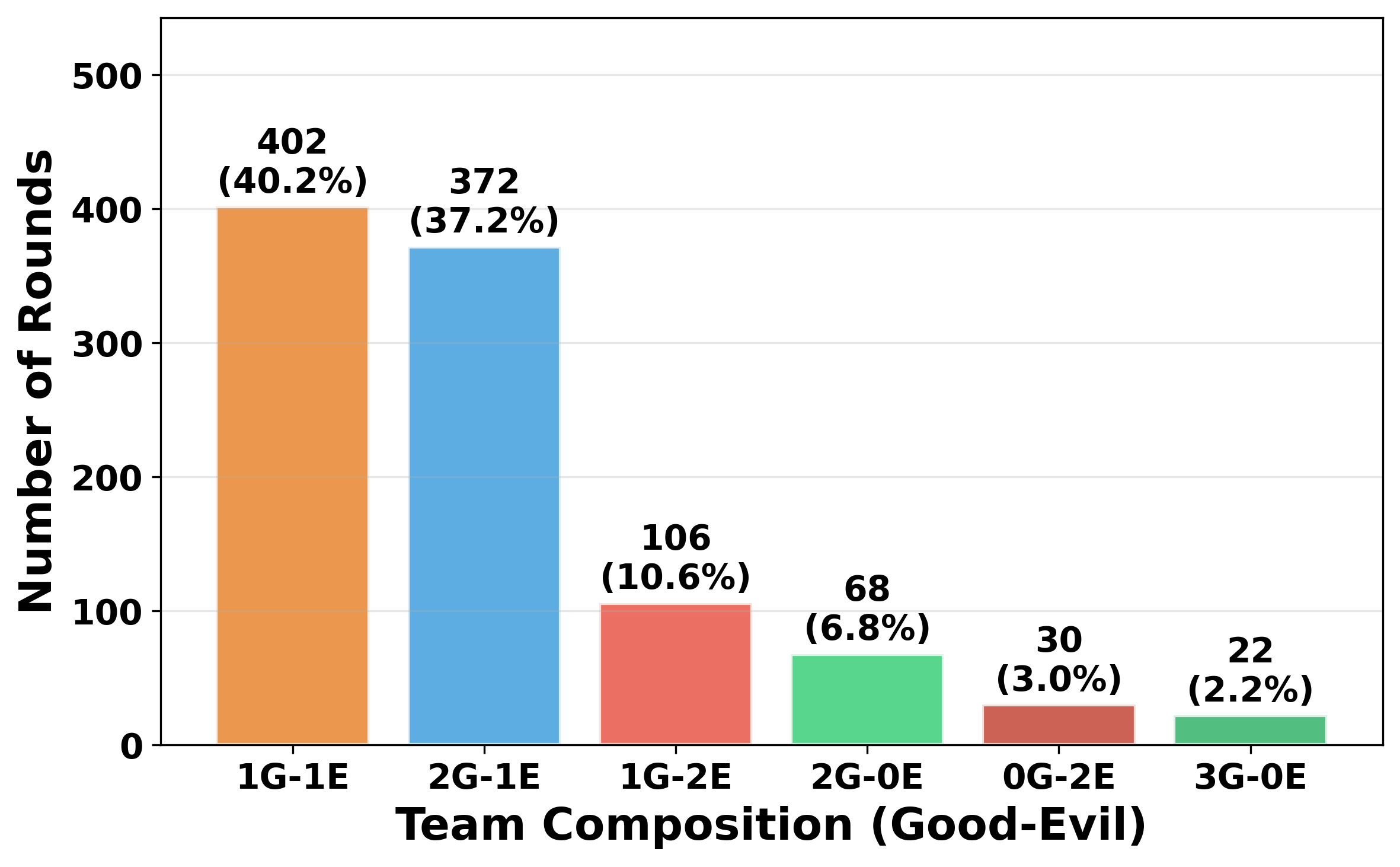}
\caption{Team composition distribution across 1,000 rounds. Labels denote the exact count of Good (G) and Evil (E) players on each team. Mixed teams containing exactly one Evil player (1G-1E and 2G-1E) account for \qtyP{77.4} of all rounds, reflecting the role-aware selection rule.}
\label{fig:team-composition}
\end{figure}

\begin{figure}[!htb]
\centering
\includegraphics[width=0.55\linewidth]{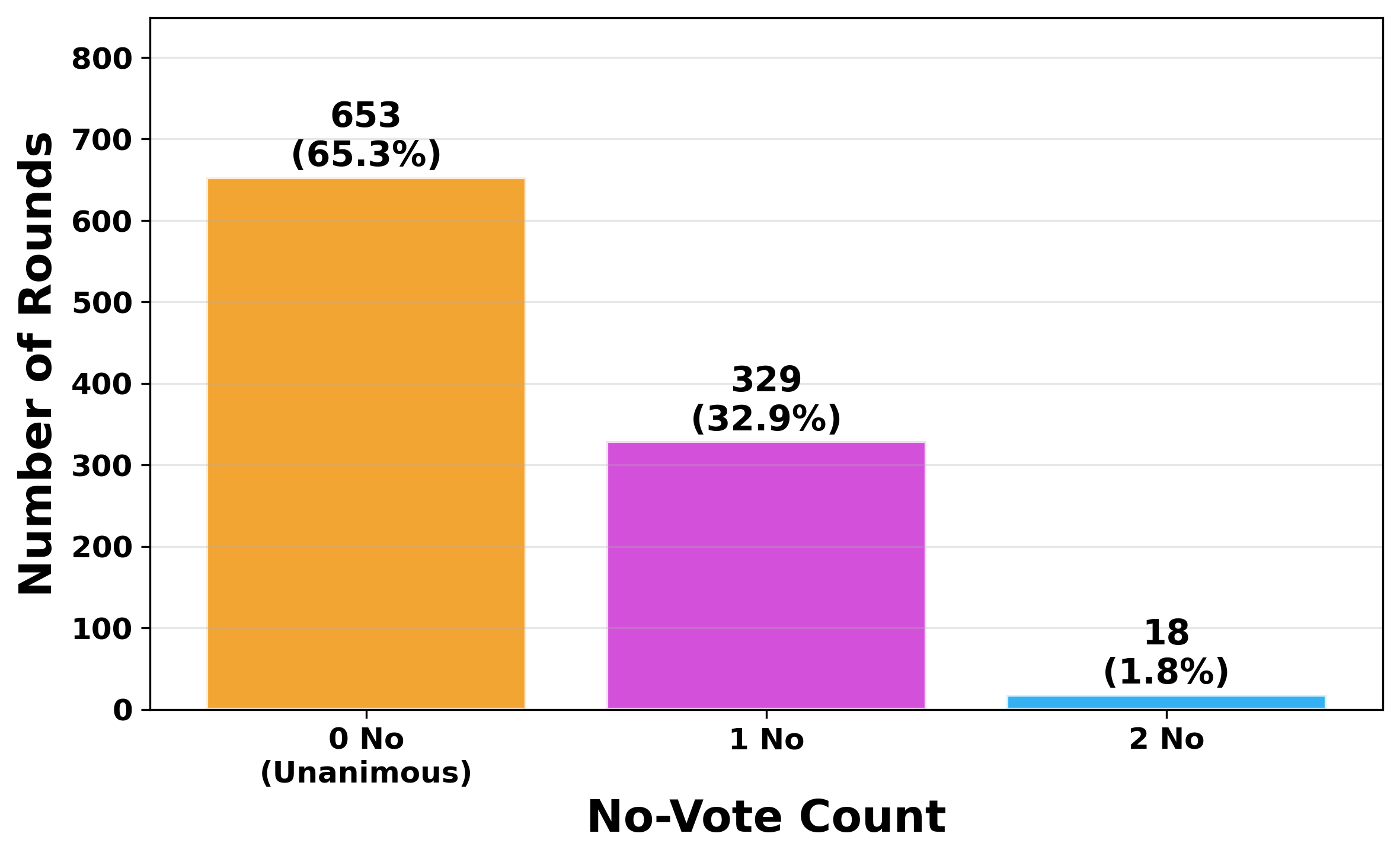}
\caption{Voting pattern distribution across 1,000 rounds. Most proposals pass unanimously (\qtyP{65.3}); a single dissenting vote occurs in \qtyP{32.9} of rounds, and the minimum two-No threshold is reached in only \qtyP{1.8}.}
\label{fig:voting-pattern}
\end{figure}

\FloatBarrier

\subsection{Stage 3}
\label{app:s3}

\begin{figure}[!htb]
  \centering
  \includegraphics[width=0.5\linewidth]{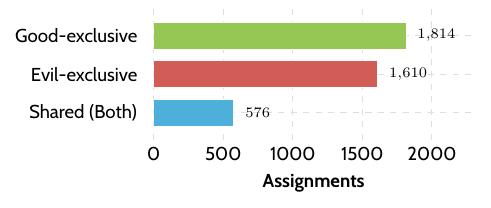}
  \caption{Tactic assignment distribution by alignment category ($n = 4{,}000$ assignments across 1,000 rounds).}
  \label{fig:tactic-dist}
\end{figure}

Figure~\ref{fig:tactic-dist} reports tactic assignment frequencies by alignment category across all 4,000 agent assignments and 
Good-exclusive tactics (Cooperative column) account for 1,814 assignments (45.4\%), Evil-exclusive tactics (Opportunistic and Adversarial columns) for 1,610 (40.3\%), and Shared tactics (Defensive/Both column) for 576 (14.4\%).
The lower Shared share reflects the level-aware assignment rule, since high-skill agents dominate numerically and draw exclusively from alignment-specific columns.

\begin{figure}[!htb]
  \centering
  \includegraphics[width=0.5\linewidth]{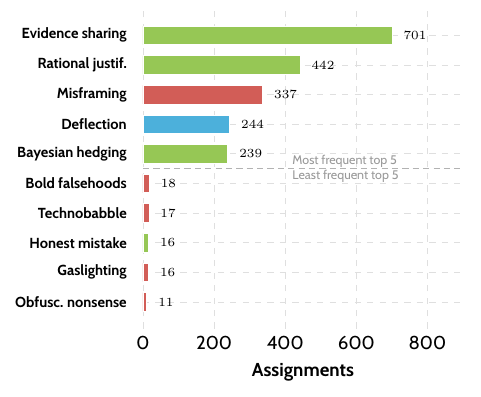}
  \caption{The five most and least frequently assigned tactics across all 1,000 samples.}
  \label{fig:tactic-freq}
\end{figure}

Table~\ref{tab:tactic-count} reports the assignment count for each of the 37 tactics, organized by the same 4$\times$4 matrix structure used in Section~\ref{sec:behavior-matrix}.
Figure~\ref{fig:tactic-freq} illustrates the five most and least occurring tactics.

\begin{table*}[htb]
\centering
\resizebox{\textwidth}{!}{
\begin{tabular}{l l l l l}
\toprule
\textbf{Information Commitment $\backslash$ Social Goal} &
\textbf{(1) Cooperative} & \textbf{(2) Defensive} & \textbf{(3) Opportunistic} & \textbf{(4) Adversarial} \\
\midrule
(A) Transparent &
\cellcolor{green!20}\begin{tabular}[c]{@{}l@{}}Evidence sharing (701)\\ Rational justification (442)\\ Perspective-taking (109)\end{tabular} &
\cellcolor{blue!20}\begin{tabular}[c]{@{}l@{}}Defensive disclosure (73)\\ Self-justification (64)\end{tabular} &
\cellcolor{red!20}\begin{tabular}[c]{@{}l@{}}Weaponized candor (87)\\ Overstated contribution (35)\end{tabular} &
\cellcolor{red!20}\begin{tabular}[c]{@{}l@{}}Truthful misdirection (188)\\ Misleading emphasis (205)\end{tabular} \\
\midrule
(B) Framing &
\cellcolor{green!20}\begin{tabular}[c]{@{}l@{}}Bayesian hedging (239)\\ Strategic uncertainty (47)\\ Pragmatic silence (30)\end{tabular} &
\cellcolor{blue!20}\begin{tabular}[c]{@{}l@{}}Omission to save face (26)\\ Half-truths (48)\end{tabular} &
\cellcolor{red!20}\begin{tabular}[c]{@{}l@{}}Cherry-picking (211)\\ False consensus (198)\\ Misframing (337)\end{tabular} &
\cellcolor{red!20}\begin{tabular}[c]{@{}l@{}}Withholding info (WIFOM) (59)\\ Strategic omission (101)\end{tabular} \\
\midrule
(C) Careless-to-truth &
\cellcolor{green!20}\begin{tabular}[c]{@{}l@{}}Noisy sincerity (57)\\ Over-explaining (146)\end{tabular} &
\cellcolor{blue!20}\begin{tabular}[c]{@{}l@{}}Face-saving blather (46)\\ Vague filler (32)\end{tabular} &
\cellcolor{red!20}\begin{tabular}[c]{@{}l@{}}Bullshitting (21)\\ Technobabble (17)\end{tabular} &
\cellcolor{red!20}\begin{tabular}[c]{@{}l@{}}Obfuscatory nonsense (11)\\ Confusion tactics (18)\end{tabular} \\
\midrule
(D) Counterfactual &
\cellcolor{green!20}\begin{tabular}[c]{@{}l@{}}Honest mistake (16)\\ Sincere false belief (27)\end{tabular} &
\cellcolor{blue!20}\begin{tabular}[c]{@{}l@{}}Deflection (244)\\ Self-deception (21)\\ Rationalized error (22)\end{tabular} &
\cellcolor{red!20}\begin{tabular}[c]{@{}l@{}}Reverse accusations (47)\\ Strategic fabrication (19)\end{tabular} &
\cellcolor{red!20}\begin{tabular}[c]{@{}l@{}}Hard lying (22)\\ Gaslighting (16)\\ Bold falsehoods (18)\end{tabular} \\
\bottomrule
\end{tabular}
}
\caption{Per-tactic assignment counts ($n = 4{,}000$).
Each entry shows the tactic name and its total count across all 1,000 samples.
Column totals confirm the alignment-category breakdown reported in Figure~\ref{fig:tactic-dist}: Cooperative (Good-exclusive) = 1,814; Defensive (Shared) = 576; Opportunistic + Adversarial (Evil-exclusive) = 1,610.
The total of 4,000 is verified.}
\label{tab:tactic-count}
\end{table*}

\noindent Prompt Boxes~\ref{box:s3-pass1} and~\ref{box:s3-pass2} show the structure of the Pass~1 and Pass~2 prompts for a Round~4 example.
Content common to all games is abbreviated with $[\ldots]$.

\begin{tcolorbox}[
  colback=bamber,
  colframe=camber,
  coltitle=black,
  fonttitle=\sffamily\small\bfseries,
  title={Prompt Box~\thepromptbox. Pass~1 Tactic Precompute (Round~4 Example)},
  step and label={promptbox}{box:s3-pass1},
  breakable,
  top=4pt, bottom=4pt, left=6pt, right=6pt,
]
\ttfamily\small
\textbf{System:} You are an expert in social deduction game strategy and deception theory.\\[3pt]
\textbf{User:}\\
You are an expert in social deduction game strategy and deception theory.
Your task: Pre-assign the communication tactic for each speaking player in Round 4 of an Avalon game.
Base your choices on: (1) each player's Round 3 tactic as a seed anchor, (2) the current game state.\\[3pt]
\# GAME CONTEXT\\
\textbf{Game ID:} G042 \quad
\textbf{Assigning tactics for:} Round 4\\
\textbf{Silent protagonist (observer):} P3 (Good)\\
\textbf{Player Roles:} P1: Evil, P2: Good, P3: Good (protagonist), P4: Evil, P5: Good\\[3pt]
\# CUMULATIVE PUBLIC HISTORY (Rounds 1 through 4)\\
Format: Round | Leader | Team | Votes | Quest Outcome\\
Round 1 | Leader: P1 | Team: P1, P2, P3 | Votes: P1:Y P2:Y P3:Y P4:N P5:Y | Quest 1 Outcome: PASS\\
Round 2 | Leader: P2 | Team: P2, P4, P5 | Votes: P1:Y P2:Y P3:N P4:Y P5:Y | Quest 2 Outcome: FAIL\\
Round 3 | Leader: P3 | Team: P1, P2, P3 | Votes: P1:Y P2:Y P3:Y P4:Y P5:Y | Quest 3 Outcome: PASS\\
Round 4 | Leader: P4 | Team: P2, P4, P5 | Votes: P1:Y P2:Y P3:N P4:Y P5:Y | Quest 4 Outcome: [...]\\[3pt]
\# OLDER ROUND SUMMARIES (Rounds 1 through 2)\\
These summaries cover rounds before Round~3. Use them for broader game trajectory and behavioral arc context.\\
$[\ldots\text{ condensed narrative summaries of Rounds 1--2} \ldots]$\\[3pt]
\# ROUND 3 DISCUSSION (Reference for player dialogues and dynamics)\\
$[\ldots\text{ full verbatim discussion log from Round 3} \ldots]$\\[3pt]
\# ROUND 3 TACTIC SEEDS (Anchor points for Round 4 evolution)\\
~~P1 (Evil, Mach-IV High): `Strategic Deception' [Framing $\times$ Opportunistic]\\
~~P2 (Good, GRS Moderate): `Tentative Reasoning' [Framing $\times$ Cooperative]\\
~~P4 (Evil, Mach-IV Moderate): `Deflection' [Framing $\times$ Defensive]\\
~~P5 (Good, GRS High): `Honest Disclosure' [Transparent $\times$ Cooperative]\\[3pt]
\# ALIGNMENT POOLS\\
~~P1 (Evil): must select from Evil pool (27 tactics)\\
~~P2 (Good): must select from Good pool (19 tactics)\\
~~P4 (Evil): must select from Evil pool (27 tactics)\\
~~P5 (Good): must select from Good pool (19 tactics)\\[3pt]
\# TACTIC EVOLUTION GUIDELINES\\
Scale levels are FIXED. $[\ldots]$ Only the tactic may change.\\
Round 4 (Mid): mix Opportunistic and Defensive --- begin shaping the narrative more actively.\\
$[\ldots\text{ Good and Evil player guidance} \ldots]$\\[3pt]
\textbf{Output:} JSON object: \{"P1": \{"tactic": "...", "row": "...", "col": "...", "scale": "Mach-IV", "level": "High"\}, ...\}
\end{tcolorbox}

\begin{tcolorbox}[
  colback=bamber,
  colframe=camber,
  coltitle=black,
  fonttitle=\sffamily\small\bfseries,
  title={Prompt Box~\thepromptbox. Pass~2 Dialogue Generation (Round~4 Example)},
  step and label={promptbox}{box:s3-pass2},
  breakable,
  top=4pt, bottom=4pt, left=6pt, right=6pt,
]
\ttfamily\small
\textbf{System:} You are an expert in social deception games and deception theory.
You generate realistic game discussions with tactical communication patterns.\\[3pt]
\textbf{User:}\\
Generate a realistic discussion log for Round 4 of an Avalon game based on the provided game context.\\[3pt]
\# AVALON RULES AND PUBLIC HISTORY FORMAT\\
$[\ldots\text{ game mechanics: team selection, approval votes, anonymous quest outcome} \ldots]$\\[3pt]
\# 4$\times$4 BEHAVIOR MATRIX\\
$[\ldots\text{ row/column definitions: Transparent, Framing, Careless-to-truth, Counterfactual} \times \text{Cooperative, Defensive, Opportunistic, Adversarial} \ldots]$\\[3pt]
\# SCALE DEFINITIONS\\
GRS (Good players): High = evidence-based, Moderate = some hedging, Low = confused or overwhelmed.\\
Mach-IV (Evil players): High = subtle manipulation, Moderate = situational, Low = impulsive.\\[3pt]
\# CUMULATIVE PUBLIC HISTORY (Rounds 1 through 4)\\
$[\ldots\text{ same public history block as Pass~1} \ldots]$\\[3pt]
\# OLDER ROUND SUMMARIES (Rounds 1 through 2)\\
$[\ldots\text{ condensed narrative summaries of Rounds 1--2} \ldots]$\\[3pt]
\# ROUND 3 DISCUSSION\\
Use this as context for how each player communicates. Maintain character voice consistency.\\
$[\ldots\text{ full verbatim discussion log from Round 3} \ldots]$\\[3pt]
\# PRE-ASSIGNED TACTICS FOR ROUND 4\\
Each speaker's tactic has been pre-assigned based on their role, scale, and current game state.\\
~~P1 (Evil, Mach-IV High): `Aggressive Misdirection' [Counterfactual $\times$ Adversarial]\\
~~P2 (Good, GRS Moderate): `Tentative Reasoning' [Framing $\times$ Cooperative]\\
~~P4 (Evil, Mach-IV Moderate): `False Accusation' [Counterfactual $\times$ Opportunistic]\\
~~P5 (Good, GRS High): `Evidence Synthesis' [Transparent $\times$ Cooperative]\\[3pt]
\textbf{Output:} JSON with two fields.\\
\textbf{discussion\_log}: string starting with "Discussion after Quest 4:" followed by four P\#: "dialogue" lines (one blank line between speakers; exactly 4 speakers; speaking order agenda-driven, not P1--P4 default).\\
\textbf{matrix\_tactic\_scale}: JSON object mapping each speaker to their row, col, tactic, scale, and level.
\end{tcolorbox}

\FloatBarrier

\subsection{Stage 4}
\label{app:s4}

\begin{figure}[!htb]
\centering
\includegraphics[width=0.75\linewidth]{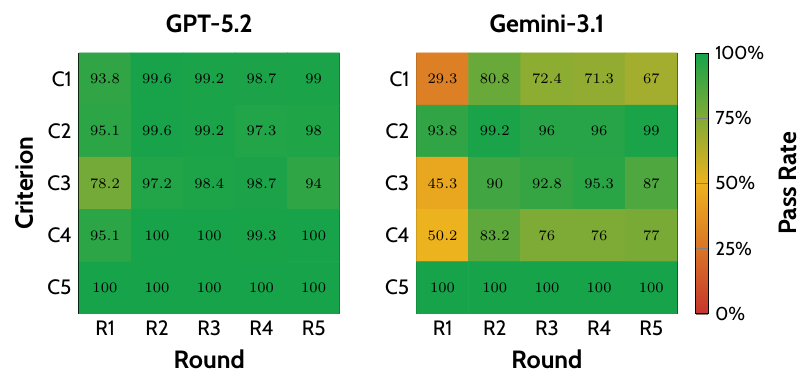}
\caption{Per-criterion pass rates (\%). Left: GPT-5.2; right: Gemini~3.1 Pro. Criteria: C1 Coherence \& Tactic Fit, C2 History Alignment, C3 Tactic-Dialogue Alignment, C4 Authenticity \& Skill Consistency, C5 Format Compliance.}
\label{fig:s4-heatmap}
\end{figure}

\begin{figure}[!htb]
\centering
\includegraphics[width=0.6\linewidth]{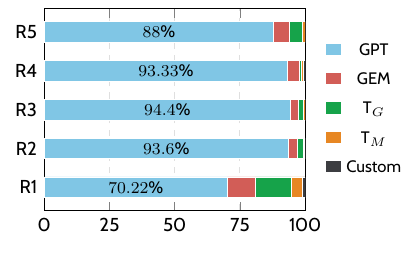}
\caption{Selection and correction breakdown (\%) per round. GPT: GPT-5.2 direct; GEM: Gemini~3.1 Pro direct; T\textsubscript{G} (for GPT)\,/\,T\textsubscript{M} (for GEM): targeted correction; Custom: custom regeneration.}
\label{fig:s4-bars}
\end{figure}

\begin{figure*}[!htb]
\centering
\subcaptionsetup{width=0.9\linewidth}
\begin{subfigure}[t]{0.47\textwidth}
  \centering
  \includegraphics[width=0.99\linewidth]{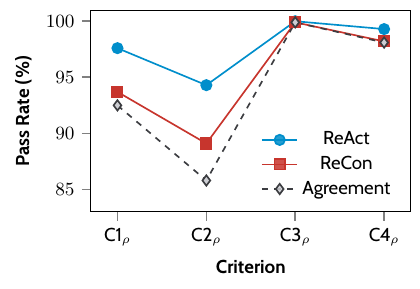}
  \caption{Mean judge score $\bar{A}$ with $\pm$1\,SD error bars.}
  \label{fig:s4-trend-avg}
\end{subfigure}
\hfill
\begin{subfigure}[t]{0.47\textwidth}
  \centering
  \includegraphics[width=0.99\linewidth]{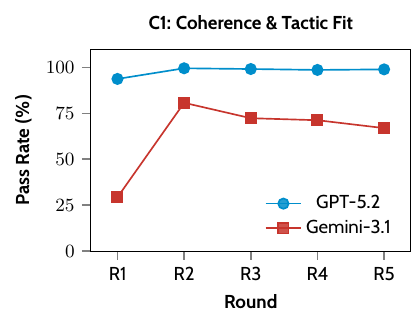}
  \caption{C1: Coherence \& Tactic Situational Fit.}
  \label{fig:s4-trend-c1}
\end{subfigure}
\\[1ex]
\begin{subfigure}[t]{0.47\textwidth}
  \centering
  \includegraphics[width=0.99\linewidth]{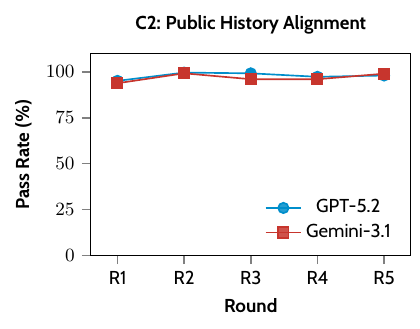}
  \caption{C2: Public History Alignment.}
  \label{fig:s4-trend-c2}
\end{subfigure}
\hfill
\begin{subfigure}[t]{0.47\textwidth}
  \centering
  \includegraphics[width=0.99\linewidth]{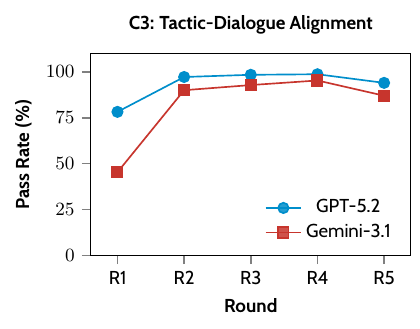}
  \caption{C3: Tactic-Dialogue Alignment.}
  \label{fig:s4-trend-c3}
\end{subfigure}
\\[1ex]
\begin{subfigure}[t]{0.47\textwidth}
  \centering
  \includegraphics[width=0.99\linewidth]{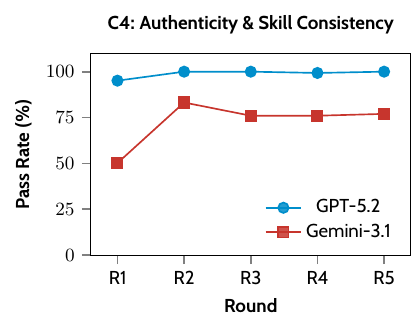}
  \caption{C4: Authenticity \& Skill Consistency.}
  \label{fig:s4-trend-c4}
\end{subfigure}
\hfill
\begin{subfigure}[t]{0.47\textwidth}
  \centering
  \includegraphics[width=0.99\linewidth]{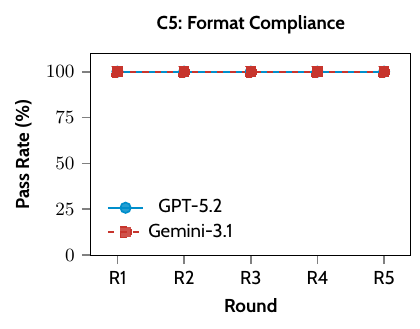}
  \caption{C5: Format Compliance (both models 100\% every round).}
  \label{fig:s4-trend-c5}
\end{subfigure}
\caption{Per-round score trajectories across R1--R5. Panel~(a) shows mean judge scores ($\bar{A}$) out of~5 with $\pm$1\,SD bands; panels~(b)--(f) show per-criterion pass rates (\%). GPT-5.2 shown in blue; Gemini~3.1 Pro in red.}
\label{fig:s4-trends}
\end{figure*}

Table~\ref{tab:verifier-results} reports the per-round Claude Sonnet~4.5 judge scores, selections, and corrections for GPT-5.2 and Gemini~3.1 Pro candidates.
Table~\ref{tab:per-criterion} reports per-criterion pass rates across all five rounds.
Coherence and Tactic Situational Fit is the weakest criterion for Gemini~3.1 Pro across all rounds (29.3--80.8\%), whereas GPT-5.2 is weakest on Tactic-Dialogue Alignment in R1 (78.2\%) and otherwise near-ceiling on all criteria.
Format Compliance reaches 100\% for both models in every round.
Figure~\ref{fig:s4-heatmap} shows the per-criterion pass-rate heatmap.
Figure~\ref{fig:s4-bars} shows the per-round selection and correction breakdown.
Figure~\ref{fig:s4-trends} shows per-round score trajectories for the average judge score and each criterion.

\begin{table*}[!htb]
\sisetup{
  text-family-to-math = true,
  text-series-to-math = true,
  round-mode          = places
}
\centering
\caption{Per-round verification outcomes by Claude Sonnet~4.5. $\bar{A}_{\text{GPT}}$ and $\bar{A}_{\text{GEM}}$ are mean judge scores out of~5 for GPT-5.2 and Gemini~3.1 Pro. GPT and GEM indicate direct-selection rates; T\fsub{GPT} and T\fsub{GEM} denote targeted-correction rates; Custom denotes full custom-regeneration rate. The five columns partition all samples (\%) and sum to 100. Pair is the pairwise-tiebreak rate; L2 marks rounds that successfully underwent a second-layer recheck.}
\label{tab:verifier-results}
\rowcolors{2}{white}{orange!10}
\resizebox{0.95\textwidth}{!}{%
\begin{tabular}{l
  S[table-format=3,   round-precision=0]
  S[table-format=1.2, round-precision=2]
  S[table-format=1.2, round-precision=2]
  S[table-format=2.2, round-precision=2]
  S[table-format=2.2, round-precision=2]
  S[table-format=2.2, round-precision=2]
  S[table-format=1.2, round-precision=2]
  S[table-format=1.2, round-precision=2]
  S[table-format=2.2, round-precision=2]
  c
}
\toprule
& &
\multicolumn{2}{c}{\textbf{Mean Scores}} &
\multicolumn{5}{c}{\textbf{Selection \& Correction} (\%, $\Sigma{=}100$)} &
\multicolumn{2}{c}{\textbf{Additional}} \\
\cmidrule(lr){3-4}\cmidrule(lr){5-9}\cmidrule(lr){10-11}
{\textbf{Round}} & {\textbf{N}} &
{$\bar{A}_{\text{GPT}}$} & {$\bar{A}_{\text{GEM}}$} &
{GPT} & {GEM} & {T\fsub{GPT}} & {T\fsub{GEM}} & {Custom} &
{Pairwise} & {L2} \\
\midrule
R1 & 225 & 4.62 & 3.19 & 70.22 & 10.67 & 13.78 & 4.44 & 0.89 & 3.56  & {\color{cgreen}$\checkmark$} \\
R2 & 250 & 4.96 & 4.53 & 93.60 &  3.60 &  2.00 & 0.40 & 0.40 & 66.00 & {\color{cgreen}$\checkmark$} \\
R3 & 250 & 4.97 & 4.37 & 94.40 &  3.20 &  1.60 & 0.80 & 0.00 & 58.80 & {\color{cgreen}$\checkmark$} \\
R4 & 150 & 4.94 & 4.39 & 93.33 &  4.67 &  0.67 & 0.67 & 0.67 & 56.67 & {\color{cgreen}$\checkmark$} \\
R5 & 100 & 4.91 & 4.30 & 88.00 &  6.00 &  5.00 & 1.00 & 0.00 & 51.00 & {\color{cgreen}$\checkmark$} \\
\bottomrule
\end{tabular}%
}
\end{table*}

\begin{table*}[htb]
\sisetup{
  text-family-to-math = true,
  text-series-to-math = true,
  round-mode          = places,
  round-precision     = 1
}
\centering
\caption{Per-criterion pass rates (\%) for Gemini~3.1 Pro (GEM) and GPT-5.2 (GPT) across all five rounds.
Criteria: (1) Coherence and Tactic Situational Fit, (2) Public History Alignment, (3) Tactic-Dialogue Alignment, (4) Authenticity and Skill Consistency, (5) Format Compliance.}
\label{tab:per-criterion}
\resizebox{\textwidth}{!}{%
\begin{tabular}{l *{10}{S[table-format=3.1]}}
\toprule
& \multicolumn{2}{c}{\textbf{\textcolor{ncsuDarkPurple}{R1}} ($n$=225)}
& \multicolumn{2}{c}{\textbf{\textcolor{ncsuDarkPurple}{R2}} ($n$=250)}
& \multicolumn{2}{c}{\textbf{\textcolor{ncsuDarkPurple}{R3}} ($n$=250)}
& \multicolumn{2}{c}{\textbf{\textcolor{ncsuDarkPurple}{R4}} ($n$=150)}
& \multicolumn{2}{c}{\textbf{\textcolor{ncsuDarkPurple}{R5}} ($n$=100)} \\
\cmidrule(lr){2-3}\cmidrule(lr){4-5}\cmidrule(lr){6-7}\cmidrule(lr){8-9}\cmidrule(lr){10-11}
{\textbf{Criterion}} & {GPT} & {GEM} & {GPT} & {GEM} & {GPT} & {GEM} & {GPT} & {GEM} & {GPT} & {GEM} \\
\midrule
\rowcolor{mint!8}
(1) Coherence       & 93.8 & 29.3 & 99.6 & 80.8 & 99.2 & 72.4 & 98.7 & 71.3 & 99.0 & 67.0 \\
(2) History         & 95.1 & 93.8 & 99.6 & 99.2 & 99.2 & 96.0 & 97.3 & 96.0 & 98.0 & 99.0 \\
\rowcolor{mint!8}
(3) Tactic-Dialogue & 78.2 & 45.3 & 97.2 & 90.0 & 98.4 & 92.8 & 98.7 & 95.3 & 94.0 & 87.0 \\
(4) Authenticity    & 95.1 & 50.2 & 100.0 & 83.2 & 100.0 & 76.0 & 99.3 & 76.0 & 100.0 & 77.0 \\
\rowcolor{mint!8}
(5) Format          & 100.0 & 100.0 & 100.0 & 100.0 & 100.0 & 100.0 & 100.0 & 100.0 & 100.0 & 100.0 \\
\bottomrule
\end{tabular}%
}
\end{table*}

\FloatBarrier

Prompt Boxes~\ref{box:s4-layer1} and~\ref{box:s4-layer2} show the structure of the Layer~1 pair evaluation and Layer~2 single-dialogue recheck prompts for a Round~4 example.
Content common to all games is abbreviated with $[\ldots]$.

\begin{tcolorbox}[
  colback=bamber,
  colframe=camber,
  coltitle=black,
  fonttitle=\sffamily\small\bfseries,
  title={Prompt Box~\thepromptbox. Layer~1: Pair Evaluation (Round~4 Example)},
  step and label={promptbox}{box:s4-layer1},
  breakable,
  top=4pt, bottom=4pt, left=6pt, right=6pt,
]
\ttfamily\small
\textbf{System:} You are an expert verifier for the Avalon social deception game with deep knowledge of deception tactics and game mechanics.
Your task is to evaluate pairs of AI-generated dialogue responses, assess their quality, and determine the best outcome: select the superior response as-is, apply targeted corrections to fix specific issues in the better response, or generate an entirely new dialogue with appropriate tactics and scales if both responses are inadequate.\\[3pt]
\textbf{User:}\\
You are evaluating two AI-generated Avalon Round~4 discussion logs to determine which is superior.
Both responses were generated for the same game context.
Your task is to validate both using binary pass/fail checks on 5 criteria, then recommend the best outcome.\\[3pt]
\# CONTEXT\\
\textbf{Game:} G042 \quad \textbf{Round:} 4\\
\textbf{Player Roles:} P1: Evil, P2: Good, P3: Good (protagonist), P4: Evil, P5: Good\\
\textbf{Protagonist (does NOT speak in dialogue):} P3\\[3pt]
\textbf{Prior Round Summaries:}\\
$[\ldots\text{ condensed narrative summaries of Rounds 1--3 for game G042} \ldots]$\\[3pt]
\textbf{Public History (current round):}\\
$[\ldots\text{ same cumulative public history block as Stage~3 Pass~1 and Pass~2} \ldots]$\\[3pt]
---\\[3pt]
\textbf{RESPONSE A} (Gemini~3.1 Pro --- model identity withheld from judge):\\
Discussion after Quest 4:\\
~~P5: ``The R2 failure with P2/P4/P5 and the R1 pass with P1/P2/P3 make P4 the clearest suspect.''\\
~~P4: ``P3 has been suspiciously quiet and didn't vote for our team. Maybe P3 is evil.''\\
~~$[\ldots\text{ P2 and P1 utterances} \ldots]$\\[3pt]
\textbf{matrix\_tactic\_scale (Response A):}\\
~~P5 (Good, GRS High): `Evidence Synthesis' [Transparent $\times$ Cooperative]\\
~~P4 (Evil, Mach-IV Moderate): `False Accusation' [Counterfactual $\times$ Opportunistic]\\
~~P2 (Good, GRS Moderate): `Tentative Reasoning' [Framing $\times$ Cooperative]\\
~~P1 (Evil, Mach-IV High): `Aggressive Misdirection' [Counterfactual $\times$ Adversarial]\\[3pt]
---\\[3pt]
\textbf{RESPONSE B} (GPT-5.2 --- model identity withheld from judge):\\
$[\ldots\text{ same game context; independent Round~4 dialogue generated by GPT-5.2} \ldots]$\\[3pt]
\textbf{matrix\_tactic\_scale (Response B):}\\
$[\ldots\text{ GPT-5.2 tactic annotations for the same four speakers} \ldots]$\\[3pt]
---\\[3pt]
\textbf{TACTIC ANNOTATION FRAMEWORK:} Both responses were generated from identical pre-assigned tactic plans.
Neither matrix is canonical; treat as starting reference, not ground truth.\\
$[\ldots\text{ full 37-tactic taxonomy, row/column definitions, and scale descriptions} \ldots]$\\[3pt]
\textbf{EVALUATION CRITERIA (1 = PASS, 0 = FAIL):}\\
(1) \textbf{Coherence \& Tactic Situational Fit}: tactic is socially congruous with the discussion dynamics (conversational position, what others are saying, how tensions develop).\\
(2) \textbf{Public History Alignment}: consistent with quest outcomes, teams, and voting from the current round and prior round summaries.\\
(3) \textbf{Tactic-Dialogue Alignment}: annotation correctly names what was said; tactic is valid in the taxonomy; row/col match the 4$\times$4 matrix; scale and level are accurate.\\
(4) \textbf{Authenticity \& Skill Consistency}: believable Avalon exchange; utterance sophistication matches the player's assigned skill level.\\
(5) \textbf{Format Compliance}: starts with ``Discussion after Quest 4:'', exactly 4 nonprotagonist speakers, each speaks exactly once.\\[3pt]
\textbf{DECISION LOGIC:}\\
~~1. Both 5/5 $\Rightarrow$ \texttt{PAIRWISE\_TIEBREAKER}: qualitative comparison selects the stronger response\\
~~2. One 5/5, other $<$5 $\Rightarrow$ select the 5/5 response directly\\
~~3. At least one 4/5 $\Rightarrow$ \texttt{CORRECTION\_A} or \texttt{CORRECTION\_B}: targeted inline fix of the failing criterion; if both 4/5, pairwise comparison selects the base\\
~~4. Both $\leq$3/5 $\Rightarrow$ \texttt{CUSTOM\_GENERATION}: generate a new dialogue from scratch\\[3pt]
\textbf{Output:} JSON with \texttt{response\_a}/\texttt{response\_b} (per-criterion scores and totals), \texttt{recommendation}, \texttt{pairwise\_winner}, \texttt{failed\_criterion}, \texttt{targeted\_correction} (corrected dialogue + matrix), \texttt{custom\_dialogue} (new dialogue + matrix), and \texttt{reasoning}.
\end{tcolorbox}

\begin{tcolorbox}[
  colback=bamber,
  colframe=camber,
  coltitle=black,
  fonttitle=\sffamily\small\bfseries,
  title={Prompt Box~\thepromptbox. Layer~2: Single-Dialogue Recheck (Round~4 Example)},
  step and label={promptbox}{box:s4-layer2},
  breakable,
  top=4pt, bottom=4pt, left=6pt, right=6pt,
]
\ttfamily\small
\textbf{System:} $[\ldots\text{ same system prompt as Layer~1} \ldots]$\\[3pt]
\textbf{User:}\\
You are evaluating a single AI-generated Avalon discussion log in Round~4 for quality.
The dialogue has already undergone one round of correction.
Your default is to \textbf{ACCEPT} it; only flag a criterion as failing when there is a clear, objective, nondebatable violation.
You must not be nitpicky or subjective.
Score it on 5 binary criteria and, if it fails any criterion, apply the minimum fix required to pass that criterion.\\[3pt]
\# CONTEXT\\
\textbf{Game:} G042 \quad \textbf{Round:} 4\\
\textbf{Player Roles:} P1: Evil, P2: Good, P3: Good (protagonist), P4: Evil, P5: Good\\
\textbf{Protagonist (does NOT speak in dialogue):} P3\\[3pt]
\textbf{Prior Round Summaries:}\\
$[\ldots\text{ same condensed narrative summaries as Layer~1} \ldots]$\\[3pt]
\textbf{Public History (current round):}\\
$[\ldots\text{ same public history block as Layer~1} \ldots]$\\[3pt]
---\\[3pt]
\textbf{DIALOGUE TO EVALUATE} (Layer~1 targeted correction output):\\
Discussion after Quest 4:\\
~~P5: ``The R2 failure with P2/P4/P5 and the R1 pass make P4 the clearest suspect on the evidence.''\\
~~P4: ``P3 voted against our team in R4 --- that warrants scrutiny before we finalize R5.''\\
~~$[\ldots\text{ P2 and P1 utterances} \ldots]$\\[3pt]
\textbf{Matrix Tactic Scale:}\\
~~P5 (Good, GRS High): `Evidence Synthesis' [Transparent $\times$ Cooperative]\\
~~P4 (Evil, Mach-IV Moderate): `False Accusation' [Counterfactual $\times$ Opportunistic]\\
~~P2 (Good, GRS Moderate): `Tentative Reasoning' [Framing $\times$ Cooperative]\\
~~P1 (Evil, Mach-IV High): `Aggressive Misdirection' [Counterfactual $\times$ Adversarial]\\[3pt]
---\\[3pt]
\textbf{TACTIC ANNOTATION FRAMEWORK:}\\
$[\ldots\text{ same 37-tactic taxonomy, row/column definitions, and scale descriptions as Layer~1} \ldots]$\\[3pt]
\textbf{EVALUATION CRITERIA (1 = PASS, 0 = FAIL):}\\
$[\ldots\text{ same 5 criteria as Layer~1} \ldots]$\\
Benefit-of-the-doubt rule: when uncertain whether a flaw reaches the 0/1 threshold, always score as passing (1).
Only assign 0 when the failure is absolutely clear, objective, and nondebatable.\\[3pt]
\textbf{DECISION LOGIC:}\\
~~1. 5/5 $\Rightarrow$ \texttt{ACCEPT}: accept the dialogue as-is\\
~~2. 4/5 $\Rightarrow$ \texttt{TARGETED\_CORRECTION}: fix the single failing criterion in the provided dialogue\\
~~3. $\leq$3/5 $\Rightarrow$ \texttt{FULL\_REGENERATION}: generate a completely new dialogue that passes all 5 criteria\\[3pt]
\textbf{Output:} JSON with \texttt{criteria} (per-criterion scores and explanations), \texttt{total}, \texttt{recommendation}, \texttt{failed\_criterion}, \texttt{targeted\_correction} (corrected dialogue + matrix), \texttt{regenerated\_dialogue} (new dialogue + matrix), and \texttt{reasoning}.
\end{tcolorbox}

\FloatBarrier

\subsection{Stage 5}
\label{app:s5}

Prompt Boxes~\ref{box:s5-react} and~\ref{box:s5-recon} show the structure of the \fsl{ReAct} and \fsl{ReCon} generation prompts for a Round~4 example.
Content common to all games is abbreviated with $[\ldots]$.

\begin{tcolorbox}[
  colback=ncsuLightBlue!18,
  colframe=cslate,
  colbacktitle=cslate,
  coltitle=white,
  fonttitle=\sffamily\small\bfseries,
  title={Prompt Box~\thepromptbox. ReAct: Reasoning Trace Generation (Round~4 Example)},
  step and label={promptbox}{box:s5-react},
  breakable,
  top=4pt, bottom=4pt, left=6pt, right=6pt,
]
\ttfamily\small
\textbf{System:} You are P3, the Investigator --- a Good player in a modified 5-player Avalon game.
Reason carefully and produce only valid JSON.\\[3pt]
\textbf{User:}\\
$[$\ldots\texttt{Avalon game overview, setup, round sequence, Investigator role, Critical Self-Knowledge, and Internal Reasoning Framework (Steps 1--6: Abduction, Suspicion, First-Order Beliefs, Second-Order Beliefs, Statement, Deduction)} \ldots$]$\\[3pt]
---\\[3pt]
\# CURRENT GAME CONTEXT\\
\textbf{Game:} G042 \quad \textbf{Round:} R4 \quad \textbf{Protagonist:} P3\\
\textbf{Player speaking order this round:} P5, P4, P2, P1\\[3pt]
\textbf{**CUMULATIVE PUBLIC HISTORY**}\\
Round 1 | Leader: P1 | Team: P1, P2, P3 | Votes: P1:Y P2:Y P3:Y P4:N P5:Y | Quest 1: PASS\\
Round 2 | Leader: P2 | Team: P2, P4, P5 | Votes: P1:Y P2:Y P3:N P4:Y P5:Y | Quest 2: FAIL\\
Round 3 | Leader: P3 | Team: P1, P2, P3 | Votes: P1:Y P2:Y P3:Y P4:Y P5:Y | Quest 3: PASS\\
Round 4 | Leader: P4 | Team: P2, P4, P5 | Votes: P1:Y P2:Y P3:N P4:Y P5:Y | Quest 4: FAIL\\[3pt]
\textbf{**CUMULATIVE DISCUSSION LOG**}\\
$[\ldots\text{ verbatim discussion logs for Rounds 1--4} \ldots]$\\[3pt]
\textbf{**PRIOR ROUND REASONING (ReAct chain)**}\\
$[\ldots\text{ R1--R3 ReAct reasoning traces produced by the same framework} \ldots]$\\[3pt]
---\\[3pt]
Reasoning (ReAct $\to$ adapted for modified Avalon):\\[3pt]
\textbf{Thought 1:} Identify the single most diagnostic unresolved question about player alignments after this round.
Focus on quest outcomes, team composition, and player statements relative to your own participation.\\[3pt]
\textbf{Action 1:} Select one concrete, named piece of evidence from the cumulative public history above.
State it explicitly.\\[3pt]
\textbf{Observation 1:} State precisely what that evidence implies about one or more players' alignments.
Apply the Definitive Deduction Rule if applicable.\\[3pt]
\textbf{Thought 2:} Based on Observation~1, state your updated hypothesis.
Which player is your top suspect and on what basis?
(Anti-loop: if Thought~2 would only restate Thought~1 with no new evidence, examine a genuinely different clue before continuing.)\\[3pt]
\textbf{Action 2:} Cross-check your hypothesis against one specific statement or voting behavior from the cumulative discussion log.
Cite the player and the content explicitly.\\[3pt]
\textbf{Observation 2:} Does the cross-check confirm, revise, or reinforce your Thought~2 hypothesis?
State your final confidence level.\\[3pt]
\textbf{Thought 3 (Synthesis):} Based on your two-step analysis, synthesize your final assessment.
Which players warrant second-order belief analysis (Strong/Extreme suspicion)?
What would you say to the group, grounded only in observable evidence?
State your final deduction of 1--3 Evil players.\\[3pt]
Now produce the final JSON, maintaining speaking order (P5, P4, P2, P1) in abduction, suspicion, and beliefs:\\
$[$\ldots\texttt{ output schema: abduction (good\_expl, evil\_expl, choice per player), suspicion (level, reason), depth, beliefs ($\beta_1^i$ all players; $\beta_2^i$ Strong/Extreme only), statement, deduction} \ldots $]$
\end{tcolorbox}

\begin{tcolorbox}[
  colback=ncsuLightBlue!18,
  colframe=cslate,
  colbacktitle=cslate,
  coltitle=white,
  fonttitle=\sffamily\small\bfseries,
  title={Prompt Box~\thepromptbox. ReCon: Reasoning Trace Generation (Round~4 Example)},
  step and label={promptbox}{box:s5-recon},
  breakable,
  top=4pt, bottom=4pt, left=6pt, right=6pt,
]
\ttfamily\small
\textbf{System:} You are P3, the Investigator --- a Good player in a modified 5-player Avalon game.
Reason carefully and produce only valid JSON.\\[3pt]
\textbf{User:}\\
$[\ldots\text{ same static Avalon overview, setup, and Internal Reasoning Framework as ReAct} \ldots]$\\[3pt]
---\\[3pt]
\# CURRENT GAME CONTEXT\\
\textbf{Game:} G042 \quad \textbf{Round:} R4 \quad \textbf{Protagonist:} P3\\
\textbf{Player speaking order this round:} P5, P4, P2, P1\\[3pt]
\textbf{**CUMULATIVE PUBLIC HISTORY**}\\
$[\ldots\text{ same four-round public history block as ReAct} \ldots]$\\[3pt]
\textbf{**CUMULATIVE DISCUSSION LOG**}\\
$[\ldots\text{ verbatim discussion logs for Rounds 1--4} \ldots]$\\[3pt]
\textbf{**PRIOR ROUND REASONING (ReCon chain)**}\\
$[\ldots\text{ R1--R3 ReCon reasoning traces produced by the same framework} \ldots]$\\[3pt]
---\\[3pt]
Reasoning (ReCon $\to$ adapted for modified Avalon):\\[3pt]
\textbf{Formulation contemplation} (first-order perspective transition):\\
For each non-protagonist player in speaking order (P5, P4, P2, P1), infer their probable alignment based on quest participation, voting record, and speech in the discussion log.
Adopt each player's perspective: given what they know, would an Evil player in their position behave differently from a Good player here?
Form an initial role assessment and suspicion level for each player, grounded only in observable evidence.\\[3pt]
\textbf{Refinement contemplation} (second-order perspective transition):\\
Revisit your initial formulation from the perspectives of other players.
Check three issues: (a) \textbf{Evidence Sufficiency} --- are your strongest claims supported by explicit observable evidence (quest outcomes, votes, or quoted statements)?
(b) \textbf{Cross-Round Consistency} --- does your deduction stay consistent with prior rounds unless new evidence justifies a revision?
(c) \textbf{Public Statement Alignment} --- your statement field must be what you would plausibly say aloud to other players, strategic yet grounded in observable evidence.
Revise suspicion levels, beliefs, and statement to eliminate any issues.\\[3pt]
Now produce the final JSON, maintaining speaking order (P5, P4, P2, P1) in abduction, suspicion, and beliefs:\\
$[\ldots\text{ same output schema as ReAct} \ldots]$
\end{tcolorbox}

\FloatBarrier

\subsection{Stage 6}
\label{app:s6}

\begin{figure}[t]
\centering
\subcaptionsetup{width=0.9\linewidth}
\begin{subfigure}[t]{0.49\textwidth}
  \centering
  \includegraphics[width=0.99\linewidth]{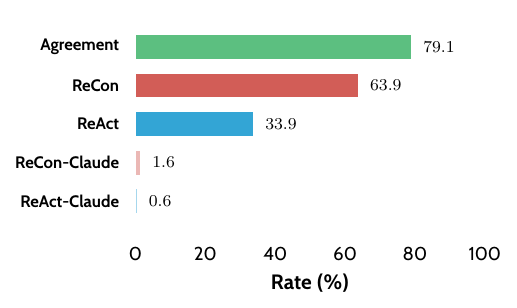}
    \caption{Reasoning framework selection and pairwise score agreement rates.
    Agreements reports the proportion of pairs where both traces score 4/4, requiring a pairwise tiebreaker.
    ReAct-Claude and ReCon-Claude denote corrected outputs from the Layer~2 recheck.}
\label{fig:s6-sel}
\end{subfigure}
\hfill
\begin{subfigure}[t]{0.49\textwidth}
  \centering
  \includegraphics[width=0.99\linewidth]{tikz/pdf/line_chart}
    \caption{Per-criterion pass rates (\%) for ReAct and ReCon traces in Stage~6 ($N{=}1{,}000$).
    Agreement reports the proportion of pairs receiving the same binary score per criterion.
    $\text{C1}_{\rho}$: Abduction Plausibility; $\text{C2}_{\rho}$: Suspicion Level Consistency; $\text{C3}_{\rho}$: Gated Depth Reasoning; $\text{C4}_{\rho}$: Statement \& Deduction Coherence.}
    \label{fig:s6-criteria}
\end{subfigure}
\label{fig:s6-rate-criterion}
\caption{Stage~6 reasoning framework (ReAct, ReCon) selection rates by Claude Sonnet~4.6 and passing rates of their criteria along with agreements.}
\end{figure}

Table~\ref{tab:s6-full} reports comprehensive per-framework verification statistics for Stage~6.
Table~\ref{tab:s6-criteria} provides a focused view of per-criterion pass rates and inter-framework agreement.
Figure~\ref{fig:s6-sel} shows the reasoning framework selection rates by Claude Sonnet~4.6, along with the agreements where both frameworks achieve a 4/4 score.
Figure~\ref{fig:s6-criteria} shows the corresponding per-criterion pass rate trends across frameworks.

\begin{table*}[htb]
\sisetup{
  text-family-to-math = true,
  text-series-to-math = true,
  round-mode          = places,
  round-precision     = 1,
}
\centering
\caption{%
  Comprehensive Stage~6 verification statistics for ReAct and ReCon traces ($N{=}1{,}000$).
  \emph{Pass rates}: binary pass rate per criterion (\%).
  \emph{Score}: mean total score with SD; distribution of 3/4, 2/4, and ${\leq}$1/4 outcomes (\%).
  \emph{Selection}: rate of direct selection, corrected-trace selection (ReAct-Claude or ReCon-Claude), and combined total (\%).
  $\Delta$ row shows ReAct minus ReCon.
  The pairwise statistics row reports per-criterion agreement (\%) under C1--C4, and pair outcome fractions aligned under the selection columns: both traces score 4/4 (resolved by pairwise tiebreaker), exactly one scores 4/4 (direct selection), or both score below 4/4 (targeted correction).}
\label{tab:s6-full}
\resizebox{\textwidth}{!}{%
\begin{tabular}{l
  S[table-format=3.1]
  S[table-format=3.1]
  S[table-format=3.1]
  S[table-format=3.1]
  S[table-format=3.1]
  S[table-format=1.2]
  S[table-format=1.2]
  S[table-format=2.1]
  S[table-format=2.1]
  S[table-format=2.1]
  S[table-format=2.1]
  S[table-format=2.1]
  S[table-format=2.1]
}
\toprule
& \multicolumn{5}{c}{\textbf{Pass Rate (\%)}}
& \multicolumn{5}{c}{\textbf{Score Distribution}}
& \multicolumn{3}{c}{\textbf{Selection (\%)}} \\
\cmidrule(lr){2-6}\cmidrule(lr){7-11}\cmidrule(lr){12-14}
{\textbf{Framework}}
& {\textbf{C1}} & {\textbf{C2}} & {\textbf{C3}} & {\textbf{C4}} & {\textbf{4/4}}
& {\textbf{Mean}} & {\textbf{SD}}
& {\textbf{3/4}} & {\textbf{2/4}} & {$\mathbf{\leq}$\textbf{1/4}}
& {\textbf{Direct}} & {\textbf{Corr.}} & {\textbf{Total}} \\
\midrule
ReAct & 97.6 & 94.3 & 100.0 & 99.3 & 92.5 & 3.91 & 0.33 & 6.2 & 1.3 & 0.0 & 33.9 & 0.6 & 34.5 \\
ReCon & 93.7 & 89.1 & 99.9  & 98.2 & 84.4 & 3.81 & 0.48 & 12.2 & 3.3 & 0.1 & \bfseries 63.9 & 1.6 & 65.5 \\
\midrule
\multicolumn{1}{l}{$\Delta$}
  & \multicolumn{1}{c}{$+$3.9} & \multicolumn{1}{c}{$+$5.2} & \multicolumn{1}{c}{$+$0.1} & \multicolumn{1}{c}{$+$1.1} & \multicolumn{1}{c}{$+$8.1}
  & \multicolumn{1}{c}{$+$0.10} & \multicolumn{1}{c}{$-$0.15}
  & \multicolumn{1}{c}{$-$6.0} & \multicolumn{1}{c}{$-$2.0} & \multicolumn{1}{c}{$-$0.1}
  & \multicolumn{1}{c}{$-$30.0} & \multicolumn{1}{c}{$-$1.0} & \multicolumn{1}{c}{$-$31.0} \\
\midrule\midrule
\multicolumn{1}{l}{\textit{Agreement (\%):}}
  & \multicolumn{1}{c}{92.5} & \multicolumn{1}{c}{85.8} & \multicolumn{1}{c}{99.9} & \multicolumn{1}{c}{98.1}
  & \multicolumn{1}{c}{---}
  & \multicolumn{6}{c}{---}
  & \multicolumn{2}{c}{---} \\
\multicolumn{1}{l}{\textit{Pair outcome (\%):}}
  & \multicolumn{10}{c}{---}
  & \multicolumn{1}{c}{\bfseries 79.1} & \multicolumn{1}{c}{18.7} & \multicolumn{1}{c}{2.2} \\
\bottomrule
\end{tabular}%
}
\end{table*}

\begin{table*}[!htb]
\sisetup{round-mode=places, round-precision=1}
\centering
\caption{Per-criterion pass rates for ReAct and ReCon traces in Stage~6 ($N{=}1{,}000$). Agreement reports the percentage of pairs where both traces receive 4/4 score for that criterion.}
\label{tab:s6-criteria}
\rowcolors{2}{white}{ncsuLightBlue!25}
\begin{tabular}{l
  S[table-format=3.1]
  S[table-format=3.1]
  S[table-format=3.1]
}
\toprule
{\textbf{Criterion}} &
{\textbf{ReAct (\%)}} &
{\textbf{ReCon (\%)}} &
{\textbf{Agree. (\%)}} \\
\midrule
Abduction Plausibility  & 97.6  & 93.7  & 92.5 \\
Suspicion Consistency   & 94.3  & 89.1  & 85.8 \\
Gated Depth Reasoning   & 100.0 & 99.9  & 99.9 \\
Statement \& Deduction  & 99.3  & 98.2  & 98.1 \\
\midrule
\textbf{Overall (4/4)}  & \multicolumn{1}{c}{\textbf{92.5}} & \multicolumn{1}{c}{\textbf{84.4}} & \multicolumn{1}{c}{---} \\
\bottomrule
\end{tabular}
\end{table*}

\noindent Prompt Boxes~\ref{box:s6-layer1} and~\ref{box:s6-layer2} show the structure of the Layer~1 pair evaluation and Layer~2 single-trace recheck prompts for a Round~4 example.
Content common to all games is abbreviated with $[\ldots]$.

\begin{tcolorbox}[
  colback=ncsuLightBlue!18,
  colframe=cslate,
  colbacktitle=cslate,
  coltitle=white,
  fonttitle=\sffamily\small\bfseries,
  title={Prompt Box~\thepromptbox. Layer~1: Pair Evaluation (Round~4 Example)},
  step and label={promptbox}{box:s6-layer1},,
  breakable,
  top=4pt, bottom=4pt, left=6pt, right=6pt,
]
\ttfamily\small
\textbf{System:} You are an expert verifier for Avalon Theory of Mind reasoning traces.
Evaluate Trace~A and Trace~B using 4 binary PASS/FAIL criteria, apply the decision logic strictly, and return only valid JSON.
Judge from the Investigator's perspective using observable evidence only, not factual alignment truth.
Only fail a criterion when the violation is clear.\\[3pt]
\textbf{User:}\\
You are evaluating two AI-generated Theory of Mind reasoning traces for the same Avalon game round.
Apply 4 binary PASS/FAIL criteria, then apply the decision logic strictly.
Judge from the Investigator's perspective using observable evidence only.\\[3pt]
\# GAME CONTEXT\\
\textbf{Game:} G042 \quad \textbf{Round:} R4 \quad \textbf{Investigator:} P3 (always Good; always PASS; does NOT speak)\\
\textbf{Player Roles (ground truth):} P1: Evil, P2: Good, P3: Good, P4: Evil, P5: Good\\
\textbf{Is Final Round:} No\\[3pt]
$[$\ldots\texttt{ Avalon mechanics reference: team sizes, quest rules, Investigator constraints, Definitive Deduction Rule, gated depth rule, suspicion scale} \ldots$]$\\[3pt]
\textbf{Cumulative Public History (Rounds 1--4):}\\
$[\ldots\text{ same four-round history block as Stage~5 prompts} \ldots]$\\[3pt]
\textbf{Cumulative Round Summaries (Rounds 1--4):}\\
$[\ldots\text{ condensed narrative summaries of each round} \ldots]$\\[3pt]
\textbf{Cumulative Discussion Log (Rounds 1--4):}\\
$[\ldots\text{ verbatim discussion logs} \ldots]$\\[3pt]
\textbf{Prior Round Reasoning Golds (Rounds 1--3):}\\
$[\ldots\text{ gold reasoning traces selected in prior rounds} \ldots]$\\[3pt]
---\\[3pt]
\textbf{TRACE A} (ReAct --- framework identity withheld from judge):\\
$[$\ldots\texttt{ ReAct reasoning trace for G042 R4: abduction, suspicion, depth, beliefs, statement, deduction} \ldots$]$\\[3pt]
\textbf{TRACE B} (ReCon --- framework identity withheld from judge):\\
$[\ldots\text{ ReCon reasoning trace for G042 R4: same six fields} \ldots]$\\[3pt]
---\\[3pt]
\textbf{EVALUATION CRITERIA (1\,=\,PASS, 0\,=\,FAIL):}\\
(1) \textbf{Abduction Plausibility}: hypotheses $h_i$ cite specific observable evidence; each alignment choice reflects the strongest available inference; the Definitive Deduction Rule is enforced for any two-person failed quest involving the Investigator.\\
(2) \textbf{Suspicion Level Consistency}: suspicion levels $s_i$ are proportionate to cumulative public evidence; the Definitive Deduction Rule mandates Extreme for the two-person failed quest partner; Strong requires multiple converging cues.\\
(3) \textbf{Gated Depth Reasoning}: first-order beliefs $\beta_1^i$ exist for all players; second-order beliefs $\beta_2^i$ exist only for players at Strong or Extreme suspicion; depth field is set to~2 iff at least one player has Strong or Extreme suspicion.\\
(4) \textbf{Statement and Deduction Coherence}: final utterance $\hat{u}$ is grounded in observable evidence and plausible as in-character dialogue; deduction $\hat{\alpha}$ includes all Strong and Extreme suspicion players; final round requires exactly two named Evil players.\\[3pt]
\textbf{DECISION LOGIC:}\\
~~1. Both 4/4 $\Rightarrow$ \texttt{PAIRWISE\_TIEBREAKER}: choose the stronger trace, closer to ground-truth alignment.\\
~~2. One 4/4, other $<$4/4 $\Rightarrow$ select the 4/4 trace directly.\\
~~3. Both $<$4/4 $\Rightarrow$ \texttt{CORRECTION\_A} or \texttt{CORRECTION\_B}: correct the higher-scoring base; pairwise comparison breaks ties.\\
Hard rules: \texttt{CORRECTION\_A/B} only when both traces score $<$4/4; if one trace is 4/4, it must be selected.\\[3pt]
\textbf{EVALUATION PROCESS:}\\
~~1. Score Trace~A on all 4 criteria. \quad 2. Score Trace~B on all 4 criteria. \quad 3. Count passes for each trace.\\
~~4. Apply the decision logic. \quad 5. Return JSON in the format below.\\[3pt]
\textbf{Output:} JSON with \texttt{trace\_a}/\texttt{trace\_b} (per-criterion scores and totals), \texttt{recommendation}, \texttt{pairwise\_winner}, \texttt{pairwise\_reasoning}, \texttt{failed\_criteria}, and \texttt{targeted\_correction} (full corrected trace when applicable).
\end{tcolorbox}

\begin{tcolorbox}[
  colback=ncsuLightBlue!18,
  colframe=cslate,
  colbacktitle=cslate,
  coltitle=white,
  fonttitle=\sffamily\small\bfseries,
  title={Prompt Box~\thepromptbox. Layer~2: Single-Trace Recheck (Round~4 Example)},
  step and label={promptbox}{box:s6-layer2},
  breakable,
  top=4pt, bottom=4pt, left=6pt, right=6pt,
]
\ttfamily\small
\textbf{System:} $[\ldots\text{ same system prompt as Layer~1} \ldots]$\\[3pt]
\textbf{User:}\\
You are re-evaluating a single AI-generated Theory of Mind reasoning trace for an Avalon game round.
This trace has been selected as the best candidate and received a targeted correction.
Re-score it on the same 4 binary criteria.
The default is to \textbf{ACCEPT}; only fail a criterion when the violation is clear, objective, and non-debatable.\\[3pt]
\# GAME CONTEXT\\
\textbf{Game:} G042 \quad \textbf{Round:} R4 \quad \textbf{Investigator:} P3 (always Good; always PASS; does NOT speak)\\
\textbf{Player Roles (ground truth):} P1: Evil, P2: Good, P3: Good, P4: Evil, P5: Good\\
\textbf{Is Final Round:} No\\[3pt]
$[\ldots\text{ same Avalon mechanics reference as Layer~1} \ldots]$\\[3pt]
\textbf{Cumulative Public History (Rounds 1--4):}\\
$[\ldots\text{ same history block as Layer~1} \ldots]$\\[3pt]
\textbf{Cumulative Round Summaries (Rounds 1--4):}\\
$[\ldots\text{ condensed narrative summaries} \ldots]$\\[3pt]
\textbf{Cumulative Discussion Log (Rounds 1--4):}\\
$[\ldots\text{ verbatim discussion logs} \ldots]$\\[3pt]
\textbf{Prior Round Reasoning Golds (Rounds 1--3):}\\
$[\ldots\text{ gold traces from prior rounds} \ldots]$\\[3pt]
---\\[3pt]
\textbf{TRACE TO EVALUATE} (Layer~1 targeted correction output):\\
$[$\ldots\texttt{ the corrected reasoning trace from Layer~1: abduction, suspicion, depth, beliefs, statement, deduction} \ldots$]$\\[3pt]
---\\[3pt]
\textbf{EVALUATION CRITERIA (1\,=\,PASS, 0\,=\,FAIL):}\\
$[\ldots\text{ same 4 criteria as Layer~1} \ldots]$\\
Benefit-of-the-doubt rule: when uncertain whether a flaw reaches the pass/fail threshold, always score as passing.
Only assign 0 when the failure is absolutely clear, objective, and non-debatable.\\[3pt]
\textbf{Output:} JSON with per-criterion scores, \texttt{total}, \texttt{recommendation} (\texttt{ACCEPT} or \texttt{TARGETED\_CORRECTION}), \texttt{failed\_criteria}, and \texttt{targeted\_correction} (full corrected trace when applicable; null otherwise).
\end{tcolorbox}

\FloatBarrier

\section{Trace Deduction Accuracy}
\label{app:gold-ceiling}

Table~\ref{tab:gold-verdict} reports the deduction accuracy by game-ending round, whereas Table~\ref{tab:gold-perround} reports deduction accuracy for all 1,000 traces, grouped by round.
Accuracy rises from R1 to R5 as quest outcomes and discussions accumulate.
This rise indicates that the traces infer incrementally rather than apply a fixed heuristic.
The F1 over all rounds (0.842) is lower than the game-ending F1 (0.962) because early rounds lack sufficient evidence to identify the Evil pair.
Reporting the pooled F1 alone would understate the corpus; reporting it with the game-ending verdict shows where the difficulty lies.

\begin{table}[ht]
\sisetup{
  text-family-to-math = true,
  text-series-to-math = true,
  round-precision      = 3,
  round-mode           = places
}
\centering
\caption{Gold-trace deduction accuracy at game-ending round. \fbf{Exact} indicates the percentage of games that correctly identify both Evil players and no one else.}
\begin{tabular}{lc S[table-format=2.1,round-precision=1] *3{S[table-format=1.3]}}
\toprule
\rowcolor{ncsuBlue!15}
\textbf{\textcolor{ncsuBlue}{Ends}} &
{\textbf{\textcolor{ncsuBlue}{n}}} &
{\textbf{\textcolor{ncsuBlue}{Exact~(\si{\percent})}}} &
{\textbf{\textcolor{ncsuBlue}{P}}} &
{\textbf{\textcolor{ncsuBlue}{R}}} &
{\textbf{\textcolor{ncsuBlue}{F1}}} \\
\midrule
R3 & 100 & 93.0 & 0.960 & 0.960 & 0.960 \\
R4 &  50 & 86.0 & 0.930 & 0.930 & 0.930 \\
R5 & 100 & 97.0 & 0.980 & 0.980 & 0.980 \\
\midrule
\rowcolor{orange!10}
\textbf{All} & {\bfseries 250} & {\bfseries 93.2} & {\bfseries 0.962} & {\bfseries 0.962} & {\bfseries 0.962} \\
\bottomrule
\end{tabular}
\label{tab:gold-verdict}
\end{table}

\begin{table}[ht]
\sisetup{
  text-family-to-math = true,
  text-series-to-math = true,
  round-precision      = 3,
  round-mode           = places
}
\centering
\caption{Gold-trace deduction accuracy by round, all 1,000 traces. Same column annotation as Table~\ref{tab:gold-verdict}.}
\begin{tabular}{l l S[table-format=2.1,round-precision=1] *3{S[table-format=1.3]}}
\toprule
\rowcolor{ncsuBlue!15}
\textbf{\textcolor{ncsuBlue}{Round}} &
{\textbf{\textcolor{ncsuBlue}{n}}} &
{\textbf{\textcolor{ncsuBlue}{Exact~(\si{\percent})}}} &
{\textbf{\textcolor{ncsuBlue}{P}}} &
{\textbf{\textcolor{ncsuBlue}{R}}} &
{\textbf{\textcolor{ncsuBlue}{F1}}} \\
\midrule
R1 & 250 & 38.4 & 0.695 & 0.688 & 0.691 \\
R2 & 250 & 60.0 & 0.804 & 0.828 & 0.816 \\
R3 & 250 & 81.6 & 0.904 & 0.908 & 0.906 \\
R4 & 150 & 86.7 & 0.930 & 0.937 & 0.934 \\
R5 & 100 & 97.0 & 0.980 & 0.980 & 0.980 \\
\midrule
\rowcolor{ncsuOrange!10}
\textbf{All} & {\bfseries 1000} & {\bfseries 67.7} & {\bfseries 0.839} & {\bfseries 0.845} & {\bfseries 0.842} \\
\bottomrule
\end{tabular}
\label{tab:gold-perround}
\end{table}

\paragraph{What quest outcomes alone determine.}
For each game, there is a provision for six pairs ($\binom{4}{2}$) of Evil players, and a failed quest includes at least one Evil player while the Investigator is always Good.
To determine the valid Evil pairs remaining from quest outcomes, we keep each pair that includes a member of every failed team.
For instance, in G002, the Investigator P1 and player P4 fail the R1 quest.
Thus, we know that P4 is Evil, and the R2 team of P2, P3, and P5 also fails, so three pairs remain: (P2, P4), (P3, P4), and (P4, P5).
Table~\ref{tab:logic-only} groups the 250 games by the number of pairs that remain.
On \fsc{Avcorp}, quest outcomes single out one pair in 12 games, and two to six pairs remain in the other 238.
A reasoner that uses only quest outcomes and picks one of $k$ allowed pairs at random is correct with probability $1/k$, so a group of $n$ such games yields $n/k$ correct games in expectation (\fbf{Expected} in Table~\ref{tab:logic-only}).
Across the 238 games, this sum is $18.5 + 22 + 14 + 7 + 7.33 = 68.83$, an expected exact match of $68.83/238 = 28.9\%$, whereas the gold traces resolve 221 (92.9\%).
The difference comes from the dialogue that the traces use but the quest record lacks.

\begin{table}[t]
\centering
\caption{Games grouped by the number of Evil pairs that their quest outcomes allow. \fbf{Logic only} is the expected exact match of a random choice among the allowed pairs, and \fbf{Expected} is the number of games that this choice resolves in expectation; \fbf{Gold} counts the games that the gold traces resolve.}
\label{tab:logic-only}
\begin{tabular}{c c c c c}
\toprule
\rowcolor{ncsuBlue!15}
\textbf{\textcolor{ncsuBlue}{Allowed pairs}} & \textbf{\textcolor{ncsuBlue}{Games}} & \textbf{\textcolor{ncsuBlue}{Logic only (\%)}} & \textbf{\textcolor{ncsuBlue}{Expected}} & \textbf{\textcolor{ncsuBlue}{Gold}} \\
\midrule
1 & 12 & 100.0 & 12.00 & 12 \\
2 & 37 & 50.0 & 18.50 & 37 \\
3 & 66 & 33.3 & 22.00 & 63 \\
4 & 56 & 25.0 & 14.00 & 53 \\
5 & 35 & 20.0 & 7.00 & 29 \\
6 & 44 & 16.7 & 7.33 & 39 \\
\midrule
2 to 6 & 238 & 28.9 & 68.83 & 221 \\
\bottomrule
\end{tabular}
\end{table}

\paragraph{Review of divergent traces.}
We manually reviewed the games whose final verdict differed from $\alpha$.
One game-ending verdict, G124/4, excluded every member of a failed team.
It also broke the game's speaker order, indicating a disrupted generation.
We regenerated the failing trace under human supervision, and the trace now names the correct pair.
In each of the 17 games that still diverge, the trace names a pair that the quest record allows.
In several cases, the Evil players pass a shared quest to avoid suspicion.
No final verdict, therefore, contradicts its quest record.

\section{Annotator Agreement}
\label{app:human-study}

We sample 25 whole games (100 rounds) so that the two annotators follow each Investigator's reasoning across rounds, as the generator did.
The sample is representative: 79 of its 100 items are tiebreak cases (Table~\ref{tab:iaa}), against 79.1\% of the corpus.
The judge decided the other 21 on criterion scores alone.
The annotators receive the same Trace~A and Trace~B pairs as the judge.
They receive these pairs in a random order without a fixed position.
Since most games allow multiple Evil pairs consistent with the quest record (Appendix~\ref{app:gold-ceiling}), annotators score the reasoning rather than the correctness of its conclusion. 
The judge uses $\alpha$ to resolve tied pairs.
Judge agreement compares the judge's selection with each annotator's selection and their shared selection.
If the tiebreak rewarded correct conclusions over sound reasoning, judge agreement would drop on the tiebreak items.

\paragraph{Selection agreement.}
Confidence intervals for $\kappa$ are 95\% bootstrap intervals (2,000 resamples over items).
We compute $\kappa$ with scikit-learn and draw the bootstrap resamples with NumPy.
The judge matches the annotators' shared choice on 95 of the 97 rounds where they agree ($\kappa = .96$, 95\% CI $[.89, 1.00]$).
Table~\ref{tab:iaa-confusion} reports the selections behind these values; its last panel shows that the annotators share the judge's preference for ReCon.

\begin{table}[t]
\centering
\caption{Agreement on the final trace selection. A1 and A2 are the annotators and J the Stage~6 judge. \fbf{Agree} counts items on which the pair selects the same trace; \fbf{Both} is the judge against the annotators' shared choice, over the items on which they agree. Tiebreak items were decided by the judge using $\alpha$; clean-win items on criterion scores alone. PABAK is $2P_o - 1$, where $P_o$ is the share of items on which the pair agrees.}
\resizebox{0.95\textwidth}{!}{%
\begin{tabular}{l ccc ccc ccc}
\toprule
\rowcolor{mint!15}
 \whitecell{} & \multicolumn{3}{c}{\textbf{\textcolor{black}{All (100)}}}
 & \multicolumn{3}{c}{\textbf{\textcolor{black}{Tiebreak (79)}}}
 & \multicolumn{3}{c}{\textbf{\textcolor{black}{Clean win (21)}}} \\
\cmidrule(lr){2-4}\cmidrule(lr){5-7}\cmidrule(lr){8-10}
\rowcolor{mint!15}
\textbf{\textcolor{black}{Pair}} &
\textbf{\textcolor{black}{Agree}} & \textbf{\textcolor{black}{$\kappa$}} & \textbf{\textcolor{black}{PABAK}} &
\textbf{\textcolor{black}{Agree}} & \textbf{\textcolor{black}{$\kappa$}} & \textbf{\textcolor{black}{PABAK}} &
\textbf{\textcolor{black}{Agree}} & \textbf{\textcolor{black}{$\kappa$}} & \textbf{\textcolor{black}{PABAK}} \\
\midrule
A1 vs.\ A2  & 97/100 & 0.937 & 0.940 & 76/79 & 0.918 & 0.924 & 21/21 & 1.000 & 1.000 \\
A1 vs.\ J   & 96/100 & 0.917 & 0.920 & 77/79 & 0.946 & 0.949 & 19/21 & 0.809 & 0.810 \\
A2 vs.\ J   & 97/100 & 0.937 & 0.940 & 78/79 & 0.973 & 0.975 & 19/21 & 0.809 & 0.810 \\
\midrule
J vs.\ Both & 95/97  & 0.957 & 0.959 & 76/76 & 1.000 & 1.000 & 19/21 & 0.809 & 0.810 \\
\bottomrule
\end{tabular}
}
\label{tab:iaa}
\end{table}

\begin{table}[ht]
\centering
\small
\caption{Trace selections on the 100 items. In the first three panels, rows give the first rater and columns the second, and A and B are the positions in which the annotators saw the two traces: ReCon was Trace~A in 50 items and Trace~B in the other 50, and ReAct the reverse. The last panel counts each rater's selections by framework, regardless of position.}
\label{tab:iaa-confusion}
\begin{subtable}[t]{0.22\linewidth}
\centering
\caption*{\hlc[orange!15]{\mbox{A1 (rows) vs.\ A2}}}
\begin{tabular}{l ccc}
\toprule
 & A & B & All \\
\midrule
A & 59 & 1 & 60 \\
B & 2 & 38 & 40 \\
\midrule
All & 61 & 39 & 100 \\
\bottomrule
\end{tabular}
\end{subtable}\hfill
\begin{subtable}[t]{0.22\linewidth}
\centering
\caption*{\hlc[orange!15]{\mbox{A1 (rows) vs.\ J}}}
\begin{tabular}{l ccc}
\toprule
 & A & B & All \\
\midrule
A & 58 & 2 & 60 \\
B & 2 & 38 & 40 \\
\midrule
All & 60 & 40 & 100 \\
\bottomrule
\end{tabular}
\end{subtable}\hfill
\begin{subtable}[t]{0.22\linewidth}
\centering
\caption*{\hlc[orange!15]{\mbox{A2 (rows) vs.\ J}}}
\begin{tabular}{l ccc}
\toprule
 & A & B & All \\
\midrule
A & 59 & 2 & 61 \\
B & 1 & 38 & 39 \\
\midrule
All & 60 & 40 & 100 \\
\bottomrule
\end{tabular}
\end{subtable}\hfill
\begin{subtable}[t]{0.26\linewidth}
\centering
\caption*{\hlc[mint!15]{Selections by framework}}
\begin{tabular}{l ccc}
\toprule
 & A1 & A2 & J \\
\midrule
\textbf{ReCon} & 56 & 57 & 56 \\
ReAct & 44 & 43 & 44 \\
\bottomrule
\end{tabular}
\end{subtable}
\end{table}

\paragraph{Sampled games.}
We used stratified random sampling over each game's ending round (seed~42).
We allocated games in proportion to their frequency in the corpus (40\%, 20\%, and 40\% for R3, R4, and R5 endings).
R3-ending (10 games, 30 items): G004, G014, G015, G018, G059, G062, G066, G202, G207, G215.
R4-ending (5 games, 20 items): G026, G128, G175, G178, G228.
R5-ending (10 games, 50 items): G034, G035, G042, G086, G088, G090, G185, G192, G198, G242.
Three of the 25 sampled games (G062, G178, G228) are among the 17 games whose final verdict diverges from $\alpha$ (Section~\ref{sec:gold-ceiling}).

\begin{table}[t]
\centering
\caption{Per-criterion agreement, Trace~A and Trace~B pooled ($n = 200$ per row). \fbf{Pass} gives each rater's pass rate. $\kappa$ is undefined when both raters pass every trace.}
\label{tab:iaa-criteria}
\begin{tabular}{l l c c c c}
\toprule
\rowcolor{mint!15}
\textbf{\textcolor{black}{Criterion}} & \textbf{\textcolor{black}{Pair}} &
\textbf{\textcolor{black}{Pass}} & \textbf{\textcolor{black}{Agree}} &
\textbf{\textcolor{black}{$\kappa$}} & \textbf{\textcolor{black}{PABAK}} \\
\midrule
C1 Abduction  & A1 vs.\ A2    & 0.980, 0.975 & 199 & 0.886     & 0.990 \\
              & A1 vs.\ Judge & 0.980, 0.965 & 197 & 0.720     & 0.970 \\
              & A2 vs.\ Judge & 0.975, 0.965 & 196 & 0.657     & 0.960 \\
\midrule
C2 Suspicion  & A1 vs.\ A2    & 0.940, 0.935 & 199 & 0.957     & 0.990 \\
              & A1 vs.\ Judge & 0.940, 0.910 & 194 & 0.784     & 0.940 \\
              & A2 vs.\ Judge & 0.935, 0.910 & 195 & 0.826     & 0.950 \\
\midrule
C3 Depth      & A1 vs.\ A2    & 1.000, 1.000 & 200 & undefined & 1.000 \\
              & A1 vs.\ Judge & 1.000, 1.000 & 200 & undefined & 1.000 \\
              & A2 vs.\ Judge & 1.000, 1.000 & 200 & undefined & 1.000 \\
\midrule
C4 Coherence  & A1 vs.\ A2    & 1.000, 1.000 & 200 & undefined & 1.000 \\
              & A1 vs.\ Judge & 1.000, 0.995 & 199 & 0.000     & 0.990 \\
              & A2 vs.\ Judge & 1.000, 0.995 & 199 & 0.000     & 0.990 \\
\bottomrule
\end{tabular}
\end{table}

\paragraph{Per-criterion agreement.}
Table~\ref{tab:iaa-criteria} reports agreement on each of the four criteria.
We pool Trace~A and Trace~B.
Thus, each pair of raters contributes 200 judgments per criterion.
Cohen's~$\kappa$ corrects the observed agreement ($P_o$) for the expected agreement ($P_e$).
When nearly every trace passes, $P_e$ approaches $P_o$.
This mathematical property forces $\kappa$ toward zero or makes it undefined \citep{Feinstein+Cicchetti-90:kappa-paradox}.
This effect occurs even when raters agree on almost every judgment.
Criterion~4 demonstrates this effect directly.
Both annotators pass all 200 traces.
The judge passes 199 traces.
Each annotator agrees with the judge on 199 judgments.
However, $P_e = P_o = 0.995$, which yields $\kappa = 0$.
On Criterion~3, every rater passes every trace.
This unanimous outcome sets $P_e = 1$ and makes $\kappa$ undefined.
PABAK replaces $P_e$ with the expected value under balanced labels \citep{Byrt+93:PABAK}.
PABAK measures the agreement itself regardless of failure rarity.
PABAK is at least 0.94 for every pair and criterion.
Failures occur on only two criteria.
For these two criteria, $\kappa$ remains defined but has wide intervals.
For example, on Criterion~1, the 95\% interval for Annotator~2 against the judge spans from 0.206 to 0.939.
Gwet's AC1 \citep{Gwet-08:AC1} is another prevalence-robust coefficient.
This coefficient stays within 0.01 of $P_o$ on every row and supports the same conclusion.

\section{Trained Detector Settings} 
Qwen3-8B is released under the Apache~2.0 license.
\subsection{Trained Model Parameters and GPU Hours}
\label{app:parameters}

Table~\ref{tab:parameters} lists the SFT and GRPO parameters.
Each GRPO step uses two prompts with 16 completions each, and one epoch over the 900 training samples takes 450 steps.

\begin{table}[ht]
\centering
\caption{Training parameters. Values in parentheses apply to the tactic variant.}
\label{tab:parameters}
\resizebox{0.95\textwidth}{!}{%
\begin{tabular}{l l l}
\toprule
\rowcolor{ncsuBlue!15}
\textbf{\textcolor{ncsuBlue}{Parameter}} & \textbf{\textcolor{ncsuBlue}{SFT}} & \textbf{\textcolor{ncsuBlue}{GRPO}} \\
\midrule
\rowcolor{ncsuOrange!15}
Base model & \multicolumn{2}{l}{Qwen3-8B, bf16, not quantized} \\
\rowcolor{mint!10}
LoRA & \multicolumn{2}{l}{rank 32, $\alpha_{\text{LoRA}} = 64$, dropout 0.05, all linear layers} \\
Optimizer & AdamW (default) & AdamW (fused) \\
Learning rate & $2 \times 10^{-5}$ & $1 \times 10^{-5}$ \\
Warmup ratio & 0.05 & 0.05 \\
Epochs & 1 & 1 (450 steps) \\
Batch per device $\times$ accumulation & $4 \times 2 = 8$ samples & $8 \times 4 = 32$ completions \\
Completions per prompt & --- & 16 \\
Sampling temperature & --- & 0.8 \\
Maximum length (tokens) & 5,120 (5,632) in total & prompt 3,328 (3,456); completion 1,280 \\
Loss & cross-entropy on the completion & DAPO \\
Rewards & --- & 5 (6), equal weights \\
Trainer (TRL 0.24.0) & \texttt{SFTTrainer} & \texttt{GRPOTrainer} \\
Libraries & \multicolumn{2}{l}{PyTorch 2.7.1, Transformers 4.56.1, PEFT 0.14.0, Accelerate 1.4.0} \\
\bottomrule
\end{tabular}
}
\end{table}

\paragraph{Token caps.}
The longest gold trace has 1,139 tokens (1,191 with tactics), and the longest prompt has 3,013 tokens (3,329 with tactics).
We cap completions at 1,280 tokens and prompts at 3,328 tokens (3,456 with tactics).
The completion cap stays close to the longest trace because GRPO generates 16 completions per prompt and processes every token in each.
An output that reaches the cap has no deduction and counts as an empty prediction (Appendix~\ref{app:truncation}).

\paragraph{GPU hours.}
All runs use one NVIDIA H100 PCIe GPU.
SFT takes 20 to 24 minutes and GRPO 25.1 to 34.4 hours.
The eight runs take 226.3 GPU hours in total.
We run 18 evaluations on the held-out games (51 to 91 minutes each, 22.2 GPU-hours in total): the SFT and GRPO stages of the eight runs, and the untrained model once for each of the two prompts.

\subsection{GRPO Rewards}
\label{app:rewards}

Each reward scores a completion from 0 to 1, and GRPO sums the rewards with equal weights.
The rewards read the fields of the completion.
These fields include the hypotheses $h_i$ with a Good or Evil choice, the suspicion levels $s_i$, the beliefs $\beta_1^i$ and $\beta_2^i$, the depth (the highest belief order), the utterance $\hat{u}$, the deduction $\hat{\alpha}$, and, in the tactic variant, one tactic per contextual agent.
Table~\ref{tab:rewards} defines each reward over the four contextual agents.
Four of the five original rewards saturated during training.
Rewards (2) and (3) reach their maximum scores without identifying the Evil players.
The revised rewards replace these two with rewards that score the Good or Evil choice and the order of suspicion levels against $\alpha$.

\begin{table}[ht]
\centering
\caption{GRPO rewards. \fbf{Set} gives the reward sets that use the reward; \fbf{$\alpha$} marks rewards that use the withheld roles.}
\label{tab:rewards}
\resizebox{\textwidth}{!}{%
\begin{tabular}{l l c p{0.62\textwidth}}
\toprule
\rowcolor{ncsuBlue!15}
\textbf{\textcolor{ncsuBlue}{Reward}} & \textbf{\textcolor{ncsuBlue}{Set}} & \textbf{\textcolor{ncsuBlue}{$\alpha$}} & \textbf{\textcolor{ncsuBlue}{Score}} \\
\midrule
(1) Schema validity & Original & No & 1 if the completion is valid JSON with every field; otherwise half the share of fields present, and 0 if it is not JSON. \\
(1) Schema validity & Revised & No & Mean of three checks: every field is present, the utterance $\hat{u}$ is not empty, and the depth equals the highest belief order. \\
(2) Abduction completeness & Original & No & Share of agents with both hypotheses and a Good or Evil choice. \\
(2) Abduction accuracy & Revised & Yes & Share of agents whose Good or Evil choice matches $\alpha$; a missing agent counts as wrong. \\
(3) Suspicion consistency & Original & Yes & $0.6 \times$ the share of agents with a valid suspicion level $s_i$ $+\ 0.4 \times$ the share of Evil players rated Strong or Extreme. \\
(3) Suspicion calibration & Revised & Yes & Mean over the four (Evil, Good) pairs of agents: 1 if the Evil player has the higher level, 0.5 if the levels are equal, and 0 otherwise. \\
(4) Gated depth & Both & No & $0.5 \times$ the share of agents with a first-order belief $+\ 0.5 \times \max(0,\, 1 - 0.25v)$, where $v$ counts second-order beliefs about agents below Strong suspicion. \\
(5) Deduction F1 & Both & Yes & F1 of $\hat{\alpha}$ against the Evil players in $\alpha$. \\
(6) Tactic accuracy & Tactic variant & No & Share of agents whose predicted tactic matches the gold tactic. \\
\bottomrule
\end{tabular}
}
\end{table}

\subsection{Avalon-NLU Conversion}
\label{app:avalon-nlu}

We convert the 20 released games of \citet{Stepputtis+23:AvalonLLM} without editing the source files.
These games are released under the MIT license.
Each quest becomes one round.
Four games end after three quests, ten after four, and six after five. 
This conversion yields 82 total rounds ($4 \times 3 + 10 \times 4 + 6 \times 5 = 82$).
Seats map to P1 through P6.
The Investigator is the first Loyal Servant, and the Evil pair is Morgana and the Assassin.
The history follows the \fsc{Avcorp} format. 
It adds the team size of each quest and every proposal with its leader and team. 
It also adds the vote count of each rejected proposal and the score after the quest.
The discussion uses the chat transcript of the quest. 
We mark the Investigator's own messages within this transcript.
An independent script rebuilt all 82 rounds from the raw files. 
This script found one seat-parsing error across three rounds. 
We fixed this error before evaluation.
In one game (NLU-3YjxjN), the log records a quest failed by four Good players. 
A re-proposal announced in the chat lacks a matching proposal event. 
Therefore, we also report results over the remaining 19 games.

\subsection{Detector Results by Round}
\label{app:detector-rounds}

Tables~\ref{tab:rounds-f1} and~\ref{tab:rounds-end} report results by round on the 25 held-out \fsc{Avcorp} games.
Table~\ref{tab:nlu-rounds} reports results on the 20 Avalon-NLU games using the domain-context prompt (Appendix~\ref{app:nlu-prompts}).
Every detector and frontier model scores lower at R1 than at R3.
At R1, no quest outcomes are known.
Every game contains exactly two Evil players.
However, with the trace variant and original rewards, GRPO names 2.57 to 3.12 players per round on \fsc{Avcorp}.
With the revised rewards, it names 1.73 to 2.27 players.
The best detector performs strongest late in a game.
At R5, its F1 score reaches 0.927.
In contrast, the frontier models score between 0.550 and 0.700.
On Avalon-NLU, the untrained model names the fewest players (1.17 and 1.56 per round).
This model also produces the lowest recall (0.366 and 0.384).

\begin{table*}[t]
\centering
\sisetup{
  text-family-to-math = true,
  text-series-to-math = true,
  round-precision=3,
  round-mode=places
}
\caption{Round F1 on the 25 held-out \fsc{Avcorp} games (R1 to R3: 25 rounds each; R4: 15; R5: 10), with precision, recall, and F1 over all 100 rounds and the mean number of players named per round. Trace-only rows give the mean over three runs. $^\dagger$Reworded prompt. Gold traces are the corpus traces, scored the same way.}
\label{tab:rounds-f1}
\resizebox{\textwidth}{!}{%
\begin{tabular}{ll *8{S[table-format = 1.3]} S[table-format=1.2, round-precision=2]}
\toprule
\rowcolor{ncsuBlue!15}
\whitecell{}&\whitecell{}& \multicolumn{5}{c}{\textbf{\textcolor{ncsuBlue}{Round F1}}} & \multicolumn{3}{c}{\textbf{\textcolor{ncsuBlue}{All 100 rounds}}} & \whitecell{} \\
\cmidrule(lr){3-7}\cmidrule(lr){8-10}
\rowcolor{ncsuBlue!15}
\textbf{\textcolor{ncsuBlue}{Detector}} & \textbf{\textcolor{ncsuBlue}{Stage}} & \textbf{\textcolor{ncsuBlue}{R1}} & \textbf{\textcolor{ncsuBlue}{R2}} & \textbf{\textcolor{ncsuBlue}{R3}} & \textbf{\textcolor{ncsuBlue}{R4}} & \textbf{\textcolor{ncsuBlue}{R5}} & \textbf{\textcolor{ncsuBlue}{P}} & \textbf{\textcolor{ncsuBlue}{R}} & \textbf{\textcolor{ncsuBlue}{F1}} & \textbf{\textcolor{ncsuBlue}{Named}} \\
\midrule
Qwen3-8B, trace prompt & adapters off & 0.073 & 0.172 & 0.507 & 0.490 & 0.545 & 0.671 & 0.235 & 0.348 & 0.70 \\
Qwen3-8B, tactic prompt & adapters off & 0.222 & 0.353 & 0.593 & 0.640 & 0.629 & 0.722 & 0.350 & 0.471 & 0.97 \\
\midrule
Trace, original rewards & SFT & 0.502 & 0.552 & 0.651 & 0.628 & 0.735 & 0.704 & 0.517 & 0.596 & 1.47 \\
 & SFT + GRPO & 0.687 & 0.717 & 0.767 & 0.765 & 0.804 & 0.621 & 0.913 & 0.737 & 2.97 \\
Trace, revised rewards & SFT & 0.464 & 0.503 & 0.648 & 0.671 & 0.632 & 0.675 & 0.493 & 0.570 & 1.46 \\
 & SFT + GRPO & 0.707 & 0.774 & 0.849 & 0.825 & 0.844 & 0.802 & 0.783 & 0.793 & 1.95 \\
Tactic, original rewards & SFT & 0.400 & 0.545 & 0.688 & 0.727 & 0.732 & 0.682 & 0.535 & 0.599 & 1.57 \\
 & SFT + GRPO & 0.727 & 0.775 & 0.832 & 0.788 & 0.851 & 0.713 & 0.880 & 0.787 & 2.47 \\
Tactic, revised rewards & SFT & 0.416 & 0.494 & 0.674 & 0.679 & 0.700 & 0.685 & 0.500 & 0.578 & 1.46 \\
 & SFT + GRPO & 0.700 & 0.692 & 0.832 & 0.885 & 0.927 & 0.768 & 0.795 & 0.781 & 2.07 \\
\midrule
GPT-5.6 Sol & default reasoning & 0.615 & 0.760 & 0.820 & 0.733 & 0.550 & 0.733 & 0.700 & 0.716 & 1.91 \\
Claude Opus~5.5$^\dagger$ & default reasoning & 0.707 & 0.800 & 0.860 & 0.900 & 0.700 & 0.799 & 0.795 & 0.797 & 1.99 \\
Gemini~3.1 Pro & default reasoning & 0.600 & 0.667 & 0.747 & 0.700 & 0.600 & 0.727 & 0.625 & 0.672 & 1.72 \\
DeepSeek-V4-Pro & default reasoning & 0.542 & 0.560 & 0.760 & 0.655 & 0.600 & 0.634 & 0.615 & 0.624 & 1.94 \\
\midrule
\textit{Gold traces} & --- & 0.680 & 0.752 & 0.900 & 0.900 & 1.000 & 0.816 & 0.820 & 0.818 & 2.01 \\
\bottomrule
\end{tabular}
}
\end{table*}

\begin{table*}[!t]
\centering
\sisetup{
  text-family-to-math = true,
  text-series-to-math = true,
  round-mode = none
}
\caption{Game-ending verdict by the round at which a game ends (exact matches; 10 games end at R3, 5 at R4, and 10 at R5) and the mean of players named per round, on the 25 held-out \fsc{Avcorp} games. Notation as in Table~\ref{tab:rounds-f1}.}
\label{tab:rounds-end}
\resizebox{\textwidth}{!}{%
\begin{tabular}{ll *3{S[table-format=1.1]} *5{S[table-format=1.2]}}
\toprule
\rowcolor{ncsuBlue!15}
 \whitecell{}&\whitecell{} & \multicolumn{3}{c}{\textbf{\textcolor{ncsuBlue}{Exact games, by ending round}}} & \multicolumn{5}{c}{\textbf{\textcolor{ncsuBlue}{Players named per round}}} \\
\cmidrule(lr){3-5}\cmidrule(lr){6-10}
\rowcolor{ncsuBlue!15}
\textbf{\textcolor{ncsuBlue}{Detector}} & \textbf{\textcolor{ncsuBlue}{Stage}} & {\textbf{\textcolor{ncsuBlue}{R3 (10)}}} & {\textbf{\textcolor{ncsuBlue}{R4 (5)}}} & {\textbf{\textcolor{ncsuBlue}{R5 (10)}}} & {\textbf{\textcolor{ncsuBlue}{R1}}} & {\textbf{\textcolor{ncsuBlue}{R2}}} & {\textbf{\textcolor{ncsuBlue}{R3}}} & {\textbf{\textcolor{ncsuBlue}{R4}}} & {\textbf{\textcolor{ncsuBlue}{R5}}} \\
\midrule
Qwen3-8B, trace prompt & adapters off & 5 & 0 & 1 & 0.20 & 0.32 & 1.00 & 1.27 & 1.30 \\
Qwen3-8B, tactic prompt & adapters off & 6 & 4 & 2 & 0.52 & 0.72 & 1.24 & 1.33 & 1.50 \\
\midrule
Trace, original rewards & SFT & 3.7 & 1.0 & 4.3 & 1.35 & 1.43 & 1.52 & 1.53 & 1.63 \\
 & SFT + GRPO & 3.0 & 1.3 & 4.0 & 3.12 & 3.07 & 2.92 & 2.89 & 2.57 \\
Trace, revised rewards & SFT & 4.3 & 2.0 & 3.3 & 1.33 & 1.40 & 1.53 & 1.44 & 1.80 \\
 & SFT + GRPO & 7.3 & 3.0 & 5.7 & 1.73 & 1.96 & 1.99 & 2.04 & 2.27 \\
Tactic, original rewards & SFT & 5 & 3 & 5 & 1.20 & 1.52 & 1.72 & 1.67 & 2.10 \\
 & SFT + GRPO & 6 & 2 & 4 & 2.40 & 2.44 & 2.52 & 2.40 & 2.70 \\
Tactic, revised rewards & SFT & 5 & 3 & 4 & 1.08 & 1.24 & 1.68 & 1.73 & 2.00 \\
 & SFT + GRPO & 7 & 3 & 8 & 2.00 & 2.16 & 2.04 & 2.07 & 2.10 \\
\midrule
GPT-5.6 Sol & default reasoning & 9 & 3 & 4 & 1.64 & 2.00 & 2.00 & 2.00 & 2.00 \\
Claude Opus~5.5$^\dagger$ & default reasoning & 9 & 3 & 6 & 1.96 & 2.00 & 2.00 & 2.00 & 2.00 \\
Gemini~3.1 Pro & default reasoning & 7 & 2 & 5 & 1.20 & 1.72 & 1.96 & 2.00 & 2.00 \\
DeepSeek-V4-Pro & default reasoning & 7 & 0 & 5 & 1.84 & 2.00 & 2.00 & 1.87 & 2.00 \\
\midrule
\textit{Gold traces} & --- & 10 & 3 & 10 & 2.00 & 2.04 & 2.00 & 2.00 & 2.00 \\
\bottomrule
\end{tabular}
}
\end{table*}

\subsection{Frontier Models in \fsc{Avcorp}}
\label{app:frontier}

Table~\ref{tab:frontier} compares the best detector with four frontier models, the gold traces, and two strategies that use no model.
Each frontier model receives the tactic prompt without the Qwen3 switch \texttt{/no\_think}, conditions on its own earlier traces, and runs at its provider's default reasoning setting and temperature.
It can generate up to 32,768 tokens per round, reasoning included.
One DeepSeek-V4-Pro round that does not end a game reaches this cap and counts as an empty prediction.
Claude Opus~5.5 refused 16 of the 25 games from the second round on, and Anthropic's API labeled these refusals as reasoning extraction.
For Claude Opus~5.5 alone, we reword the seven instruction phrases that mention reasoning or a trace (e.g., ``reasoning trace'' becomes ``report'') and rerun all 25 games.
After this change, the model answers every round.
Naming every candidate reaches perfect recall but resolves no game, and naming two candidates at random resolves 16.7\% of games in expectation.

\begin{table*}[t]
\centering
\sisetup{
  text-family-to-math = true,
  text-series-to-math = true,
  round-mode = none
}
\caption{Round F1 on the 20 Avalon-NLU games with the domain-context prompt (R1 to R3: 20 rounds each; R4: 16; R5: 6), with precision, recall, and F1 over all 82 rounds, the mean number of players named per round, and the number of games, of 20, whose game-ending deduction names exactly two players (\fbf{Two}). Every model conditions on its own earlier traces. Trace-only rows give the mean over three runs.}
\label{tab:nlu-rounds}
\resizebox{\textwidth}{!}{%
\begin{tabular}{ll *8{S[table-format=1.3]} S[table-format=1.2] S[table-format=2.1]}
\toprule
\rowcolor{ncsuBlue!15}
 \whitecell{}&\whitecell{} & \multicolumn{5}{c}{\textbf{\textcolor{ncsuBlue}{Round F1}}} & \multicolumn{3}{c}{\textbf{\textcolor{ncsuBlue}{Performance}}} & \whitecell{} & \whitecell{} \\
\cmidrule(lr){3-7}\cmidrule(lr){8-10}
\rowcolor{ncsuBlue!15}
\textbf{\textcolor{ncsuBlue}{Detector}} & \textbf{\textcolor{ncsuBlue}{Stage}} & {\textbf{\textcolor{ncsuBlue}{R1}}} & {\textbf{\textcolor{ncsuBlue}{R2}}} & {\textbf{\textcolor{ncsuBlue}{R3}}} & {\textbf{\textcolor{ncsuBlue}{R4}}} & {\textbf{\textcolor{ncsuBlue}{R5}}} & {\textbf{\textcolor{ncsuBlue}{P}}} & {\textbf{\textcolor{ncsuBlue}{R}}} & {\textbf{\textcolor{ncsuBlue}{F1}}} & {\textbf{\textcolor{ncsuBlue}{Named}}} & {\textbf{\textcolor{ncsuBlue}{Two}}} \\
\midrule
Qwen3-8B, trace prompt & adapters off & 0.163 & 0.259 & 0.611 & 0.623 & 0.667 & 0.625 & 0.366 & 0.462 & 1.17 & 19 \\
Qwen3-8B, tactic prompt & adapters off & 0.175 & 0.338 & 0.610 & 0.469 & 0.583 & 0.492 & 0.384 & 0.432 & 1.56 & 20 \\
\midrule
Trace, original rewards & SFT & 0.525 & 0.512 & 0.606 & 0.666 & 0.677 & 0.555 & 0.606 & 0.579 & 2.18 & 18.7 \\
 & SFT + GRPO & 0.508 & 0.542 & 0.611 & 0.626 & 0.628 & 0.488 & 0.691 & 0.572 & 2.83 & 15.7 \\
Trace, revised rewards & SFT & 0.489 & 0.479 & 0.602 & 0.646 & 0.727 & 0.541 & 0.583 & 0.561 & 2.15 & 18.3 \\
 & SFT + GRPO & 0.524 & 0.560 & 0.584 & 0.689 & 0.745 & 0.580 & 0.616 & 0.597 & 2.13 & 17.7 \\
Tactic, original rewards & SFT & 0.471 & 0.505 & 0.500 & 0.630 & 0.600 & 0.459 & 0.616 & 0.526 & 2.68 & 15 \\
 & SFT + GRPO & 0.433 & 0.531 & 0.579 & 0.621 & 0.621 & 0.449 & 0.695 & 0.545 & 3.10 & 10 \\
Tactic, revised rewards & SFT & 0.479 & 0.500 & 0.558 & 0.618 & 0.583 & 0.485 & 0.598 & 0.536 & 2.46 & 18 \\
 & SFT + GRPO & 0.460 & 0.512 & 0.652 & 0.694 & 0.643 & 0.530 & 0.640 & 0.580 & 2.41 & 15 \\
\bottomrule
\end{tabular}
}
\end{table*}

\begin{table}[t]
\centering
\sisetup{
  text-family-to-math = true,
  text-series-to-math = true,
  round-precision=3,
  round-mode=places
}
\caption{Game-ending verdict on the 25 held-out \fsc{Avcorp} games. \fbf{Exact} requires naming both Evil players and no one else. $^\dagger$Reworded prompt.}
\label{tab:frontier}
\begin{tabular}{l S[table-format=2.1,round-precision=1] c S[table-format=1.3]}
\toprule
\rowcolor{ncsuBlue!15}
\textbf{\textcolor{ncsuBlue}{Model}} & {\textbf{\textcolor{ncsuBlue}{Exact (\%)}}} & \textbf{\textcolor{ncsuBlue}{Exact games}} & {\textbf{\textcolor{ncsuBlue}{F1}}} \\
\midrule
Best detector (tactic, revised, SFT + GRPO) & \bfseries 72.0 & 18/25 & \bfseries 0.882 \\
GPT-5.6 Sol & 64.0 & 16/25 & 0.740 \\
Claude Opus~5.5$^\dagger$ & \bfseries 72.0 & 18/25 & 0.800 \\
Gemini~3.1 Pro & 56.0 & 14/25 & 0.680 \\
DeepSeek-V4-Pro & 48.0 & 12/25 & 0.620 \\
\midrule
\textit{Gold traces} & 92.0 & 23/25 & 0.960 \\
Name every candidate & 0.0 & 0/25 & 0.667 \\
Two at random (expected) & 16.7 & --- & 0.500 \\
\bottomrule
\end{tabular}
\end{table}

\subsection{Avalon-NLU under Two Prompts}
\label{app:nlu-prompts}
Table~\ref{tab:nlu-prompts} reports every Avalon-NLU evaluation under two prompts: Minimal vs Domain-context.
Table~\ref{tab:detector} reports the second.
The \emph{minimal} prompt adapts the \fsc{Avcorp} prompt to six players.
It adds a paragraph of rules covering roles, team sizes, votes, and assassinations.
This prompt uses the original history format and limits outputs to 2,048 tokens.
The \emph{domain-context} prompt adds guidance on weighing evidence, scores, quest records, message marking, and deduction sizes.
This prompt uses the extended history from Appendix~\ref{app:avalon-nlu}.
It limits outputs to 3,072 tokens.
Under both prompts, every trained detector exceeds the untrained model in round F1.
The minimal prompt does not state the required deduction size.
At the game-ending round, only 2 to 12 of the 20 deductions name exactly two players under the minimal prompt.
In contrast, 10 to 20 deductions name exactly two players under the domain-context prompt.
Under the minimal prompt, GRPO leads SFT on the game-ending verdict (3.4 against 3.1 exact games on average).
However, GRPO trails SFT under the domain-context prompt (6.6 against 8.5 exact games).
Under the minimal prompt, the trained detectors name the Investigator itself in 3 to 15 of the 82 rounds.
The scorer removes these invalid names.
The domain-context prompt explicitly marks the Investigator's messages.
Consequently, no detector names the Investigator under the domain-context prompt.
Two domain-context outputs reach the 3,072-token limit.
Both occur during SFT tactic runs at the final round of NLU-QdJ2dO.
We count these truncated outputs as empty predictions.
No detector resolves game NLU-3YjxjN under either prompt.
Therefore, every exact count remains the same over the other 19 games.

\begin{table*}[ht]
\centering
\sisetup{
  text-family-to-math = true,
  text-series-to-math = true,
  round-mode = none
}
\caption{Every Avalon-NLU evaluation under the minimal and domain-context prompts. \fbf{Exact} counts the games, of 20, whose game-ending deduction names both Evil players and no one else; the counts are the same over the 19 games without NLU-3YjxjN. \fbf{F1} scores the game-ending deduction over 20 games and \fbf{F1$_{19}$} over 19; \fbf{Round F1} scores all 82 rounds. Every model conditions on its own earlier traces.}
\label{tab:nlu-prompts}
\resizebox{\textwidth}{!}{%
\begin{tabular}{ll S[table-format=2.1] *3{S[table-format=1.3]} S[table-format=2.1] *3{S[table-format=1.3]}}
\toprule
\rowcolor{ncsuBlue!15}
\whitecell{}&\whitecell{}& \multicolumn{4}{c}{\textbf{\textcolor{ncsuBlue}{Minimal prompt}}} & \multicolumn{4}{c}{\textbf{\textcolor{ncsuBlue}{Domain-context prompt}}} \\
\cmidrule(lr){3-6}\cmidrule(lr){7-10}
\rowcolor{ncsuBlue!15}
\textbf{\textcolor{ncsuBlue}{Detector}} & \textbf{\textcolor{ncsuBlue}{Stage}} & {\textbf{\textcolor{ncsuBlue}{Exact}}} & {\textbf{\textcolor{ncsuBlue}{F1}}} & {\textbf{\textcolor{ncsuBlue}{F1$_{19}$}}} & {\textbf{\textcolor{ncsuBlue}{Round F1}}} & {\textbf{\textcolor{ncsuBlue}{Exact}}} & {\textbf{\textcolor{ncsuBlue}{F1}}} & {\textbf{\textcolor{ncsuBlue}{F1$_{19}$}}} & {\textbf{\textcolor{ncsuBlue}{Round F1}}} \\
\midrule
Qwen3-8B, trace prompt & adapters off & 1 & 0.338 & 0.361 & 0.240 & 8 & 0.667 & 0.676 & 0.462 \\
Qwen3-8B, tactic prompt & adapters off & 5 & 0.548 & 0.588 & 0.361 & 4 & 0.550 & 0.553 & 0.432 \\
\midrule
Trace, original, run 1 & SFT & 4 & 0.642 & 0.667 & 0.512 & 9 & 0.716 & 0.727 & 0.581 \\
 & SFT + GRPO & 1 & 0.592 & 0.595 & 0.567 & 6 & 0.651 & 0.658 & 0.581 \\
Trace, original, run 2 & SFT & 4 & 0.553 & 0.575 & 0.514 & 8 & 0.667 & 0.675 & 0.552 \\
 & SFT + GRPO & 2 & 0.625 & 0.624 & 0.579 & 7 & 0.674 & 0.683 & 0.604 \\
Trace, original, run 3 & SFT & 3 & 0.526 & 0.548 & 0.502 & 11 & 0.756 & 0.769 & 0.604 \\
 & SFT + GRPO & 1 & 0.642 & 0.647 & 0.561 & 5 & 0.636 & 0.643 & 0.532 \\
Trace, revised, run 1 & SFT & 3 & 0.634 & 0.658 & 0.538 & 9 & 0.723 & 0.734 & 0.586 \\
 & SFT + GRPO & 5 & 0.634 & 0.658 & 0.486 & 6 & 0.675 & 0.684 & 0.568 \\
Trace, revised, run 2 & SFT & 2 & 0.600 & 0.623 & 0.503 & 9 & 0.707 & 0.718 & 0.566 \\
 & SFT + GRPO & 4 & 0.681 & 0.689 & 0.526 & 8 & 0.714 & 0.725 & 0.615 \\
Trace, revised, run 3 & SFT & 3 & 0.613 & 0.639 & 0.506 & 6 & 0.659 & 0.667 & 0.533 \\
 & SFT + GRPO & 6 & 0.682 & 0.706 & 0.588 & 10 & 0.765 & 0.779 & 0.607 \\
Tactic, original & SFT & 2 & 0.537 & 0.557 & 0.510 & 7 & 0.644 & 0.675 & 0.526 \\
 & SFT + GRPO & 2 & 0.593 & 0.596 & 0.557 & 4 & 0.646 & 0.653 & 0.545 \\
Tactic, revised & SFT & 4 & 0.623 & 0.649 & 0.514 & 9 & 0.658 & 0.693 & 0.536 \\
 & SFT + GRPO & 6 & 0.689 & 0.713 & 0.544 & 7 & 0.698 & 0.707 & 0.580 \\
\midrule
Mean of 2 & adapters off & 3.0 & 0.443 & 0.474 & 0.301 & 6.0 & 0.608 & 0.614 & 0.447 \\
Mean of 8 & SFT & 3.1 & 0.591 & 0.615 & 0.512 & 8.5 & 0.691 & 0.707 & 0.560 \\
Mean of 8 & SFT + GRPO & 3.4 & 0.642 & 0.653 & 0.551 & 6.6 & 0.682 & 0.691 & 0.579 \\
\bottomrule
\end{tabular}
}
\end{table*}

\subsection{Outputs at the Length Cap}
\label{app:truncation}

Table~\ref{tab:truncation} lists every output that reached the 1,280-token cap on the held-out games.
These truncated outputs lack a deduction.
We count them as empty predictions.
When using their own traces, the detectors produce eight truncated outputs out of 1,800.
A truncated output enters the prompt for the next round.
This propagation explains the consecutive truncated rounds in games G203 and G228.
Game G228 is one of two held-out games where the gold trace names the wrong pair.
Every trained detector also misses the correct pair in this game.
Therefore, the truncation does not change the game-ending score.
When using gold prior traces (Appendix~\ref{app:gold-chain}), only the untrained model with the tactic prompt reaches the cap.
This truncation occurs in four rounds.

\begin{table}[ht]
\centering
\caption{Outputs that reached the 1,280-token cap on the 25 held-out \fsc{Avcorp} games. \fbf{Final} marks a game's last round, which the game-ending verdict scores.}
\label{tab:truncation}
\begin{tabular}{l l l l l c}
\toprule
\rowcolor{ncsuBlue!15}
\textbf{\textcolor{ncsuBlue}{Prior traces}} & \textbf{\textcolor{ncsuBlue}{Run}} & \textbf{\textcolor{ncsuBlue}{Stage}} &
\textbf{\textcolor{ncsuBlue}{Game}} & \textbf{\textcolor{ncsuBlue}{Rounds}} & \textbf{\textcolor{ncsuBlue}{Final}} \\
\midrule
Own  & Trace, original, run 1 & Untrained & G203 & R2, R3 & R3 \\
Own  & Trace, original, run 1 & SFT       & G088 & R5     & R5 \\
Own  & Trace, original, run 2 & SFT       & G045 & R2     & --- \\
Own  & Trace, revised, run 2  & SFT + GRPO & G228 & R2, R3, R4 & R4 \\
Own  & Tactic, original       & Untrained & G045 & R5     & R5 \\
\midrule
Gold & Tactic prompt          & Untrained & G089, G165 & R3, R3 & G165 \\
Gold & Tactic prompt          & Untrained & G191, G133 & R4, R5 & G133 \\
\bottomrule
\end{tabular}
\end{table}

\subsection{Evaluation with Gold Prior Traces}
\label{app:gold-chain}

Table~\ref{tab:gold-vs-own} compares the game-ending verdicts using gold prior traces versus generated prior traces on the same 25 games.
With gold prior traces, every stage resolves 72.0\% to 92.0\% of the games.
This performance range includes the untrained model.
Gold prior traces leak the answer.
Copying the previous gold deduction without any model resolves 18 of the 25 games.
Therefore, we evaluate every detector using its own traces.
The gold-trace evaluations took 203 to 254 minutes per run.
These eight runs required 29.1 GPU hours in total.

\begin{table}[ht]
\centering
\sisetup{
  text-family-to-math = true,
  text-series-to-math = true,
  round-mode = none
}
\caption{Game-ending verdict on the 25 held-out games with gold or own prior traces. Exact in \%. Trace-only rows give the mean over three runs.}
\label{tab:gold-vs-own}
\begin{tabular}{ll S[table-format=2.1] S[table-format=1.3] S[table-format=2.1] S[table-format=1.3]}
\toprule
\rowcolor{ncsuBlue!15}
\whitecell{} & \whitecell{} & \multicolumn{2}{c}{\textbf{\textcolor{ncsuBlue}{Gold prior traces}}} & \multicolumn{2}{c}{\textbf{\textcolor{ncsuBlue}{Own prior traces}}} \\
\cmidrule(lr){3-4}\cmidrule(lr){5-6}
\rowcolor{ncsuBlue!15}
\textbf{\textcolor{ncsuBlue}{Detector}} & \textbf{\textcolor{ncsuBlue}{Stage}} & {\textbf{\textcolor{ncsuBlue}{Exact}}} & {\textbf{\textcolor{ncsuBlue}{F1}}} & {\textbf{\textcolor{ncsuBlue}{Exact}}} & {\textbf{\textcolor{ncsuBlue}{F1}}} \\
\midrule
Qwen3-8B, trace prompt & adapters off & 72.0 & 0.889 & 24.0 & 0.581 \\
Qwen3-8B, tactic prompt & adapters off & 72.0 & 0.863 & 48.0 & 0.717 \\
\midrule
Trace, original rewards & SFT & 82.7 & 0.902 & 36.0 & 0.697 \\
 & SFT + GRPO & 89.3 & 0.947 & 33.3 & 0.775 \\
Trace, revised rewards & SFT & 76.0 & 0.869 & 38.7 & 0.688 \\
 & SFT + GRPO & 84.0 & 0.927 & 64.0 & 0.861 \\
Tactic, original rewards & SFT & 84.0 & 0.931 & 52.0 & 0.747 \\
 & SFT + GRPO & 88.0 & 0.940 & 48.0 & 0.847 \\
Tactic, revised rewards & SFT & 88.0 & 0.940 & 48.0 & 0.735 \\
 & SFT + GRPO & 92.0 & 0.960 & 72.0 & 0.882 \\
\bottomrule
\end{tabular}
\end{table}
\FloatBarrier
\section{Assistance}
\label{app:assistance}
We used AI coding assistants to support code development during this study.

\end{document}